\documentclass[10pt,twocolumn,letterpaper]{article}

\usepackage[pagenumbers]{cvpr} 

\usepackage{graphicx}
\usepackage{amsmath}
\usepackage{amssymb}
\usepackage{booktabs}

\usepackage{multirow}
\usepackage[table,xcdraw,dvipsnames]{xcolor}
\usepackage{tabularx}
\usepackage{kotex}
\usepackage{mathtools}
\usepackage{dsfont}
\usepackage{bbm}
\DeclareMathOperator*{\argmin}{\arg\min}
\DeclareMathOperator*{\argmax}{\arg\max}
\usepackage{enumitem}
\usepackage{diagbox}
\renewcommand{\paragraph}[1]{\vspace{1mm}\noindent\textbf{#1}}

\usepackage[pagebackref,breaklinks,colorlinks]{hyperref}

\usepackage[capitalize]{cleveref}
\crefname{section}{Sec.}{Secs.}
\Crefname{section}{Section}{Sections}
\Crefname{table}{Table}{Tables}
\crefname{table}{Tab.}{Tabs.}

\def\cvprPaperID{} 
\def\confName{CVPR}
\def\confYear{2022}

\begin{document}

\title{WildNet: Learning Domain Generalized Semantic Segmentation from the Wild}

\author{Suhyeon Lee \quad\quad Hongje Seong \quad\quad Seongwon Lee \quad\quad Euntai Kim\thanks{Corresponding author.}\\
School of Electrical and Electronic Engineering, Yonsei University, Seoul, Korea\\
{\tt\small \{hyeon93, hjseong, won4113, etkim\}@yonsei.ac.kr}
}
\maketitle

\begin{abstract}
We present a new domain generalized semantic segmentation network named WildNet, which learns domain-generalized features by leveraging a variety of contents and styles from the wild.
In domain generalization, the low generalization ability for unseen target domains is clearly due to overfitting to the source domain.
To address this problem, previous works have focused on generalizing the domain by removing or diversifying the styles of the source domain.
These alleviated overfitting to the source-style but overlooked overfitting to the source-content.
In this paper, we propose to diversify both the content and style of the source domain with the help of the wild.
Our main idea is for networks to naturally learn domain-generalized semantic information from the wild.
To this end, we diversify styles by augmenting source features to resemble wild styles and enable networks to adapt to a variety of styles. 
Furthermore, we encourage networks to learn class-discriminant features by providing semantic variations borrowed from the wild to source contents in the feature space.
Finally, we regularize networks to capture consistent semantic information even when both the content and style of the source domain are extended to the wild.
Extensive experiments on five different datasets validate the effectiveness of our WildNet, and we significantly outperform state-of-the-art methods.
The source code and model are available online: \url{https://github.com/suhyeonlee/WildNet}.
\end{abstract}

\vspace{-0.5em}
\section{Introduction}\label{sec:introduction}

\begin{figure}
  \centering
  \begin{subfigure}{0.49\columnwidth}
    \raisebox{-\height}{\includegraphics[width=\columnwidth, height=0.55\columnwidth]{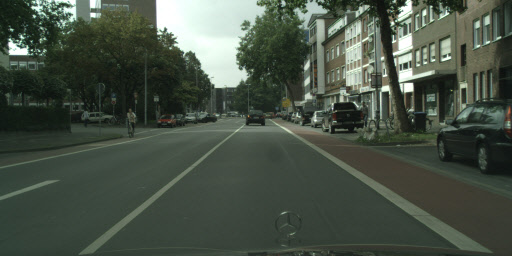}}
    \caption{Unseen domain image}
    \label{fig:compare_r50_city_img}
  \end{subfigure}
  \hfill
  \begin{subfigure}{0.49\columnwidth}
    \raisebox{-\height}{\includegraphics[width=\columnwidth, height=0.55\columnwidth]{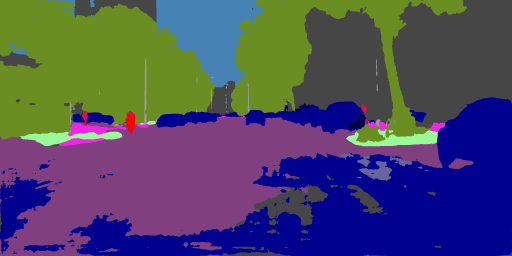}}
    \caption{Baseline (mIoU 35.16\%)}
    \label{fig:compare_r50_city_base}
  \end{subfigure}
  \begin{subfigure}{0.49\columnwidth}
    \raisebox{-\height}{\includegraphics[width=\columnwidth, height=0.55\columnwidth]{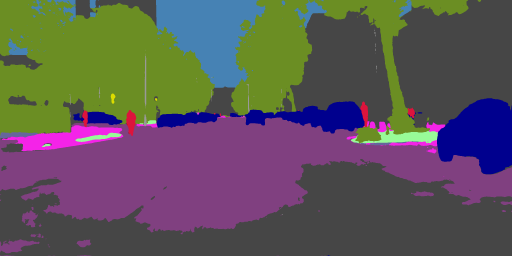}}
    \caption{RobustNet (mIoU 36.58\%)}
    \label{fig:compare_r50_city_isw}
  \end{subfigure}
  \hfill
  \begin{subfigure}{0.49\columnwidth}
    \raisebox{-\height}{\includegraphics[width=\columnwidth, height=0.55\columnwidth]{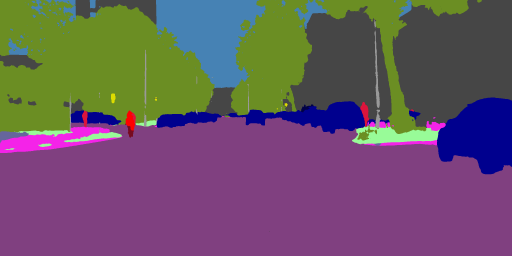}}
    \caption{Ours (mIoU 44.62\%)}
    \label{fig:compare_r50_city_ours}
  \end{subfigure}
  \vspace{-0.7em}
  \caption{Semantic segmentation results on (a) an unseen domain image. 
  The models are trained on GTAV~\cite{richter2016playing} train set and validated on Cityscapes~\cite{cordts2016cityscapes} validation set.
  (b) Baseline model overfits the source domain and performs poorly with mIoU 35.16\% on the unseen target domain.
  (c) RobustNet~\cite{choi2021robustnet}, a state-of-the-art method, improved mIoU to 36.58\% by whitening the style, but still has low generalization capability.
  (d) Our WildNet achieves superior generalization performance with mIoU 44.62\% by learning various styles and contents from the wild.
  More qualitative results on other datasets are available in 
  the supplementary material.
  }
  \vspace{-1.5em}
  \label{fig:compare_r50_city_img_base_isw_ours}
\end{figure}

Domain generalized semantic segmentation aims to better predict pixel-level semantic labels on multiple unseen target domains while learning only on the source domain. 
Unfortunately, the domain shift between the source and target domains makes a segmentation model trained on the given source data behave stupidly on the unseen target data, as shown in~\cref{fig:compare_r50_city_base}.
In domain generalization (DG), the \textit{low generalization performance} for unseen domains is obviously due to \textit{overfitting to the source domain}.
Since the model cannot see any information about the target domains in the learning process and even unlabeled target images are not provided unlike domain adaptation (DA), it over-learns the statistical distribution of the given source data.

Recently, some studies~\cite{choi2021robustnet, pan2019switchable, pan2018two, lee2021unsupervised} have proposed learning the domain-generalized content feature by `removing' domain-specific \textit{style} information from the data to prevent overfitting to the source domain.
Based on the correlation between the feature's covariance matrix and style~\cite{gatys2015texture, gatys2016image}, they assumed that only content features would remain if elements of features considered the domain-specific style were whitened~\cite{huang2018decorrelated, li2017universal, sun2016deep, roy2019unsupervised}.
However, since the content and style are not orthogonal, whitening the style may cause a loss of semantic content, which is indispensable for semantic category prediction.
As a result, they predict semantic categories from incomplete content features and have difficulty making accurate predictions, as shown in~\cref{fig:compare_r50_city_isw}.

In this paper, we propose a new domain generalized semantic segmentation network called WildNet, which learns the domain-generalized semantic feature by `extending' \textit{both content and style} to the \textit{wild}.
Although some previous works~\cite{huang2021fsdr, yue2019domain, peng2021global} utilized various styles from the wild, \eg, ImageNet~\cite{deng2009imagenet} for real styles and Painter by Numbers~\cite{nichol2016painter} for unreal styles, they overlooked that the high generalization ability comes from learning not only various styles but also various contents.
In contrast to previous studies, our main idea is to naturally learn domain-generalized semantic information by leveraging a variety of contents and styles from the wild, without forcing whitening on domain-specific styles.

To extend both content and style to the wild, we present four effective learning methods.
(i) Based on the relevance of style and feature statistics, \textit{feature stylization} diversifies the style of the source feature by transferring the statistics of the wild feature to the source feature over several layers.
(ii) To prevent overfitting to the source contents, we propose \textit{content extension learning} to increase the intra-class content variability in the latent embedding space.
Extending content from source to wild helps networks make generalized predictions on unseen contents.
(iii) To prevent overfitting to the source style, we propose \textit{style extension learning} to encourage networks to adapt to the various styles extended to the wild.
(iv) Finally, \textit{semantic consistency regularization} enables networks to capture consistent semantic information even when both the content and style of the source domain are extended to the wild.
With the proposed learning methods, our WildNet learns domain-generalized semantic features by leveraging a variety of contents and styles from the wild.
Extensive experiments over multiple domains show that our network achieves superior performance on domain generalization for semantic segmentation.

Our main contributions are as follows:
\vspace{-0.7\topsep}
\begin{itemize}
\setlength\itemsep{-0.3em}
\item We present a novel domain generalized semantic segmentation network named WildNet, which learns domain-generalized semantic features by leveraging a variety of contents and styles from the wild.
\item We propose four learning techniques to train domain-generalized networks by extending both the content and style of the source domain to the wild. 
These enable our model to make reliable predictions on various unseen target domains without training on them.
\item Our network achieves superior performance in extensive experiments on domain generalization for semantic segmentation constructed over multiple domains.
\end{itemize}

\section{Related Work}\label{sec:relatedwork}
\subsection{Domain Adaptation and Generalization}
\vspace{-0.3em}
Domain adaptation (DA) aims to increase the performance on the target domain by reducing the domain gap between the source and target domains.
In semantic segmentation, DA is exploited to tackle the effort of annotating pixel-level categories in an image.
Most DA methods train networks using the `given' target images via image translation~\cite{hoffman2017cycada, zhang2018fully, yang2020fda, he2021multi, ma2021coarse}, feature alignment~\cite{hoffman2016fcns, tsai2018learning, vu2019advent, paul2020domain, wang2020classes}, and self-training~\cite{li2019bidirectional, pan2020unsupervised, lee2021unsupervised, zhang2021prototypical, guo2021metacorrection} strategies.
However, it is hard to acquire target images from various environments during the learning process, and efforts to retrain networks are required whenever applying networks to a new target domain.

To overcome these limitations, domain generalization (DG) has recently attracted considerable attention.
However, most DG studies have focused on image classification and there are only a few recent studies on semantic segmentation.
In this study, we deal with DG for semantic segmentation.
Unlike DA, DG does not have access to the target domains during the learning process.
To make reliable predictions on various `unknown' target domains, most existing studies focus on whitening~\cite{choi2021robustnet}, normalizing~\cite{pan2018two}, and diversifying~\cite{huang2021fsdr, yue2019domain, peng2021global} styles to avoid overfitting to the style of the source domain.
This paper focuses on extending both the content and style of the source domain to the wild~\cite{deng2009imagenet}, enabling networks to learn domain-generalized semantic features from diversified contents and styles.

\subsection{Contrastive Learning}
\vspace{-0.3em}
Contrastive learning~\cite{chopra2005learning, oord2018representation} is a strategy that minimizes the distance from a positive sample and maximizes the distance from a negative sample in the embedding space.
Recently, He~\etal~\cite{he2020momentum} used a dynamic dictionary with a queue and Chen~\etal~\cite{chen2020simple} used two views of the same image as a positive pair to learn visual representations.
To diversify a positive pair, a recent work~\cite{dwibedi2021little} proposed to use the positive's nearest neighbor in the latent space as a positive.
After supervised contrastive learning~\cite{khosla2020supervised} has been proposed, there are recent efforts to apply contrastive learning to fully- and semi-supervised semantic segmentation~\cite{zhong2021pixel, zhao2021contrastive, alonso2021semi}.
To obtain positive samples, these works perform image augmentation or store features using label information in a memory bank~\cite{wu2018unsupervised}.
These enhance class discrimination in the seen source domain but do not guarantee improving class discrimination in various unseen domains.
To adapt contrastive learning to DG for semantic segmentation, we propose a learning method using the wild-stylized feature and its closest wild content as positive samples.

\begin{figure*}[t]
\centering
\includegraphics[width=0.85\textwidth]{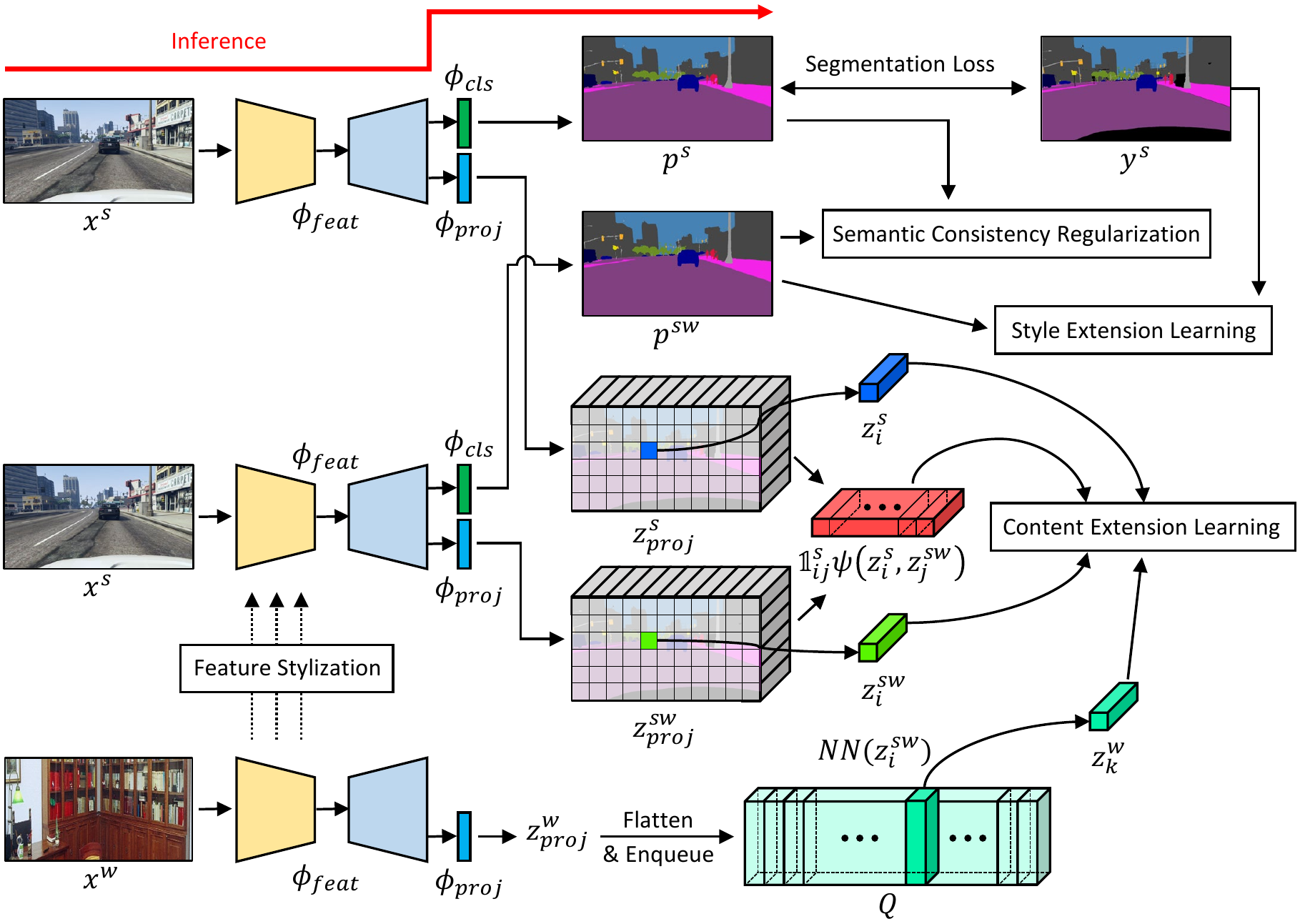}
\vspace{-1.0em}
\caption{
The overall learning process of WildNet.
Our model is trained with four proposed methods: FS, CEL, SEL, and SCR.
FS augments source features to resemble wild styles using the statistics of wild features, and the augmented features are used for CEL, SEL, and SCR.
CEL performs contrastive learning using the augmented features and the wild features closest to them as positive samples and other class features as negative samples.
SEL enables networks to learn task-specific information from features with diversified styles, and SCR regularizes networks to capture consistent semantic information from features with diversified contents and styles.
With the proposed learning methods, our model learns domain-generalized semantic features by leveraging contents and styles from the wild.
}
\vspace{-1.2em}
\label{fig:overall_architecture}
\end{figure*}

\subsection{Free ImageNet}
\vspace{-0.3em}
Most studies regard ImageNet~\cite{deng2009imagenet} as free and use it to pre-train networks.
The ImageNet pre-trained model is commonly used in various fields such as object detection~\cite{zhu2020deformable, ma2018shufflenet}, semantic segmentation~\cite{tao2020hierarchical, choi2020cars}, panoptic segmentation~\cite{cheng2020panoptic, xiong2019upsnet, mohan2021efficientps}, and video object segmentation~\cite{seong2021hierarchical}, and is considered to be the same basis.
The ImageNet pre-trained model is also used in most DA and DG for semantic segmentation methods, and ImageNet is used to borrow various styles~\cite{yue2019domain, huang2021fsdr}.
In this paper, we focus on learning domain-generalized networks by leveraging a variety of contents and styles from ImageNet.

\section{Proposed Method}\label{sec:method}
\vspace{-0.3em}
In this section, we introduce four learning techniques consisting of Feature Stylization (FS), Content Extension Learning (CEL), Style Extension Learning (SEL), and Semantic Consistency Regularization (SCR) for learning domain-generalized features by extending both the content and style of the source domain to the wild.
Our WildNet achieves superior generalization ability with them and the overall learning process is outlined in~\cref{fig:overall_architecture}.

\subsection{Problem Setup and Overview}
\vspace{-0.2em}
Domain generalization (DG) aims to enhance the generalization capability on both the seen source domain $\mathcal{S}$ and unseen target domains $\mathcal{T} = \{\mathcal{T}_1, ..., \mathcal{T}_N\}$. 
Let $\phi$ be a semantic segmentation model that outputs pixel-wise category predictions $p$ from image $x$. 
This model consists of a feature extractor $\phi_{feat}$ and classifier $\phi_{cls}$.
In DG, when we train the model, we have access to the source domain training dataset $D^s = \{(x^s,y^s) \}$ while inaccessible to the target domains, where $x^s \in {\mathbb{R}^{{H}\times{W}\times{3}}}$ is an image, $y^s \in {\mathbb{R}^{{H}\times{W}\times{K}}}$ is its pixel-wise label, and $K$ is a number of semantic categories.
The baseline model is trained with the segmentation loss
\vspace{-0.5em}
\begin{equation}\vspace{-0.5em}
    \mathcal{L}_{orig} = -\frac{1}{HW} \sum\limits_{h=1}^{H} \sum\limits_{w=1}^{W} \sum\limits_{k=1}^{K} { y^{s}_{hwk} \mathrm{log} (\phi(x^s))}.
\label{eq:orig}
\end{equation}

In this paper, we focus on extending both the content and style of the source domain to obtain high generalization performance on unknown target domains $\mathcal{T}$.
We utilize the unlabeled wild dataset $D^w = \{x^w \}$, which has various contents and styles.
At each training iteration, a random pair of source and wild images is provided as input, and the style and content of the source image are extended to the wild domain $\mathcal{W}$ in the feature space.
With the help of the wild, our network naturally learns domain-generalized semantic information from a variety of contents and styles.
After the training, the model is evaluated on validation sets of both the seen source domain $\mathcal{S}$ and unseen target domains $\mathcal{T}$.

\subsection{Feature Stylization}\label{subsec:fs}
\vspace{-0.2em}
As style is related to feature statistics~\cite{gatys2016image, gatys2017controlling, li2017demystifying, huang2017arbitrary, jing2019neural} and the distributional shift due to style differences lies mainly in shallow layers of networks~\cite{pan2018two}, the styles of features can be diversified by adjusting the statistics of features from shallow layers.
In this work, we diversify the styles of the source features with the help of wild styles by adding several AdaIN~\cite{huang2017arbitrary} layers to the feature extractor in the learning process.
This enables us to augment source features to resemble wild styles without losing spatial information.

Let $\phi_l$ be the $l$-th layer of networks $\phi$ and let $z_l$ be the feature output from $\phi_l$ when image $x$ is input into $\phi$. 
To allow networks to learn domain-generalized semantic information from various wild-style features, we swap the style of the source feature $z^s_l$ from the source image $x^s$ with the style of the wild feature $z^w_l$ from the wild image $x^w$.
In the $l$-th layer, we transfer the style of $z^w_l$ to $z^s_l$ and obtain the wild-stylized feature $z^{sw}_l$ as
\vspace{-0.5em}
\begin{equation}\label{eq:fs1}\vspace{-0.5em}
    z^{sw}_l = \sigma(z^w_l) \cfrac{z^s_l - \mu(z^s_l)}{\sigma(z^s_l)} + \mu(z^w_l)
\end{equation}
where $\mu(z_l)$ and $\sigma(z_l)$ are channel-wise mean and standard deviation of feature $z_l$, respectively.
Because the distribution of $z^{s}_l$ is re-normalized with channel-wise statistics of $z^{w}_l$, the style of $z^{s}_l$ is swapped to the wild-style while maintaining the spatial information.

The wild-stylized feature $z^{sw}_l$ is input into layer $l+1$ and $z^{sw}_{l+1} = \phi_{l+1}(z^{sw}_l)$ is output from the layer.
$z^{sw}_{l+1}$ can be swapped repeatedly in the style of ${z}^w_{l+1}$ as
\vspace{-0.5em}
\begin{equation}\label{eq:fs2}\vspace{-0.5em}
    z^{sw}_{l+1} := \sigma(z^w_{l+1}) \cfrac{z^{sw}_{l+1} - \mu(z^{sw}_{l+1})}{\sigma(z^{sw}_{l+1})} + \mu(z^w_{l+1}).
\end{equation}
By the above equation, feature $z^{sw}$ is swapped in the style of $z^{w}$ over multiple layers.
As the layer deepens, semantic information should be captured more important than style, so FS applies only to some shallow layers in this work.

\subsection{Content Extension Learning}\label{subsec:cel}
\vspace{-0.2em}
In this subsection, we propose to extend the contents in the source domain to the wild.
One of the reasons for overfitting to the source domain is that networks overlearn a limited amount of source content.
We address this issue by increasing the intra-class content variability with content extension in the latent embedding space.
To this end, we add a projection head $\phi_{proj}$ independently of the classification head $\phi_{cls}$ after the feature extractor $\phi_{feat}$ and extend the source contents to the wild in the embedding space.

\begin{figure}
  \centering
  \raisebox{-\height}{\includegraphics[width=\columnwidth]{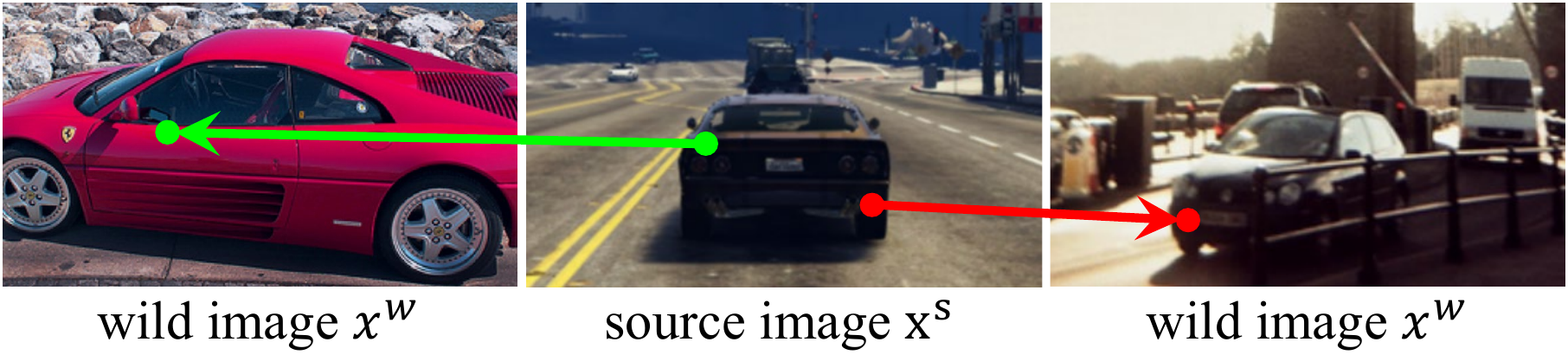}}
  \vspace{-0.2em}
  \caption{
  Visualization of source-to-wild matching pixels on the cropped source and wild images.
  We extend the source content to the wild content closest to the wild-stylized source content corresponding to the source pixel.
  The source content is encouraged to come close to the wild content in the embedding space.
  This improves the generalization ability of our model to unseen contents.
  }
  \vspace{-1.3em}
  \label{fig:cel_car}
\end{figure}

When a source image $x^s$ and wild image $x^w$ enter the feature extractor $\phi_{feat}$, it outputs the source feature $z^s$, wild-stylized source feature $z^{sw}$, and wild feature $z^w$.
The projection head $\phi_{proj}$ receives features $z^s$, $z^{sw}$, and $z^{w}$ and outputs pixel-level projected content features $z^s_{proj}$, $z^{sw}_{proj}$, and $z^{w}_{proj}$, respectively.
All projected features are normalized by $z = {z}/{max ( \left\lVert{z}\right\lVert_{2}, \epsilon ) }$ along the channel dimension.
At each training iteration, $z^{w}_{proj}$ is flattened and stored in the wild-content dictionary $Q \in {\mathbb{R}^{{C_q}\times{N_q}}}$ where $C_q$ is the number of channels of projected features and $N_q$ is the dictionary size.
Our model uses the dynamic dictionary structure in~\cite{he2020momentum} as $Q$ without a momentum update.
We diversify contents by extending the source contents to the wild-stylized source contents and then to the wild domain using $Q$.

Here we focus on that the projected source feature $z^{s}_{i}$\footnote{The subscript $proj$ is sometimes omitted for convenience.} and projected wild-stylized feature $z^{sw}_{i}$ corresponding to the $i$-th pixel of the source image $x^{s}$ contain exactly the same semantic information, but content perturbation exists.
In order to obtain reliable semantic information from unseen contents, networks should be able to cluster contents containing the same semantic information, distinguishing them from contents containing other semantic information.
To achieve this objective, we adapt the contrastive learning strategy~\cite{oord2018representation} to pixel-level instances in a supervised manner and define the source content extension loss for the $i$-th pixel as follows:
\vspace{-0.7em}
\begin{equation}
    \mathcal{L}^{i}_{SCE} = - { \text{log} \frac{ \psi(z^{s}_{i}, z^{sw}_{i}) }{ \psi(z^{s}_{i}, z^{sw}_{i}) + \sum\limits_{j=1}^{N_{z}} { \mathbbm{1}^{s}_{ij} \psi(z^{s}_{i}, z^{sw}_{j}) } } },
\label{eq:scei}
\end{equation}
\vspace{-0.8em}
\begin{equation}\vspace{-0.5em}
    \psi(z^{s}_{i}, z^{sw}_{i}) = \text{exp}( z^{s}_{i} \cdot z^{sw}_{i} / \tau ),
\end{equation}
where $\mathbbm{1}^{s}_{ij}$ is the negative pixel indicator that equals $1$ if $y^s_i$ and $y^s_j$ are different and $0$ if they are the same, $N_{z}$ is the number of pixels and the temperature parameter $\tau$ is set to 0.07.
We train the model only with reliable samples, ignoring ambiguous positive and negative samples by excluding pixels of unknown classes and pixels in other positions of the same class.
\cref{eq:scei} encourages $z^{s}_{i}$ and $z^{sw}_{i}$ to be close, while also encouraging $z^{s}_{i}$ to move away from all negative class contents. 
Then, the pixel-wise loss can be applied to the entire source image by
\vspace{-1.0em}
\begin{equation}\vspace{-0.5em}
    \mathcal{L}_{SCE} = \frac{1}{N_{z}} \sum\limits_{i=1}^{N_{z}} { \mathcal{L}^{i}_{SCE} }.
\label{eq:sce}
\end{equation}
\cref{eq:sce} encourages networks to make generalized predictions by reducing the distance in the embedding space between source contents and wild-perturbed source contents.

Next, we further extend the source contents to the wild by utilizing the wild-content dictionary $Q$.
In the learning process, $Q$ stores diverse pixel-level wild contents, which may not exist in the source domain.
Thus, if we carefully select wild-content with semantic information that each pixel needs to learn and then use it to train networks, networks become more robust to wild-content perturbations.
Since there is no class information in the wild set $D^w$, \cref{eq:scei} cannot be directly applied to this wild content extension.
We address this issue from the perspective that similar semantic contents will be located close to each other in the embedding space.
Inspired by~\cite{dwibedi2021little}, we take the wild content $z^{w}_{k}$ closest to the wild-stylized source content $z^{sw}_{i}$ from $Q$ as
\vspace{-0.7em}
\begin{equation}\vspace{-0.7em}
    z^{w}_{k} = \argmin_{q \in Q} \left\lVert{z^{sw}_{i} - q}\right\lVert_{2}
\label{eq:nn2}
\end{equation}
and encourage the source content $z^{s}_{i}$ to come close to it.
Since $z^{sw}_{i}$ and $q$ are normalized early on, \cref{eq:nn2} can be calculated efficiently using a dot product and rewritten as
\vspace{-0.7em}
\begin{equation}\vspace{-0.7em}
    z^{w}_{k} = \argmax_{q \in Q} ( z^{sw}_{i} \cdot q ).
\label{eq:nnd}
\end{equation}
\cref{fig:cel_car} shows the wild content $z^{w}_{k}$ matched to the source content $z^{s}_{i}$ using the stylized source content $z^{sw}_{i}$.
In this way, we provide various contents of the wild to the networks without category information.
Now \cref{eq:scei} can be adapted to the wild content extension as follows:
\vspace{-0.5em}
\begin{equation}\vspace{-0.5em}
    \mathcal{L}^{i}_{WCE} = - { \text{log} \frac{ \psi(z^{s}_{i}, z^{w}_{k}) }{ \psi(z^{s}_{i}, z^{w}_{k}) + \sum\limits_{j=1}^{N_{z}} { \mathbbm{1}^{s}_{ij} \psi(z^{s}_{i}, z^{sw}_{j}) } } }.
\label{eq:wcei}
\end{equation}
In the wild content extension, we reuse the negative samples of the source content extension.
Some negative extension approaches may give better performance and we left this for future work.
Then we can apply the pixel-wise wild content extension loss to the entire source image by
\vspace{-0.5em}
\begin{equation}\vspace{-0.5em}
    \mathcal{L}_{WCE} = \frac{1}{N_{z}} \sum\limits_{i=1}^{N_{z}} { \mathcal{L}^{i}_{WCE} }.
\label{eq:wce}
\end{equation}
By combining the source content extension loss and wild content extension loss, the CEL loss is defined as
\vspace{-0.5em}
\begin{equation}\vspace{-0.5em}
    \mathcal{L}_{CEL} = \mathcal{L}_{SCE} + \mathcal{L}_{WCE}.
\label{eq:cel}
\end{equation}
Our model learns to capture generalized semantic information from diverse contents by using the proposed CEL loss.

\begin{figure}
  \centering
  \begin{subfigure}{0.32\columnwidth}
    \raisebox{-\height}{\includegraphics[width=\columnwidth]{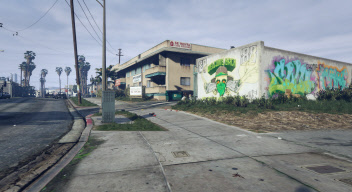}}
    \caption{$x^{s}$}
    \label{fig:fs_rec_pred_src_img}
  \end{subfigure}
  \hfill
  \begin{subfigure}{0.32\columnwidth}
    \raisebox{-\height}{\includegraphics[width=\columnwidth]{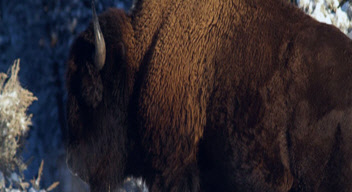}}
    \caption{$x^{w}$}
    \label{fig:fs_rec_pred_sup_img}
  \end{subfigure}
  \hfill
  \begin{subfigure}{0.32\columnwidth}
    \raisebox{-\height}{\includegraphics[width=\columnwidth]{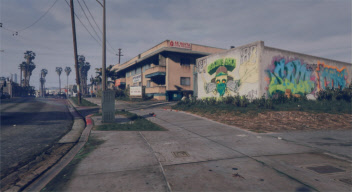}}
    \caption{rec. image from $z^{sw}$}
    \label{fig:fs_rec_pred_rec_img}
  \end{subfigure}
  \begin{subfigure}{0.32\columnwidth}
    \raisebox{-\height}{\includegraphics[width=\columnwidth]{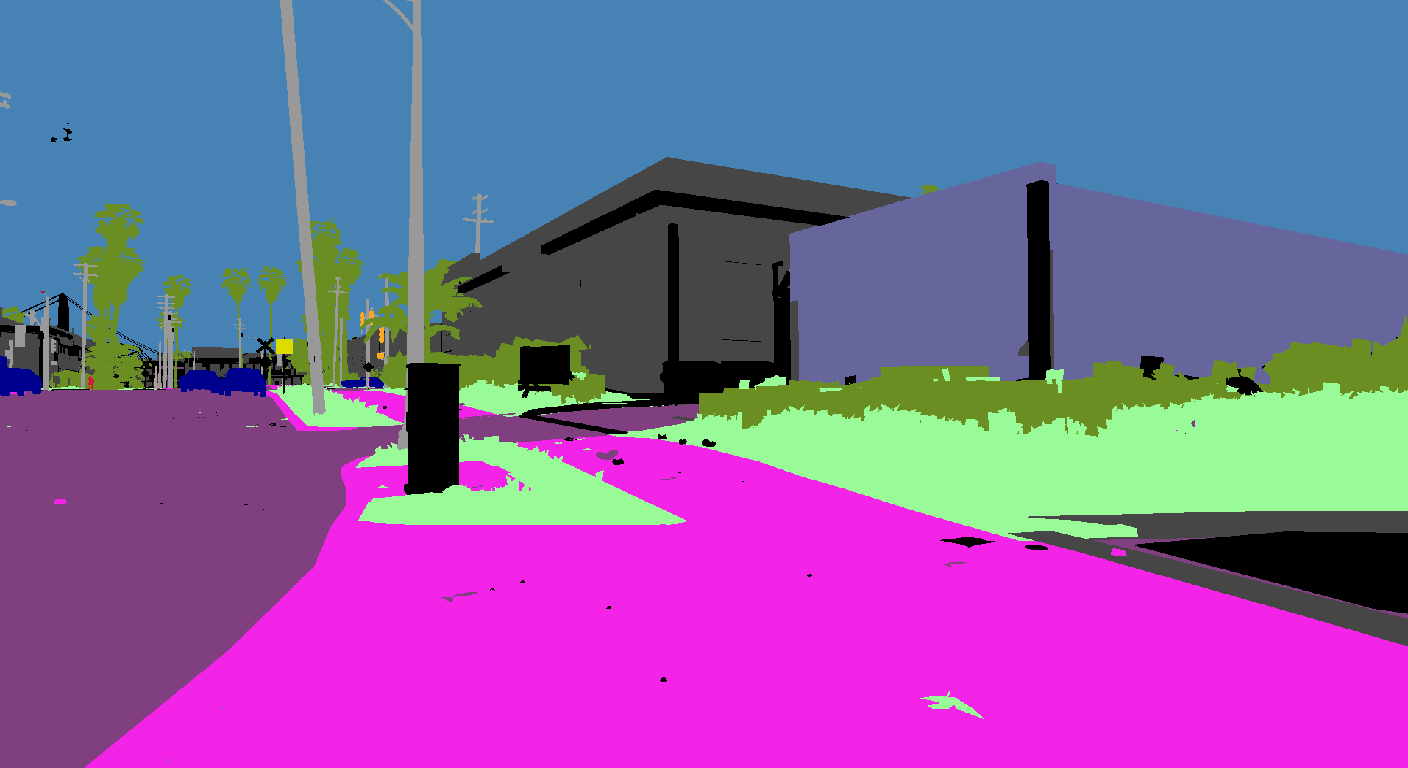}}
    \caption{$y^{s}$}
    \label{fig:fs_rec_pred_src_gt}
  \end{subfigure}
  \hfill
  \begin{subfigure}{0.32\columnwidth}
    \raisebox{-\height}{\includegraphics[width=\columnwidth]{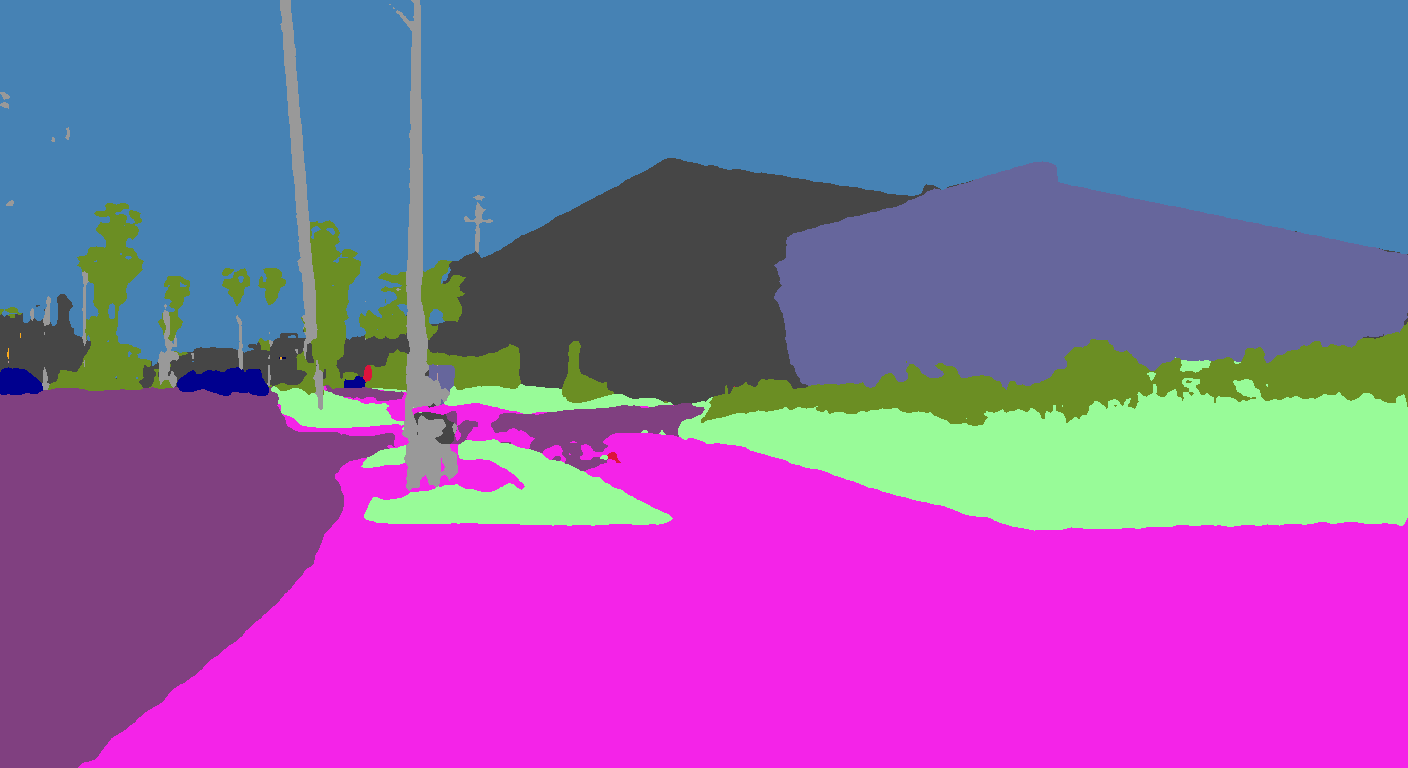}}
    \caption{prediction on $z^{s}$}
    \label{fig:fs_rec_pred_src_pred}
  \end{subfigure}
  \hfill
  \begin{subfigure}{0.32\columnwidth}
    \raisebox{-\height}{\includegraphics[width=\columnwidth]{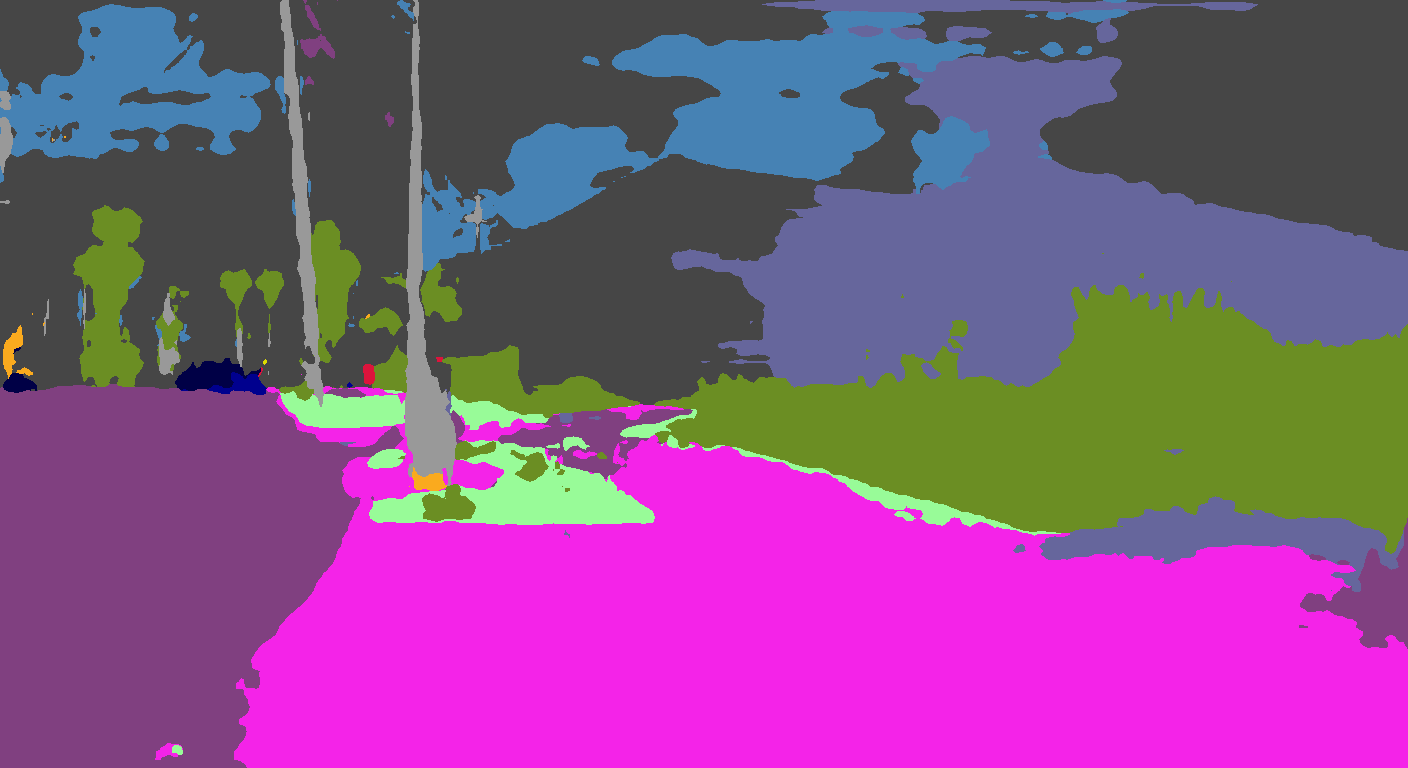}}
    \caption{prediction on $z^{sw}$}
    \label{fig:fs_rec_pred_rec_pred}
  \end{subfigure}
  \vspace{-0.2em}
  \caption{
  Given the (a) source image and (d) label, FS stylizes the source feature with the wild feature from the (b) wild image. 
  To visualize the wild-stylized feature, (c) we reconstructed an image from the wild-stylized feature using U-Net~\cite{ronneberger2015u}.
  Contrary to (e) accurate predictions from the source feature, networks (f) fail to make correct predictions from the wild-stylized feature even though the spatial information of the source feature remains the same.
  To address this issue, we apply SEL loss to allow networks to learn task-specific information from the wild-stylized features.
  }
  \vspace{-1.3em}
  \label{fig:fs_rec_pred}
\end{figure}

\subsection{Style Extension Learning}\label{subsec:sel}
\vspace{-0.2em}
Another reason for overfitting to the source domain is that networks overlearn a limited amount of the source style~\cite{zhou2021domain, nam2021reducing, park2020discover}.
To address this issue, FS has diversified the styles of the source feature with the help of the wild.
Interestingly, the style of the source features has changed while preserving spatial information, but networks fail to predict semantic categories from the wild-stylized feature as shown in~\cref{fig:fs_rec_pred_rec_pred}.
In this subsection, we propose SEL for adapting networks to diversified styles.
SEL aims to allow networks to naturally adapt to various styles by learning task-specific information from the wild-stylized feature.

When the wild-stylized source feature $z^{sw}$ enters the classification head $\phi_{cls}$, it outputs the pixel-wise softmax segmentation map $p^{sw} \in {\mathbb{R}^{{H}\times{W}\times{K}}}$.
Because $z^{sw}$ is the wild-stylized source feature in which the style of the feature from $x^{s}$ has been swapped with that of the feature from $x^w$, networks must predict the semantic label $y^{s}$ of $x^{s}$ from $z^{sw}$.
For this objective, we train networks by minimizing the following SEL loss:
\vspace{-0.7em}
\begin{equation}\vspace{-0.7em}
    \mathcal{L}_{SEL} = -\frac{1}{HW} \sum\limits_{h=1}^{H} \sum\limits_{w=1}^{W} \sum\limits_{k=1}^{K} { y^{s}_{hwk} \mathrm{log} (p^{sw}_{hwk}) }.
\label{eq:sel}
\end{equation}
Our model learns task-specific information from the wild-stylized features by applying the SEL loss.
This enables our model to naturally learn domain-generalized semantic information from various styles.

\subsection{Semantic Consistency Regularization}\label{subsec:scr}
\vspace{-0.2em}
For high generalization capability on unseen domains, the classifier should capture consistent semantic information from features~\cite{chen2019crdoco, isobe2021multi}, even if there are perturbations in both the style and content.
However, as shown in~\cref{fig:fs_rec_pred_src_pred,fig:fs_rec_pred_rec_pred}, the predicted result $p^{sw}$ of the wild-stylized source feature $z^{sw}$ differs from the predicted result $p^{s}$ of the source feature $z^{s}$.
Although SEL allows networks to learn task-specific information from $z^{sw}$, this does not guarantee that $p^{sw}$ and $p^{s}$ are identical.
To address this issue, we propose SCR that regularizes networks to capture consistent semantic information even when both the content and style of the source domain are extended to the wild.
SCR aims to train networks so that the predicted probability distributions $p^{sw}=\phi_{cls}(z^{sw})$ from the wild-stylized source features get closer to the $p^{s}=\phi_{cls}(z^{s})$ from the source features.
To this end, we adapt the Kullback-Leibler (KL) divergence loss as
\vspace{-0.7em}
\begin{equation}\vspace{-0.5em}
    \mathcal{L}_{SCR} = -\frac{1}{HW} \sum\limits_{h=1}^{H} \sum\limits_{w=1}^{W} \sum\limits_{k=1}^{K} { p^{s}_{hwk} \mathrm{log} \frac{p^{s}_{hwk}}{p^{sw}_{hwk}} }.
\label{eq:scr}
\end{equation}
With the SCR loss, our model learns consistent semantic information even with perturbations of style and content by the proposed wild extension methods.

\section{Experiments}\label{sec:experiments}

\subsection{Datasets}
\paragraph{Real semantic segmentation datasets.}
Cityscapes~\cite{cordts2016cityscapes}, BDD100K~\cite{yu2020bdd100k} and Mapillary~\cite{neuhold2017mapillary} consist of 2975, 7000, and 18000 images for train set and 500, 1000, and 2000 for validataion set. 
We consider 19 classes that are compatible with other datasets.
In all of the tables, \textbf{C}, \textbf{B}, and \textbf{M} denote Cityscapes, BDD100K, and Mapillary, respectively.

\paragraph{Synthetic semantic segmentation datasets.}
GTAV~\cite{richter2016playing} contains 24966 images rendered from the Grand Theft Auto V game engine.
It has 12403, 6382, and 6181 images for train, validation, and test sets, respectively.
SYNTHIA~\cite{ros2016synthia} contains 9400 images and we split it into 6580 and 2820 images for train and validation sets, following~\cite{choi2021robustnet}.
In all tables, \textbf{G} and \textbf{S} denote GTAV and SYNTHIA, respectively.

\paragraph{Wild dataset.}
ImageNet~\cite{deng2009imagenet} is a large-scale image classification dataset used for network pre-training in most studies.
In this paper, we use images without class labels as wild domain data.
The generalization performance according to the number of images used for training our WildNet is presented in~\Cref{tab:abl_num_wild}.

\subsection{Experimental Setup}
We conduct extensive experiments over five different semantic segmentation datasets and report the mean intersection over union (mIoU) score on several domain generalization scenarios: GTAV$\rightarrow$\{Cityscapes, BDD100K, Mapillary, SYNTHIA, GTAV\} and Cityscapes$\rightarrow$\{GTAV, BDD100K, Mapillary, SYNTHIA, Cityscapes\}.
For fair comparisons with other DG methods, we re-implement IBN-Net~\cite{pan2018two} and RobustNet~\cite{choi2021robustnet} on our baseline models and \textbf{$^\dag$} denotes our re-implemented models.
Our model is trained on one source domain train set (GTAV or Cityscapes) and validated on five domain validation sets (four unseen domains and one seen domain).
To show the overall domain generalization performance, we additionally report the average value of mIoU on five domain validation sets (\textbf{Avg}).
In all of the tables, the best results for each domain are marked in bold.

\subsection{Implementation Details}
We adapt ResNet-50, ResNet-101~\cite{he2016deep}, and VGG-16~\cite{simonyan2014very} with DeepLabV3+~\cite{chen2018encoder} as segmentation networks, and all backbones are pre-trained on ImageNet~\cite{deng2009imagenet}.
In the ResNet-based models, we use the SGD optimizer~\cite{robbins1951stochastic} with a momentum of 0.9 and weight decay of 5e-4. The initial learning rate is set to 2.5e-3 and is decreased using the polynomial policy with a power of 0.9. We train the models for 60K iterations with a batch size of 8.
In the VGG-based models, we use the Adam optimizer~\cite{kingma2015adam} with a momentum of (0.9, 0.99). The initial learning rate is set to 1e-5 and the batch size is set to 8.
Following~\cite{choi2021robustnet}, we apply random scaling within a range of [0.5, 2.0] and random cropping with a size of 768$\times$768.
The output size of the projection head is 192$\times$192 and we use uniformly sampled 64$\times$64 size feature maps for CEL to prevent memory issues.
For the diversity of the wild content dictionary, the wild feature maps are stored after uniform sampling with a size of 16$\times$16.
The FS layer replaces first batch normalization and is added immediately after the addition operation of the first two residual blocks in ResNet, and it added right after the first ReLU after the first three maxpool layers in VGG.
After training, all FS layers, projection head, and wild-content dictionary are removed, and our model can be applied to multiple unseen domains without further training on the target domains.

\begin{table}[]
    \centering
    \resizebox{0.95\columnwidth}{!}
    {
    \Large
    \begin{tabular}{l|cccc|c|c}
        \toprule
        Methods           & C              & B              & M              & S              & G              & Avg              \\
        \cmidrule{1-7}\morecmidrules \cmidrule{1-7}
        Baseline~\cite{pan2018two}   & 22.20          & -              & -              & -              & 61.00          & -              \\
        IBN-Net~\cite{pan2018two}    & 29.60          & -              & -              & -              & 64.20          & -              \\
        \cmidrule{1-7}
        Baseline~\cite{yue2019domain}   & 32.45          & 26.73          & 25.66          & -              & -              & -              \\
        DRPC~\cite{yue2019domain}       & 37.42          & 32.14          & 34.12          & -              & -              & -              \\
        \cmidrule{1-7}
        Baseline~\cite{chen2020automated}   & 23.29          & -              & -              & -              & -              & -              \\
        ASG~\cite{chen2020automated}        & 31.89          & -              & -              & -              & -              & -              \\
        \cmidrule{1-7}
        Baseline~\cite{choi2021robustnet}   & 28.95          & 25.14          & 28.18          & 26.23          & \textbf{73.45} & 36.39          \\
        RobustNet~\cite{choi2021robustnet}  & 36.58          & 35.20          & 40.33          & 28.30          & 72.10          & 42.50          \\
        \cmidrule{1-7}
        Baseline~\cite{peng2021global}   & 31.70          & -              & -              & -              & -              & -              \\
        GLTR~\cite{peng2021global}       & 38.60          & -              & -              & -              & -              & -              \\
        \cmidrule{1-7}
        Baseline          & 35.16          & 29.71          & 31.29          & 27.97          & 71.17          & 39.06              \\
        $^\dag$IBN-Net~\cite{pan2018two}   & 36.52          & 34.18          & 38.74          & 30.41          & 70.78          & 42.12              \\
        $^\dag$RobustNet~\cite{choi2021robustnet} & 38.78          & 35.64          & 40.38          & 28.97          & 70.16          & 42.78              \\
        WildNet (Ours)              & \textbf{44.62} & \textbf{38.42} & \textbf{46.09} & \textbf{31.34} & 71.20          & \textbf{46.33} \\
        \bottomrule
    \end{tabular}
    }
    \vspace{-0.6em}
    \caption{Comparison of mIoU(\%) using ResNet-50 as backbone under the domain generalization setting G$\rightarrow$\{C, B, M, S, G\}.}
    \vspace{-0.8em}
    \label{tab:gta_res50}
\end{table}

\begin{table}[]
    \centering
    \resizebox{0.95\columnwidth}{!}
    {
    \Large
    \begin{tabular}{l|cccc|c|c}
        \toprule
        Methods           & C              & B              & M              & S              & G              & Avg              \\
        \cmidrule{1-7}\morecmidrules \cmidrule{1-7}
        Baseline~\cite{yue2019domain}       & 33.56          & 27.76          & 28.33          & -              & -              & -              \\
        DRPC~\cite{yue2019domain}           & 42.53          & 38.72          & 38.05          & -              & -              & -              \\
        \cmidrule{1-7}
        Baseline~\cite{huang2021fsdr}       & 33.40          & 27.30          & 27.90          & -              & -              & -              \\
        FSDR~\cite{huang2021fsdr}           & 44.80          & 41.20          & 43.40          & -              & -              & -              \\
        \cmidrule{1-7}
        Baseline~\cite{peng2021global}      & 34.00          & 28.10          & 28.60          & -              & -              & -              \\
        GLTR~\cite{peng2021global}          & 43.70          & 39.60          & 39.10          & -              & -              & -              \\
        \cmidrule{1-7}
        Baseline                            & 35.73          & 34.06          & 33.42          & 29.06          & 71.79          & 40.81          \\
        $^\dag$IBN-Net~\cite{pan2018two}          & 37.68          & 36.64          & 36.75          & 30.84          & 70.39          & 42.46          \\
        $^\dag$RobustNet~\cite{choi2021robustnet} & 37.26          & 38.66          & 38.09          & 30.17          & 70.53          & 42.94          \\
        WildNet (Ours)                                            & \textbf{45.79} & \textbf{41.73} & \textbf{47.08} & \textbf{32.51} & \textbf{71.91} & \textbf{47.81}\\
        \bottomrule
    \end{tabular}
    }
    \vspace{-0.6em}
    \caption{Comparison of mIoU(\%) using ResNet-101 as backbone under the domain generalization setting G$\rightarrow$\{C, B, M, S, G\}.}
    \vspace{-1.0em}
    \label{tab:gta_res101}
\end{table}

\begin{table}[]
    \centering
    \resizebox{0.95\columnwidth}{!}
    {
    \Large
    \begin{tabular}{l|cccc|c|c}
        \toprule
        Methods           & C              & B              & M              & S              & G              & Avg              \\
        \cmidrule{1-7}\morecmidrules \cmidrule{1-7}
        Baseline~\cite{yue2019domain}     & 30.04          & 24.59          & 26.63          & -              & -              & -              \\
        DRPC~\cite{yue2019domain}          & 36.11          & 31.56          & 32.25          & -              & -              & -              \\
        \cmidrule{1-7}
        Baseline~\cite{chen2020automated}  & 19.89          & -              & -              & -              & -              & -              \\
        ASG~\cite{chen2020automated}       & 31.47          & -              & -              & -              & -              & -              \\
        \cmidrule{1-7}
        Baseline~\cite{huang2021fsdr}      & -              & -              & -              & -              & -              & -              \\
        FSDR~\cite{huang2021fsdr}          & 38.30          & 34.40          & 37.60          & -              & -              & -              \\
        \cmidrule{1-7}
        Baseline~\cite{peng2021global}     & 31.40          & -              & -              & -              & -              & -              \\
        GLTR~\cite{peng2021global}         & 37.20          & -              & -              & -              & -              & -              \\
        \cmidrule{1-7}
        Baseline                           & 24.68          & 26.41          & 23.60          & 24.73          & \textbf{66.36} & 33.16          \\
        $^\dag$IBN-Net~\cite{pan2018two}          & 30.25          & 30.09          & 31.87          & 26.22          & 65.47          & 36.78          \\
        $^\dag$RobustNet~\cite{choi2021robustnet} & 30.13          & 29.22          & 33.96          & 26.16          & 64.73          & 36.84          \\
        WildNet (Ours)                     & \textbf{39.18} & \textbf{34.49} & \textbf{40.75} & \textbf{27.25} & 64.57          & \textbf{41.25} \\
        \bottomrule
    \end{tabular}
    }
    \vspace{-0.6em}
    \caption{Comparison of mIoU(\%) using VGG-16 as backbone under the domain generalization setting G$\rightarrow$\{C, B, M, S, G\}.}
    \vspace{-0.8em}
    \label{tab:gta_vgg16}
\end{table}

\begin{table}[]
    \centering
    \resizebox{0.95\columnwidth}{!}
    {
    \Large
    \begin{tabular}{l|cccc|c|c}
        \toprule
        Methods           & G              & B              & M              & S              & C              & Avg              \\
        \cmidrule{1-7}\morecmidrules \cmidrule{1-7}
        Baseline~\cite{pan2018two}          & 29.40          & -              & -              & -              & 64.50          & -              \\
        IBN-Net~\cite{pan2018two}           & 37.90          & -              & -              & -              & 67.00          & -              \\
        \cmidrule{1-7}
        Baseline~\cite{choi2021robustnet}   & 42.55          & 44.96          & 51.68          & 23.29          & \textbf{77.51} & 48.00          \\
        RobustNet~\cite{choi2021robustnet}  & 45.00          & 50.73          & 58.64          & 26.20          & 76.41          & 51.40          \\
        \cmidrule{1-7}
        Baseline          & 40.50          & 42.35          & 20.67          & 8.08           & 76.30          & 37.58              \\
        $^\dag$IBN-Net~\cite{pan2018two}   & 45.28          & 46.61          & 56.78          & 26.41          & 75.47          & 50.11              \\
        $^\dag$RobustNet~\cite{choi2021robustnet} & 45.28          & 48.21          & 56.97          & 26.59          & 74.91          & 50.39              \\
        WildNet (Ours)              & \textbf{47.01} & \textbf{50.94} & \textbf{58.79} & \textbf{27.95} & 75.59          & \textbf{52.06} \\
        \bottomrule
    \end{tabular}
    }
    \vspace{-0.6em}
    \caption{Comparison of mIoU(\%) using ResNet-50 as backbone under the domain generalization setting C$\rightarrow$\{G, B, M, S, C\}.
    }
    \vspace{-0.8em}
    \label{tab:cty_res50}
\end{table}

\begin{table}[]
    \centering
    \resizebox{\columnwidth}{!}
    {
    \Huge
    \begin{tabular}{cccc|cccc|c|c}
        \toprule
        $\mathcal{L}_{orig}$      & $\mathcal{L}_{CEL}$      & $\mathcal{L}_{SEL}$      & $\mathcal{L}_{SCR}$      & C              & B              & M              & S              & G              & Avg            \\
        \cmidrule{1-10}\morecmidrules \cmidrule{1-10}
        \checkmark      &                &                &                & 35.16          & 29.71          & 31.29          & 27.97          & 71.17          & 39.06          \\
        \checkmark      & \checkmark     &                &                & 41.25          & 35.95          & 40.06          & 31.26          & 68.75          & 43.46          \\
        \checkmark      & \checkmark     & \checkmark     &                & 43.61          & \textbf{38.69} & 43.17          & \textbf{31.40} & 70.52          & 45.48          \\
        \checkmark      & \checkmark     & \checkmark     & \checkmark     & \textbf{44.62} & 38.42          & \textbf{46.09} & 31.34          & \textbf{71.20} & \textbf{46.33} \\
        \bottomrule
    \end{tabular}
    }
    \vspace{-0.6em}
    \caption{Effect of the proposed losses on the domain generalization setting G$\rightarrow$\{C,B,M,S,G\} using ResNet-50 as backbone in mIoU(\%). 
    Losses $\mathcal{L}_{orig}$, $\mathcal{L}_{CEL}$, $\mathcal{L}_{SEL}$, and $\mathcal{L}_{SCR}$ are defined in~\cref{eq:orig}, \cref{eq:cel}, \cref{eq:sel}, and \cref{eq:scr}, respectively.}
    \vspace{-0.8em}
    \label{tab:abl_loss}
\end{table}

\subsection{Comparison with DG methods}
We compare our results with existing DG methods: IBN-Net~\cite{pan2018two}, DRPC~\cite{yue2019domain}, ASG~\cite{chen2020automated}, FSDR~\cite{huang2021fsdr}, RobustNet~\cite{choi2021robustnet}, and GLTR~\cite{peng2021global}.
\Cref{tab:gta_res50} shows the generalization performance of the ResNet-50 model trained on GTAV.
We evaluate models on five validation sets consisting of four unseen domains, including the Cityscapes, BDD100K, Mapillary, and SYNTHIA datasets, and one seen domain of GTAV.
To demonstrate the high generalization ability over multiple domains, we also report the average value of the mIoU on the five domains.
Our WildNet shows remarkably superior generalization capabilities, significantly outperforming other methods in all unseen target domains except the source domain.
In particular, compared with the re-implemented results, we demonstrate that extending both the content and style is more effective in learning domain-generalized information than removing the domain-specific style.
Given in \Cref{tab:gta_res101,tab:gta_vgg16}, we achieve superior generalization ability with ResNet-101 and VGG-16 models.
Our model trained on Cityscapes also outperforms other DG methods as shown in \Cref{tab:cty_res50}.
Extensive comparative experiments of different backbones on various domains demonstrate the superiority of our model.

\begin{table*}[t]
\centering
\begin{subtable}{0.5\linewidth}
{
\centering
\resizebox{0.85\columnwidth}{!}
{
\large
\begin{tabular}{l|cccc|c|c}
    \toprule
    Num.     & C              & B              & M              & S              & G              & Avg            \\
    \cmidrule{1-7}\morecmidrules \cmidrule{1-7}
    Baseline & 35.16          & 29.71          & 31.29          & 27.97          & 71.17          & 39.06          \\
    10       & 42.43          & 36.82          & 42.15          & 30.66          & 70.92          & 44.60          \\
    100      & 43.29          & 37.71          & 43.93          & 30.67          & 70.93          & 45.31          \\
    1000     & 43.70          & 38.27          & 43.56          & 30.80          & 70.94          & 45.45          \\
    10000    & 43.87          & 37.98          & 44.19          & 31.04          & 70.85          & 45.59          \\
    All      & \textbf{44.62} & \textbf{38.42} & \textbf{46.09} & \textbf{31.34} & \textbf{71.20} & \textbf{46.33} \\
    \bottomrule
\end{tabular}
}
\caption{Number of wild images used in the training process.}
\label{tab:abl_num_wild}

\vspace{0.1cm}

\resizebox{0.85\columnwidth}{!}
{
\large
\begin{tabular}{l|cccc|c|c}
    \toprule
    \begin{tabular}[c]{@{}l@{}}Residual\\ Groups\end{tabular} & C              & B              & M              & S              & G              & Avg           \\
    \cmidrule{1-7}\morecmidrules \cmidrule{1-7}
    Baseline        & 35.16          & 29.71          & 31.29          & 27.97          & 71.17          & 39.06          \\
    1               & 43.09          & 35.28          & 41.36          & 30.51          & 71.19          & 44.29          \\
    1-2             & 43.43          & 36.90          & 41.34          & 30.36          & \textbf{71.33} & 44.67          \\
    1-3             & \textbf{44.62} & \textbf{38.42} & \textbf{46.09} & \textbf{31.34} & 71.20          & \textbf{46.33} \\
    1-4             & 44.03          & 37.83          & 43.39          & 30.24          & 70.45          & 45.19          \\
    1-5             & 43.49          & 34.91          & 43.26          & 30.00          & 70.05          & 44.34          \\
    \bottomrule
\end{tabular}
}
\caption{FS applied to different amounts of residual groups.}
\label{tab:abl_fs_layer}
}

\end{subtable}%
\begin{subtable}{0.5\linewidth}
{
\centering
\resizebox{0.85\columnwidth}{!}
{
\large
\begin{tabular}{l|cccc|c|c}
    \toprule
    Size       & C              & B              & M              & S              & G              & Avg           \\
    \cmidrule{1-7}\morecmidrules \cmidrule{1-7}
    None       & 43.85          & 38.64          & 42.70          & 28.62          & 71.07          & 44.98          \\
    2048       & 43.64          & 38.51          & 43.94          & 29.10          & 71.14          & 45.27          \\
    49152      & 43.81          & \textbf{39.01} & 44.97          & 29.41          & \textbf{71.44} & 45.73          \\
    393216     & \textbf{44.62} & 38.42          & \textbf{46.09} & \textbf{31.34} & 71.20          & \textbf{46.33} \\
    \bottomrule
\end{tabular}
}
\caption{Size of the wild content dictionary.}
\label{tab:abl_qs}

\vspace{0.25cm}

\resizebox{0.85\columnwidth}{!}
{
\large
\begin{tabular}{l|cccc|c|c}
    \toprule
    Methods & C              & B              & M              & S              & G              & Avg            \\
    \cmidrule{1-7}\morecmidrules \cmidrule{1-7}
    Baseline   & 35.16          & 29.71          & 31.29          & 27.97          & 71.17          & 39.06          \\
    Random     & 42.67          & 34.84          & 38.71          & 30.36          & 71.14          & 43.54          \\
    Wild       & \textbf{44.62} & \textbf{38.42} & \textbf{46.09} & \textbf{31.34} & \textbf{71.20} & \textbf{46.33} \\
    \bottomrule
\end{tabular}
}
\caption{Effect of FS with the statistics of the wild features.}
\label{tab:abl_fs_use_wild}

\vspace{0.25cm}

\resizebox{0.85\columnwidth}{!}
{
\large
\begin{tabular}{l|cccc|c|c}
    \toprule
    Methods & C              & B              & M              & S              & G              & Avg            \\
    \cmidrule{1-7}\morecmidrules \cmidrule{1-7}
    Random  & 43.83          & \textbf{38.75} & 45.10          & 30.65          & 71.10          & 45.89          \\
    Uniform & \textbf{44.62} & 38.42          & \textbf{46.09} & \textbf{31.34} & \textbf{71.20} & \textbf{46.33} \\
    \bottomrule
\end{tabular}
}
\caption{Comparison of sampling methods for CEL.}
\label{tab:abl_cel_sample}
}
\end{subtable}
\vspace{-0.6em}
\caption{\textbf{Ablation Study.} For each setting, we report mIoU(\%) using ResNet-50 as backbone in DG scenario: G$\rightarrow$\{C,B,M,S,G\}.}
\vspace{-1.2em}
\end{table*}

\subsection{Ablation Studies}
In this subsection, extensive experiments with ResNet-50 model on the DG scenario from GTAV to Cityscapes, BDD100K, Mapillary, SYNTHIA, and GTAV are conducted to study the effectiveness of each component in the proposed method.
\Cref{tab:abl_loss} shows the effect of the proposed losses on domain generalization.
The baseline model trained only with $\mathcal{L}_{orig}$ overfits the source domain and has poor performance on unseen domains.
Even with only $\mathcal{L}_{CEL}$ applied, our model achieves an Avg of 43.46\% with +4.40\% improvement.
This shows the importance of content diversification that is overlooked in many studies.
Further, we make the wild-stylized features learn task-specific information with $\mathcal{L}_{SEL}$ to achieve an Avg of 45.48\%, and regularize the model to learn consistent semantic information with $\mathcal{L}_{SCR}$, finally achieving an Avg of 46.33\%.
Next, we conduct more ablations for important components.

\paragraph{Number of wild images.}
In~\Cref{tab:abl_num_wild}, the number of wild images used to train our model is considered.
Even if only 10 wild images are used, the generalization performance is significantly enhanced by +5.54\% compared with the baseline by preventing overfitting to the source domain.
Moreover, the generalization performance of the model gradually improves as the number of wild images used increases.
This shows that the extension of both content and style to the wild helps networks to learn domain-generalized semantic features.

\paragraph{Amount of FS.}
\Cref{tab:abl_fs_layer} shows the influence of the amount of FS on generalization performance.
By replacing only the first batch normalization with FS, we can extend contents and styles to the wild based on diversified stylized features and improve generalization performance compared to baseline by +5.23\%.
Adding FS to some shallow layers boosts performance further.
However, applying FS to deeper layers degrades performance slightly, as semantic content should be captured more important than style as the layer deepens.
A suitable amount of FS, which does not disturb the semantic information, helps to train the generalized model by augmenting the source features to have various wild styles.

\paragraph{Size of wild content dictionary.}
\Cref{tab:abl_qs} shows sensitivity to the size of the wild content dictionary.
Extending the source content to the wild improves generalization performance, and even when extended to wild content within a mini-batch of size 2048 without a content dictionary, our model achieves higher generalization performance than without content extension.
We take size of 393216.

\paragraph{FS with wild style.}
In~\Cref{tab:abl_fs_use_wild}, we show the effect of FS using statistics of wild features on the generalization performance of the model.
To apply FS without the help of the wild, the mean and standard deviation of the source features were multiplied by random values in the range [0.5, 1.5] and then used instead of the statistics of the wild features.
The random FS improves performance compared to baseline by +4.48\%, which shows the importance of diversifying styles.
Furthermore, wild FS demonstrates that learning a natural style of the wild is much better with a gain of +7.27\%.

\paragraph{Sampling methods.}
By using sampled feature maps, CEL stores various wild contents in the fixed-size dictionary and reduces memory consumption due to pixel-level contrastive loss calculations.
Since two adjacent pixels have almost similar semantic information, uniform sampling makes learning more diverse contents than random sampling, leading to high generalization performance as shown in~\Cref{tab:abl_cel_sample}.

\begin{figure}[t]
\centering
\includegraphics[width=0.47\textwidth]{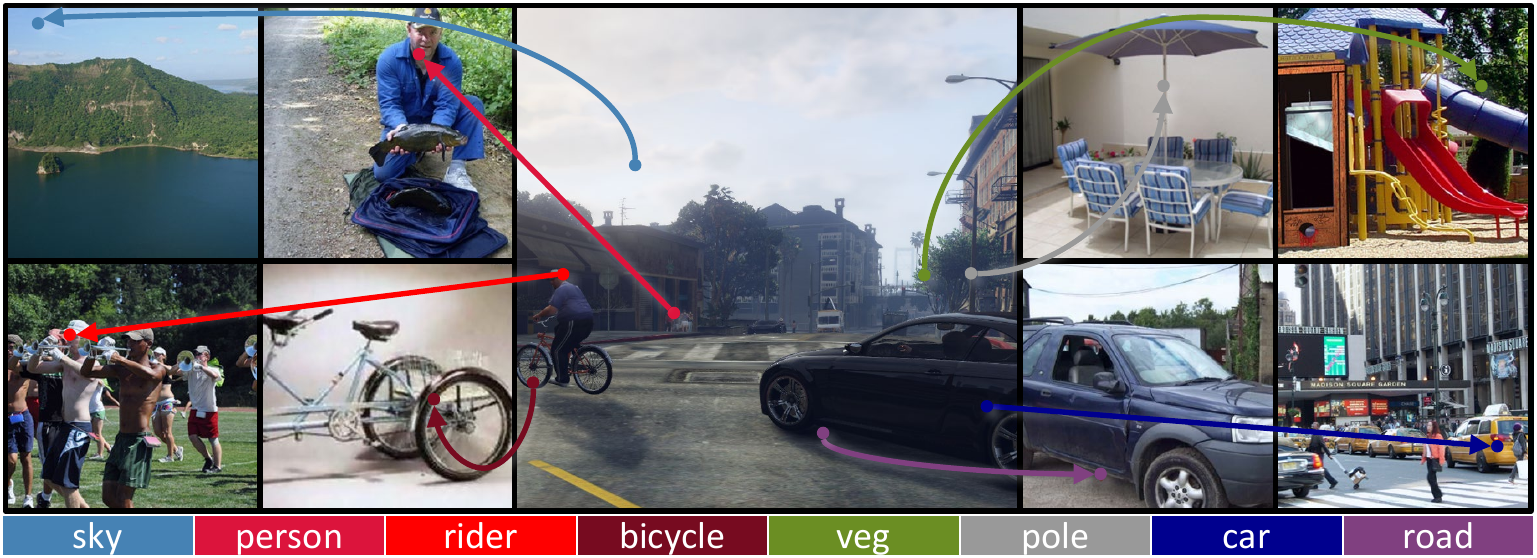}
\vspace{-0.6em}
\caption{
Visualization of extended wild contents.
More visualizations are available in the supplementary material.
}
\vspace{-0.8em}
\label{fig:cel_discussion}
\end{figure}

\section{Discussion}
\vspace{-0.1em}
\subsection{Qualitative Analysis}\label{subsec:analysis}
\vspace{-0.1em}
To analyze the wild content extension, we visualize the wild contents closest to the stylized source contents in~\cref{fig:cel_discussion}.
As can be seen in the figure, the source content is extended to wild content with semantic information similar to itself, \eg, the road under the car, the wheel of the bicycle, and the head of a man wearing a hat.
Our model learns domain-generalized features by inducing source content closer to these wild content in the feature space.
With learning various wild contents, WildNet makes reliable predictions on unseen contents.
\cref{fig:compare_r50_city_img_base_isw_ours,fig:fs_rec_pred} show segmentation results and visualization of wild-stylized features, and further analysis is provided in the supplementary material.

\subsection{Limitations and Future Works}
\vspace{-0.1em}
We have shown that the source content extends to wild content with semantic information similar to itself.
However, similar semantic information in the two contents does not guarantee that the classes of the two contents are always the same, as observed for the rider of~\cref{fig:cel_discussion}.
Extending the rider with the hat to the person with the hat may bridge between the rider-class and person-class.
Our future works will involve positive content selection using predicted class probabilities on wild images and negative content extension to further boost the class discrimination ability.

\section{Conclusion}
\vspace{-0.1em}
We presented WildNet which exploits unlabeled wild images for domain-generalized semantic segmentation.
Our approach effectively extends style and content from source to wild, resulting in drastic performance improvement even we leverage 10 wild images.
In contrast to previous studies that exploit generalization cues only from style, we additionally exploit the potential to generalize domain from content.
We thoroughly ablated to demonstrate the efficacy of our WildNet and achieved superior segmentation performance under several domain generalization scenarios.
We believe that our approach provides an opportunity to utilize huge amounts of unlabeled data for domain generalization.

\paragraph{Acknowledgement.}~This research was supported by the National Research Foundation of Korea (NRF) grant funded by the Korea government (MSIT) (NRF-2019R1A2C1007153).

{\small
\bibliographystyle{ieee_fullname}
\bibliography{main}

\begin{thebibliography}{10}\itemsep=-1pt

\bibitem{alonso2021semi}
Inigo Alonso, Alberto Sabater, David Ferstl, Luis Montesano, and Ana~C Murillo.
\newblock Semi-supervised semantic segmentation with pixel-level contrastive
  learning from a class-wise memory bank.
\newblock In {\em ICCV}, 2021.

\bibitem{chen2018encoder}
Liang-Chieh Chen, Yukun Zhu, George Papandreou, Florian Schroff, and Hartwig
  Adam.
\newblock Encoder-decoder with atrous separable convolution for semantic image
  segmentation.
\newblock In {\em ECCV}, pages 801--818, 2018.

\bibitem{chen2020simple}
Ting Chen, Simon Kornblith, Mohammad Norouzi, and Geoffrey Hinton.
\newblock A simple framework for contrastive learning of visual
  representations.
\newblock In {\em ICML}, pages 1597--1607. PMLR, 2020.

\bibitem{chen2020automated}
Wuyang Chen, Zhiding Yu, Zhangyang Wang, and Animashree Anandkumar.
\newblock Automated synthetic-to-real generalization.
\newblock In {\em ICML}, pages 1746--1756. PMLR, 2020.

\bibitem{chen2019crdoco}
Yun-Chun Chen, Yen-Yu Lin, Ming-Hsuan Yang, and Jia-Bin Huang.
\newblock Crdoco: Pixel-level domain transfer with cross-domain consistency.
\newblock In {\em CVPR}, pages 1791--1800, 2019.

\bibitem{cheng2020panoptic}
Bowen Cheng, Maxwell~D Collins, Yukun Zhu, Ting Liu, Thomas~S Huang, Hartwig
  Adam, and Liang-Chieh Chen.
\newblock Panoptic-deeplab: A simple, strong, and fast baseline for bottom-up
  panoptic segmentation.
\newblock In {\em CVPR}, pages 12475--12485, 2020.

\bibitem{choi2021robustnet}
Sungha Choi, Sanghun Jung, Huiwon Yun, Joanne~T Kim, Seungryong Kim, and Jaegul
  Choo.
\newblock Robustnet: Improving domain generalization in urban-scene
  segmentation via instance selective whitening.
\newblock In {\em CVPR}, pages 11580--11590, 2021.

\bibitem{choi2020cars}
Sungha Choi, Joanne~T Kim, and Jaegul Choo.
\newblock Cars can't fly up in the sky: Improving urban-scene segmentation via
  height-driven attention networks.
\newblock In {\em CVPR}, pages 9373--9383, 2020.

\bibitem{chopra2005learning}
Sumit Chopra, Raia Hadsell, and Yann LeCun.
\newblock Learning a similarity metric discriminatively, with application to
  face verification.
\newblock In {\em CVPR}, volume~1, pages 539--546. IEEE, 2005.

\bibitem{cordts2016cityscapes}
Marius Cordts, Mohamed Omran, Sebastian Ramos, Timo Rehfeld, Markus Enzweiler,
  Rodrigo Benenson, Uwe Franke, Stefan Roth, and Bernt Schiele.
\newblock The cityscapes dataset for semantic urban scene understanding.
\newblock In {\em CVPR}, pages 3213--3223, 2016.

\bibitem{deng2009imagenet}
Jia Deng, Wei Dong, Richard Socher, Li-Jia Li, Kai Li, and Li Fei-Fei.
\newblock Imagenet: A large-scale hierarchical image database.
\newblock In {\em CVPR}, pages 248--255. IEEE, 2009.

\bibitem{dwibedi2021little}
Debidatta Dwibedi, Yusuf Aytar, Jonathan Tompson, Pierre Sermanet, and Andrew
  Zisserman.
\newblock With a little help from my friends: Nearest-neighbor contrastive
  learning of visual representations.
\newblock In {\em ICCV}, 2021.

\bibitem{gatys2015texture}
Leon Gatys, Alexander~S Ecker, and Matthias Bethge.
\newblock Texture synthesis using convolutional neural networks.
\newblock {\em NeurIPS}, 28:262--270, 2015.

\bibitem{gatys2016image}
Leon~A Gatys, Alexander~S Ecker, and Matthias Bethge.
\newblock Image style transfer using convolutional neural networks.
\newblock In {\em CVPR}, pages 2414--2423, 2016.

\bibitem{gatys2017controlling}
Leon~A Gatys, Alexander~S Ecker, Matthias Bethge, Aaron Hertzmann, and Eli
  Shechtman.
\newblock Controlling perceptual factors in neural style transfer.
\newblock In {\em CVPR}, pages 3985--3993, 2017.

\bibitem{guo2021metacorrection}
Xiaoqing Guo, Chen Yang, Baopu Li, and Yixuan Yuan.
\newblock Metacorrection: Domain-aware meta loss correction for unsupervised
  domain adaptation in semantic segmentation.
\newblock In {\em CVPR}, pages 3927--3936, 2021.

\bibitem{he2021multi}
Jianzhong He, Xu Jia, Shuaijun Chen, and Jianzhuang Liu.
\newblock Multi-source domain adaptation with collaborative learning for
  semantic segmentation.
\newblock In {\em CVPR}, pages 11008--11017, 2021.

\bibitem{he2020momentum}
Kaiming He, Haoqi Fan, Yuxin Wu, Saining Xie, and Ross Girshick.
\newblock Momentum contrast for unsupervised visual representation learning.
\newblock In {\em CVPR}, pages 9729--9738, 2020.

\bibitem{he2016deep}
Kaiming He, Xiangyu Zhang, Shaoqing Ren, and Jian Sun.
\newblock Deep residual learning for image recognition.
\newblock In {\em CVPR}, pages 770--778, 2016.

\bibitem{hoffman2017cycada}
Judy Hoffman, Eric Tzeng, Taesung Park, Jun-Yan Zhu, Phillip Isola, Kate
  Saenko, Alexei~A Efros, and Trevor Darrell.
\newblock Cycada: Cycle-consistent adversarial domain adaptation.
\newblock In {\em ICML}, 2018.

\bibitem{hoffman2016fcns}
Judy Hoffman, Dequan Wang, Fisher Yu, and Trevor Darrell.
\newblock Fcns in the wild: Pixel-level adversarial and constraint-based
  adaptation.
\newblock {\em arXiv preprint arXiv:1612.02649}, 2016.

\bibitem{huang2021fsdr}
Jiaxing Huang, Dayan Guan, Aoran Xiao, and Shijian Lu.
\newblock Fsdr: Frequency space domain randomization for domain generalization.
\newblock In {\em CVPR}, pages 6891--6902, 2021.

\bibitem{huang2018decorrelated}
Lei Huang, Dawei Yang, Bo Lang, and Jia Deng.
\newblock Decorrelated batch normalization.
\newblock In {\em CVPR}, pages 791--800, 2018.

\bibitem{huang2017arbitrary}
Xun Huang and Serge Belongie.
\newblock Arbitrary style transfer in real-time with adaptive instance
  normalization.
\newblock In {\em ICCV}, pages 1501--1510, 2017.

\bibitem{isobe2021multi}
Takashi Isobe, Xu Jia, Shuaijun Chen, Jianzhong He, Yongjie Shi, Jianzhuang
  Liu, Huchuan Lu, and Shengjin Wang.
\newblock Multi-target domain adaptation with collaborative consistency
  learning.
\newblock In {\em CVPR}, pages 8187--8196, 2021.

\bibitem{jing2019neural}
Yongcheng Jing, Yezhou Yang, Zunlei Feng, Jingwen Ye, Yizhou Yu, and Mingli
  Song.
\newblock Neural style transfer: A review.
\newblock {\em IEEE Transactions on Visualization and Computer Graphics},
  26(11):3365--3385, 2019.

\bibitem{khosla2020supervised}
Prannay Khosla, Piotr Teterwak, Chen Wang, Aaron Sarna, Yonglong Tian, Phillip
  Isola, Aaron Maschinot, Ce Liu, and Dilip Krishnan.
\newblock Supervised contrastive learning.
\newblock In {\em NeurIPS}, volume~33, 2020.

\bibitem{kingma2015adam}
Diederik~P Kingma and Jimmy Ba.
\newblock Adam: A method for stochastic optimization.
\newblock In {\em ICLR}, 2015.

\bibitem{lee2021unsupervised}
Suhyeon Lee, Junhyuk Hyun, Hongje Seong, and Euntai Kim.
\newblock Unsupervised domain adaptation for semantic segmentation by content
  transfer.
\newblock In {\em AAAI}, pages 8306--8315, 2021.

\bibitem{li2017universal}
Yijun Li, Chen Fang, Jimei Yang, Zhaowen Wang, Xin Lu, and Ming-Hsuan Yang.
\newblock Universal style transfer via feature transforms.
\newblock In {\em NeurIPS}, pages 385--395, 2017.

\bibitem{li2017demystifying}
Yanghao Li, Naiyan Wang, Jiaying Liu, and Xiaodi Hou.
\newblock Demystifying neural style transfer.
\newblock In {\em IJCAI}, pages 2230--2236, 2017.

\bibitem{li2019bidirectional}
Yunsheng Li, Lu Yuan, and Nuno Vasconcelos.
\newblock Bidirectional learning for domain adaptation of semantic
  segmentation.
\newblock In {\em CVPR}, pages 6936--6945, 2019.

\bibitem{ma2021coarse}
Haoyu Ma, Xiangru Lin, Zifeng Wu, and Yizhou Yu.
\newblock Coarse-to-fine domain adaptive semantic segmentation with photometric
  alignment and category-center regularization.
\newblock In {\em CVPR}, pages 4051--4060, 2021.

\bibitem{ma2018shufflenet}
Ningning Ma, Xiangyu Zhang, Hai-Tao Zheng, and Jian Sun.
\newblock Shufflenet v2: Practical guidelines for efficient cnn architecture
  design.
\newblock In {\em ECCV}, pages 116--131, 2018.

\bibitem{mohan2021efficientps}
Rohit Mohan and Abhinav Valada.
\newblock Efficientps: Efficient panoptic segmentation.
\newblock {\em International Journal of Computer Vision}, 129(5):1551--1579,
  2021.

\bibitem{nam2021reducing}
Hyeonseob Nam, HyunJae Lee, Jongchan Park, Wonjun Yoon, and Donggeun Yoo.
\newblock Reducing domain gap by reducing style bias.
\newblock In {\em CVPR}, pages 8690--8699, 2021.

\bibitem{neuhold2017mapillary}
Gerhard Neuhold, Tobias Ollmann, Samuel Rota~Bulo, and Peter Kontschieder.
\newblock The mapillary vistas dataset for semantic understanding of street
  scenes.
\newblock In {\em ICCV}, pages 4990--4999, 2017.

\bibitem{nichol2016painter}
Kiri Nichol.
\newblock Painter by numbers, wikiart.
\newblock \url{https://www.kaggle.com/c/painter-by-numbers}, 2016.

\bibitem{oord2018representation}
Aaron van~den Oord, Yazhe Li, and Oriol Vinyals.
\newblock Representation learning with contrastive predictive coding.
\newblock {\em arXiv preprint arXiv:1807.03748}, 2018.

\bibitem{pan2020unsupervised}
Fei Pan, Inkyu Shin, Francois Rameau, Seokju Lee, and In~So Kweon.
\newblock Unsupervised intra-domain adaptation for semantic segmentation
  through self-supervision.
\newblock In {\em CVPR}, pages 3764--3773, 2020.

\bibitem{pan2018two}
Xingang Pan, Ping Luo, Jianping Shi, and Xiaoou Tang.
\newblock Two at once: Enhancing learning and generalization capacities via
  ibn-net.
\newblock In {\em ECCV}, pages 464--479, 2018.

\bibitem{pan2019switchable}
Xingang Pan, Xiaohang Zhan, Jianping Shi, Xiaoou Tang, and Ping Luo.
\newblock Switchable whitening for deep representation learning.
\newblock In {\em ICCV}, pages 1863--1871, 2019.

\bibitem{park2020discover}
Kwanyong Park, Sanghyun Woo, Inkyu Shin, and In~So Kweon.
\newblock Discover, hallucinate, and adapt: Open compound domain adaptation for
  semantic segmentation.
\newblock In {\em NeurIPS}, volume~33, pages 10869--10880, 2020.

\bibitem{paul2020domain}
Sujoy Paul, Yi-Hsuan Tsai, Samuel Schulter, Amit~K Roy-Chowdhury, and Manmohan
  Chandraker.
\newblock Domain adaptive semantic segmentation using weak labels.
\newblock In {\em ECCV}, pages 571--587. Springer, 2020.

\bibitem{peng2021global}
Duo Peng, Yinjie Lei, Lingqiao Liu, Pingping Zhang, and Jun Liu.
\newblock Global and local texture randomization for synthetic-to-real semantic
  segmentation.
\newblock {\em IEEE Transactions on Image Processing}, 30:6594--6608, 2021.

\bibitem{richter2016playing}
Stephan~R Richter, Vibhav Vineet, Stefan Roth, and Vladlen Koltun.
\newblock Playing for data: Ground truth from computer games.
\newblock In {\em ECCV}, pages 102--118. Springer, 2016.

\bibitem{robbins1951stochastic}
Herbert Robbins and Sutton Monro.
\newblock A stochastic approximation method.
\newblock {\em The annals of mathematical statistics}, pages 400--407, 1951.

\bibitem{ronneberger2015u}
Olaf Ronneberger, Philipp Fischer, and Thomas Brox.
\newblock U-net: Convolutional networks for biomedical image segmentation.
\newblock In {\em International Conference on Medical Image Computing and
  Computer-Assisted Intervention}, pages 234--241. Springer, 2015.

\bibitem{ros2016synthia}
German Ros, Laura Sellart, Joanna Materzynska, David Vazquez, and Antonio~M
  Lopez.
\newblock The synthia dataset: A large collection of synthetic images for
  semantic segmentation of urban scenes.
\newblock In {\em CVPR}, pages 3234--3243, 2016.

\bibitem{roy2019unsupervised}
Subhankar Roy, Aliaksandr Siarohin, Enver Sangineto, Samuel~Rota Bulo, Nicu
  Sebe, and Elisa Ricci.
\newblock Unsupervised domain adaptation using feature-whitening and consensus
  loss.
\newblock In {\em CVPR}, pages 9471--9480, 2019.

\bibitem{seong2021hierarchical}
Hongje Seong, Seoung~Wug Oh, Joon-Young Lee, Seongwon Lee, Suhyeon Lee, and
  Euntai Kim.
\newblock Hierarchical memory matching network for video object segmentation.
\newblock In {\em ICCV}, pages 12889--12898, 2021.

\bibitem{simonyan2014very}
Karen Simonyan and Andrew Zisserman.
\newblock Very deep convolutional networks for large-scale image recognition.
\newblock In {\em ICLR}, 2015.

\bibitem{sun2016deep}
Baochen Sun and Kate Saenko.
\newblock Deep coral: Correlation alignment for deep domain adaptation.
\newblock In {\em ECCV}, pages 443--450. Springer, 2016.

\bibitem{tao2020hierarchical}
Andrew Tao, Karan Sapra, and Bryan Catanzaro.
\newblock Hierarchical multi-scale attention for semantic segmentation.
\newblock {\em arXiv preprint arXiv:2005.10821}, 2020.

\bibitem{tsai2018learning}
Yi-Hsuan Tsai, Wei-Chih Hung, Samuel Schulter, Kihyuk Sohn, Ming-Hsuan Yang,
  and Manmohan Chandraker.
\newblock Learning to adapt structured output space for semantic segmentation.
\newblock In {\em CVPR}, pages 7472--7481, 2018.

\bibitem{vu2019advent}
Tuan-Hung Vu, Himalaya Jain, Maxime Bucher, Matthieu Cord, and Patrick
  P{\'e}rez.
\newblock Advent: Adversarial entropy minimization for domain adaptation in
  semantic segmentation.
\newblock In {\em CVPR}, pages 2517--2526, 2019.

\bibitem{wang2020classes}
Haoran Wang, Tong Shen, Wei Zhang, Ling-Yu Duan, and Tao Mei.
\newblock Classes matter: A fine-grained adversarial approach to cross-domain
  semantic segmentation.
\newblock In {\em ECCV}, pages 642--659. Springer, 2020.

\bibitem{wu2018unsupervised}
Zhirong Wu, Yuanjun Xiong, Stella~X Yu, and Dahua Lin.
\newblock Unsupervised feature learning via non-parametric instance
  discrimination.
\newblock In {\em CVPR}, pages 3733--3742, 2018.

\bibitem{xiong2019upsnet}
Yuwen Xiong, Renjie Liao, Hengshuang Zhao, Rui Hu, Min Bai, Ersin Yumer, and
  Raquel Urtasun.
\newblock Upsnet: A unified panoptic segmentation network.
\newblock In {\em CVPR}, pages 8818--8826, 2019.

\bibitem{yang2020fda}
Yanchao Yang and Stefano Soatto.
\newblock Fda: Fourier domain adaptation for semantic segmentation.
\newblock In {\em CVPR}, pages 4085--4095, 2020.

\bibitem{yu2020bdd100k}
Fisher Yu, Haofeng Chen, Xin Wang, Wenqi Xian, Yingying Chen, Fangchen Liu,
  Vashisht Madhavan, and Trevor Darrell.
\newblock Bdd100k: A diverse driving dataset for heterogeneous multitask
  learning.
\newblock In {\em CVPR}, pages 2636--2645, 2020.

\bibitem{yue2019domain}
Xiangyu Yue, Yang Zhang, Sicheng Zhao, Alberto Sangiovanni-Vincentelli, Kurt
  Keutzer, and Boqing Gong.
\newblock Domain randomization and pyramid consistency: Simulation-to-real
  generalization without accessing target domain data.
\newblock In {\em ICCV}, pages 2100--2110, 2019.

\bibitem{zhang2021prototypical}
Pan Zhang, Bo Zhang, Ting Zhang, Dong Chen, Yong Wang, and Fang Wen.
\newblock Prototypical pseudo label denoising and target structure learning for
  domain adaptive semantic segmentation.
\newblock In {\em CVPR}, pages 12414--12424, 2021.

\bibitem{zhang2018fully}
Yiheng Zhang, Zhaofan Qiu, Ting Yao, Dong Liu, and Tao Mei.
\newblock Fully convolutional adaptation networks for semantic segmentation.
\newblock In {\em CVPR}, pages 6810--6818, 2018.

\bibitem{zhao2021contrastive}
Xiangyun Zhao, Raviteja Vemulapalli, Philip~Andrew Mansfield, Boqing Gong,
  Bradley Green, Lior Shapira, and Ying Wu.
\newblock Contrastive learning for label efficient semantic segmentation.
\newblock In {\em ICCV}, pages 10623--10633, 2021.

\bibitem{zhong2021pixel}
Yuanyi Zhong, Bodi Yuan, Hong Wu, Zhiqiang Yuan, Jian Peng, and Yu-Xiong Wang.
\newblock Pixel contrastive-consistent semi-supervised semantic segmentation.
\newblock In {\em ICCV}, pages 7273--7282, 2021.

\bibitem{zhou2021domain}
Kaiyang Zhou, Yongxin Yang, Yu Qiao, and Tao Xiang.
\newblock Domain generalization with mixstyle.
\newblock In {\em ICLR}, 2021.

\bibitem{zhu2020deformable}
Xizhou Zhu, Weijie Su, Lewei Lu, Bin Li, Xiaogang Wang, and Jifeng Dai.
\newblock Deformable detr: Deformable transformers for end-to-end object
  detection.
\newblock In {\em ICLR}, 2020.

\end{thebibliography}
}

\clearpage

\appendix

\section{More Analysis}
\vspace{-0.2em}
In this section, we further analyze our method with additional qualitative results.
We also provide semantic segmentation results on five different datasets, which consist of four unseen domain datasets and one seen domain dataset.

\vspace{-0.2em}
\subsection{Content Extension Learning}
\begin{figure*}[t]
  \centering
  \begin{subfigure}{1.0\textwidth}
    \raisebox{-\height}{\includegraphics[width=\textwidth]{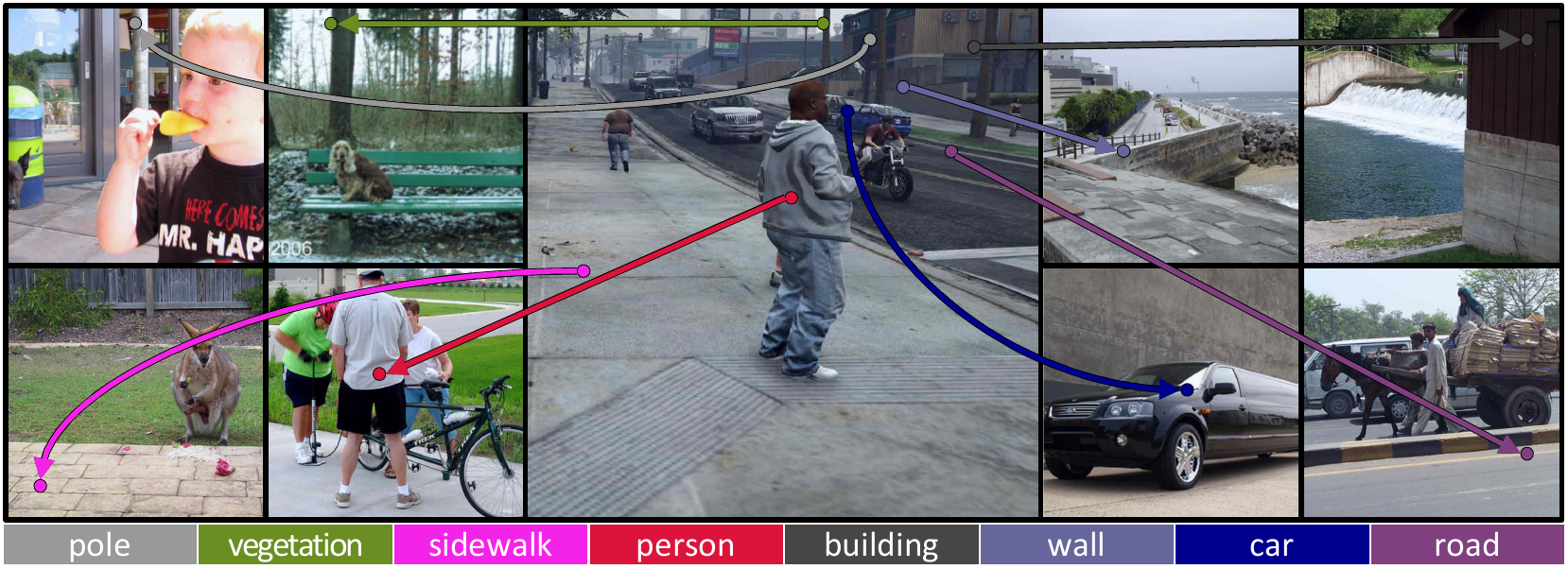}}
    \caption{}
    \label{fig:cel_supp1}
  \end{subfigure}
  \begin{subfigure}{1.0\textwidth}
    \raisebox{-\height}{\includegraphics[width=\textwidth]{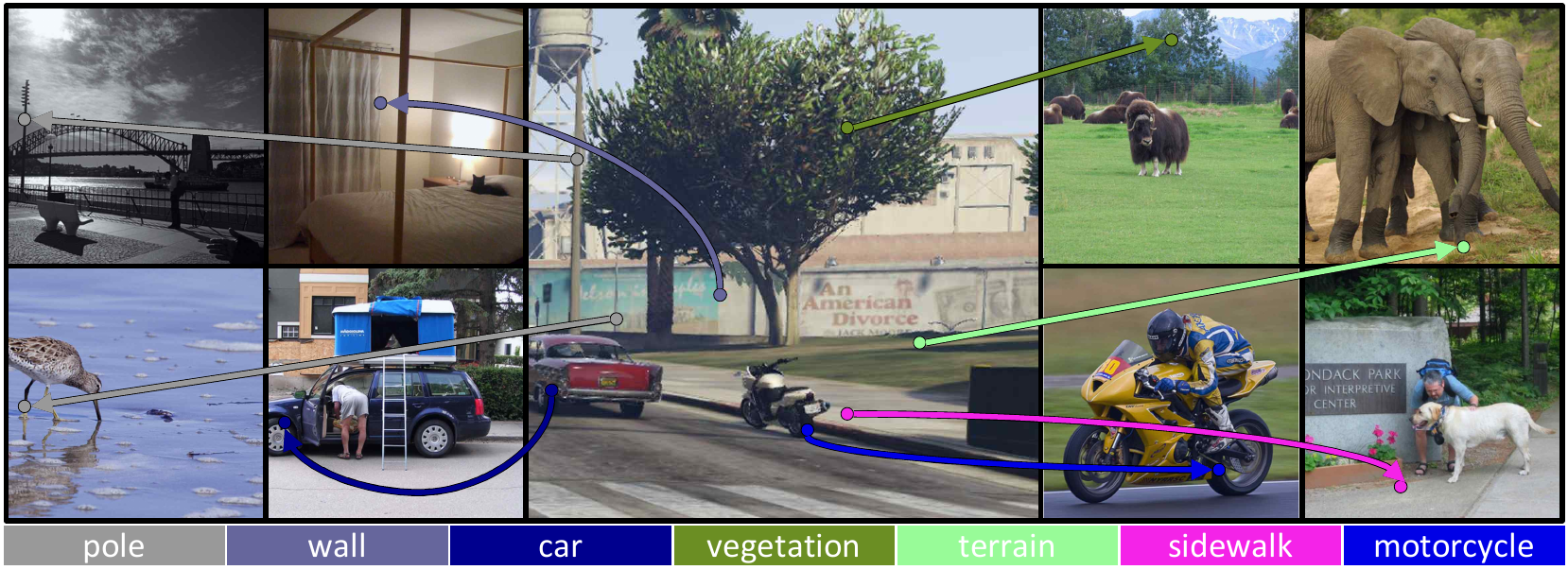}}
    \caption{}
    \label{fig:cel_supp2}
  \end{subfigure}
  \begin{subfigure}{1.0\textwidth}
    \raisebox{-\height}{\includegraphics[width=\textwidth]{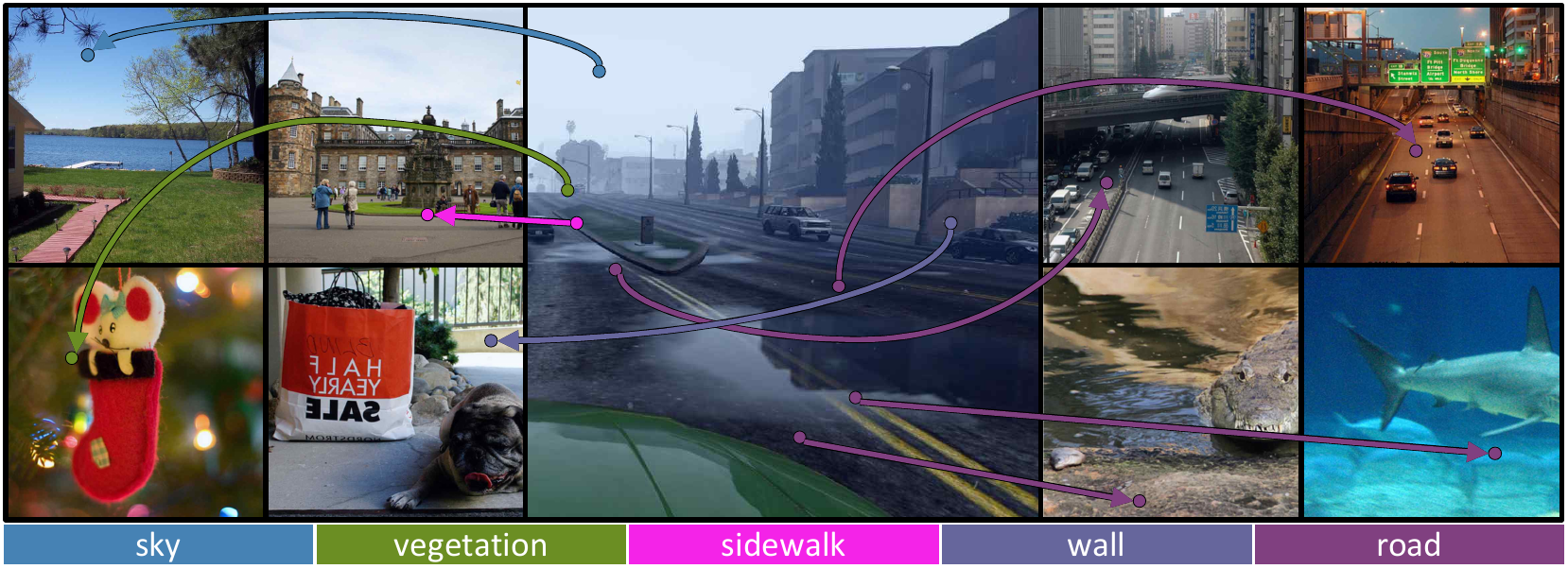}}
    \caption{}
    \label{fig:cel_supp3}
  \end{subfigure}
\caption{
Visualization of extended wild contents. 
The center image is sampled from the GTAV dataset while the remaining eight images are sampled from ImageNet.
The contents are extended from the centered GTAV image to the eight ImageNet images, and each color represents the semantic label of the content in GTAV. 
}
\label{fig:cel_supp}
\end{figure*}

\vspace{-0.2em}
\cref{fig:cel_supp} illustrates extended the wild contents from the source (\ie, GTAV~\cite{richter2016playing}) to the wild (\ie, ImageNet~\cite{deng2009imagenet}).
The eight contents are extended from a centered image in GTAV to the eight ImageNet images, and each color represents the semantic label of the content in GTAV.
After network training, we used our final ResNet-50~\cite{he2016deep} with DeepLabV3+~\cite{chen2018encoder} model to visualize the pixels in the wild image extended from each pixel in the source image.
Although the source content was extended to the wild content closest to the stylized source content in the feature space without using any wild label, the source content was extended to the wild content with the same semantic information as itself, as shown in~\cref{fig:cel_supp1}.
These content extensions increase the intra-class content variability in the latent embedding space and alleviate overfitting to the source contents.

There are various semantic classes in the wild dataset that are not considered in the source dataset, and we will refer to them as wild-only classes in this supplementary material.
The source content is sometimes extended to wild-only class content, such as the thin pole-class pixel being extended to the thin bird's leg pixel in~\cref{fig:cel_supp2}.
The proposed content extension learning is a pixel-wise approach. Therefore, if two pixels have similar features, content extension to other classes with similar shapes is observed.
This is not limited to human-annotated class labels and encourages the network to learn generalized features by reducing the distance between contents with similar semantic information in the feature space.
This may provide clues to generalization performance improvements for unseen contents.
In~\cref{fig:cel_supp3}, it was observed that some road pixels were extended to the waterside ground and underwater ground pixels.
It is expected that content extension to these wild-only classes will guide the network to correctly predict wet road and puddle pixels as road classes in rainy scenes.
With content extension learning, WildNet makes reliable predictions in various environments, such as wet vegetation in the fifth row of~\cref{fig:compare_r50_gta_img_base_isw_ours} and light-reflected road in the first row of~\cref{fig:compare_r50_bdd_img_base_isw_ours}.

\vspace{-0.2em}
\subsection{Wild-Stylized Features}
\begin{figure*}
  \centering
  \includegraphics[width=\textwidth]{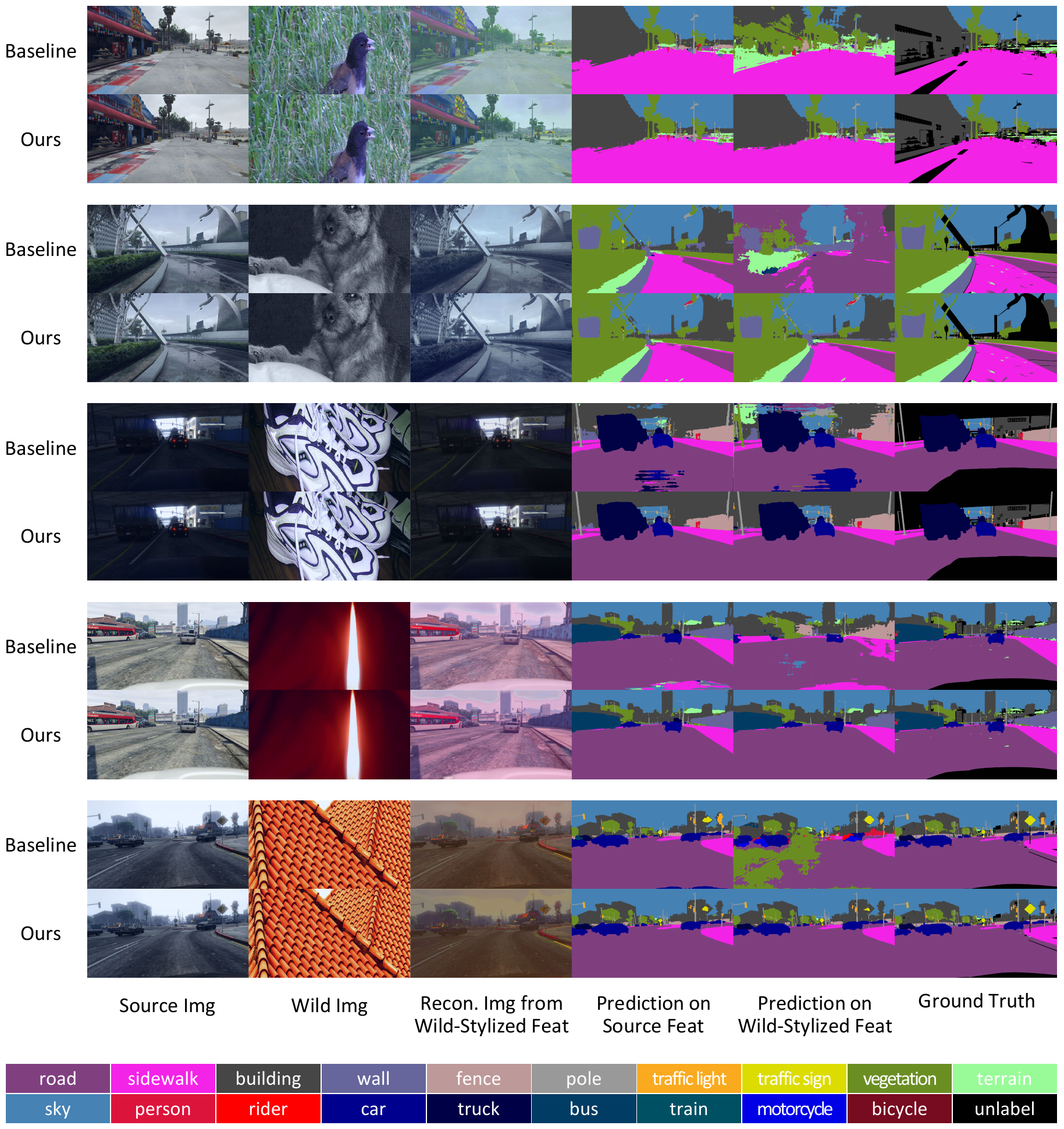}
  \caption{
  Given the source image and ground truth label, we stylize the source feature using the style of the wild feature from the given wild image. 
  To visualize the wild-stylized source feature, we reconstructed an image from the wild-stylized feature using U-Net~\cite{ronneberger2015u}.
  The reconstructed image from wild-stylized source feature includes spatial information of the source image and style information of the wild image.
  The baseline model fails to make correct predictions from wild-stylized features, as opposed to accurate predictions from source features.
  In contrast, the proposed WildNet makes accurate predictions on wild-stylized features by applying style extension learning and semantic consistency regularization in the training process.
  }
  \label{fig:fs_supp}
\end{figure*}

\vspace{-0.2em}
\cref{fig:fs_supp} shows the importance of learning task-specific information from wild-stylized features.
Given the source image and ground truth label (see the first and sixth columns in~\cref{fig:fs_supp}), we diversify source data by stylizing the source feature using the style of the wild feature from the given wild image (see the second column in~\cref{fig:fs_supp}).
To maintain the spatial information of the source feature, we apply adaptive instance normalization~\cite{huang2017arbitrary} with channel-wise mean and standard deviation for the source and wild features.
To visualize that the wild-stylized source feature contains the spatial information of the source feature and the style of the wild feature, we reconstructed the image from the wild-stylized feature using the U-Net~\cite{ronneberger2015u} structure following the process of RobustNet~\cite{choi2021robustnet} reconstructing the input image from the whitened feature.
After training the baseline model and our model on the semantic segmentation task, we freeze the weights of the pre-trained model and add a decoder to learn the image reconstruction.
The reconstructed images from the wild-stylized source features show that both the baseline model and our model transform the style while successfully maintaining the spatial information of the source features (see the third column in~\cref{fig:fs_supp}).
Nevertheless, the baseline model fails to make accurate predictions from wild-stylized source features, as opposed to making accurate predictions from the original source features (see the fifth and fourth columns in~\cref{fig:fs_supp}).

To address this issue, we train our WildNet with the proposed style extension learning and semantic consistency regularization methods.
The style extension learning enables our model to naturally adapt to various styles by learning task-specific information from the wild-stylized features.
Moreover, the semantic consistency regularization regularizes the model, enabling the capture of consistent semantic information from the wild-stylized and original source features.
As a result, our model captures generalized semantic information from features of various styles and makes correct predictions on wild-stylized source features (see the fifth column in~\cref{fig:fs_supp}).

\vspace{-0.2em}
\subsection{Qualitative Results}
\begin{figure*}
  \centering
  \begin{subfigure}{0.19\textwidth}
    \raisebox{-\height}{\includegraphics[width=\textwidth, height=0.55\textwidth]{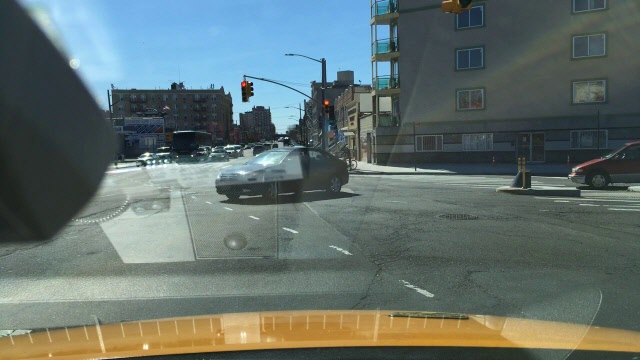}}
  \end{subfigure}
  \hfill
  \begin{subfigure}{0.19\textwidth}
    \raisebox{-\height}{\includegraphics[width=\textwidth, height=0.55\textwidth]{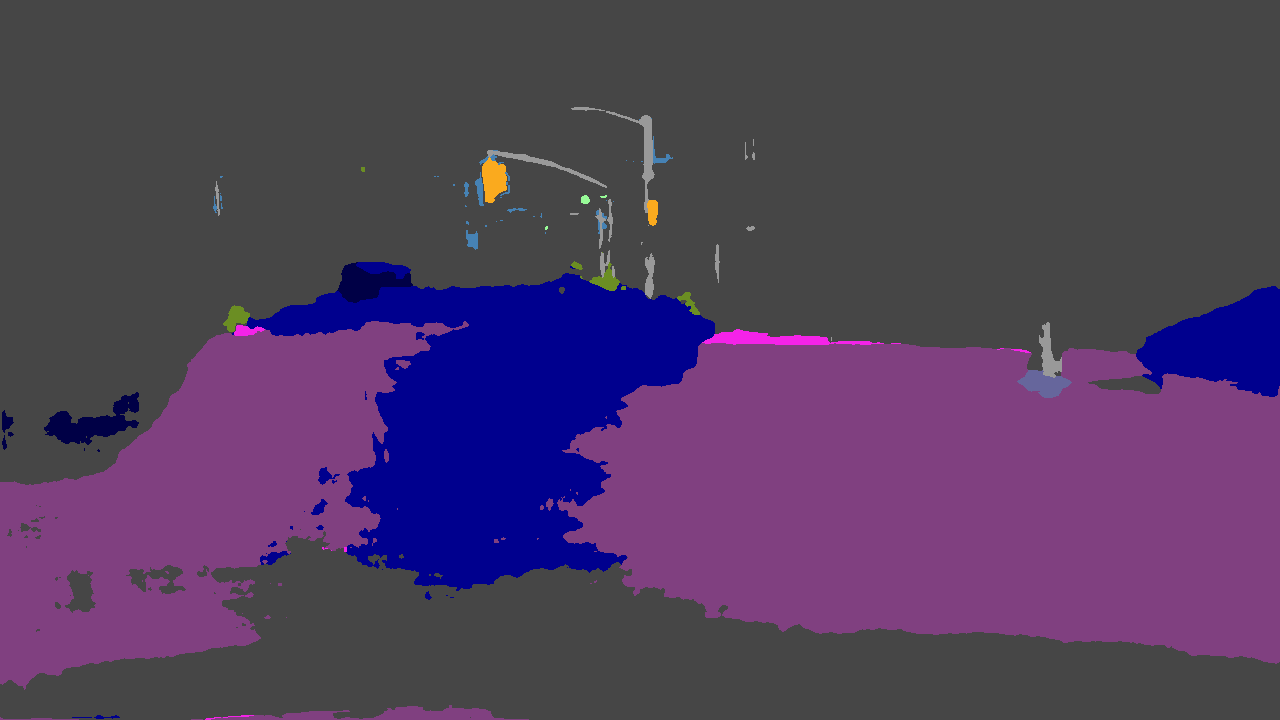}}
  \end{subfigure}
  \hfill
  \begin{subfigure}{0.19\textwidth}
    \raisebox{-\height}{\includegraphics[width=\textwidth, height=0.55\textwidth]{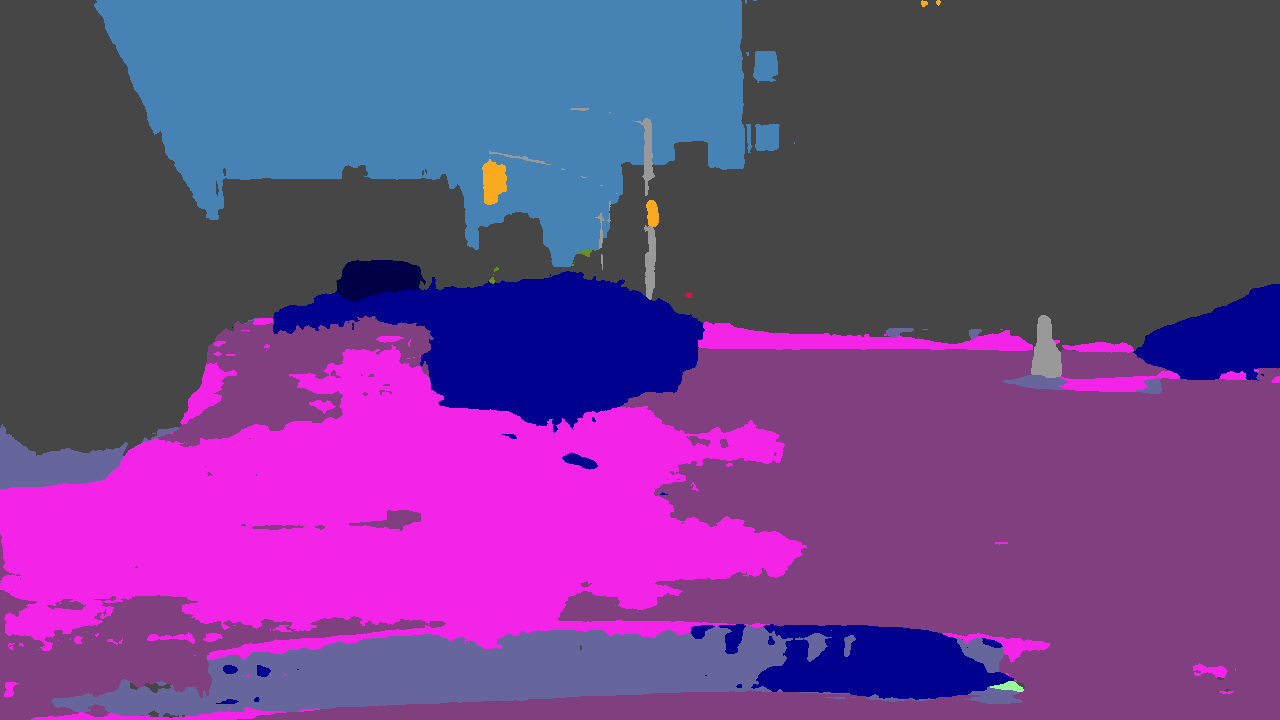}}
  \end{subfigure}
  \hfill
  \begin{subfigure}{0.19\textwidth}
    \raisebox{-\height}{\includegraphics[width=\textwidth, height=0.55\textwidth]{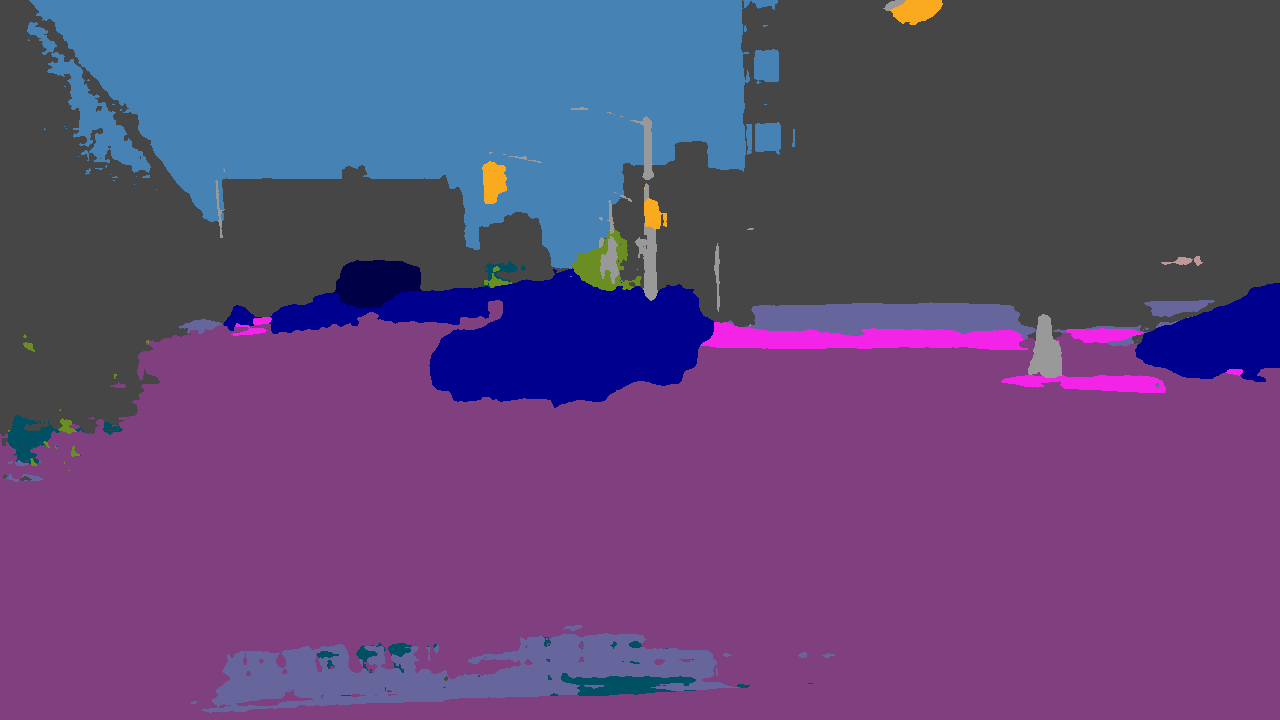}}
  \end{subfigure}
  \hfill
  \begin{subfigure}{0.19\textwidth}
    \raisebox{-\height}{\includegraphics[width=\textwidth, height=0.55\textwidth]{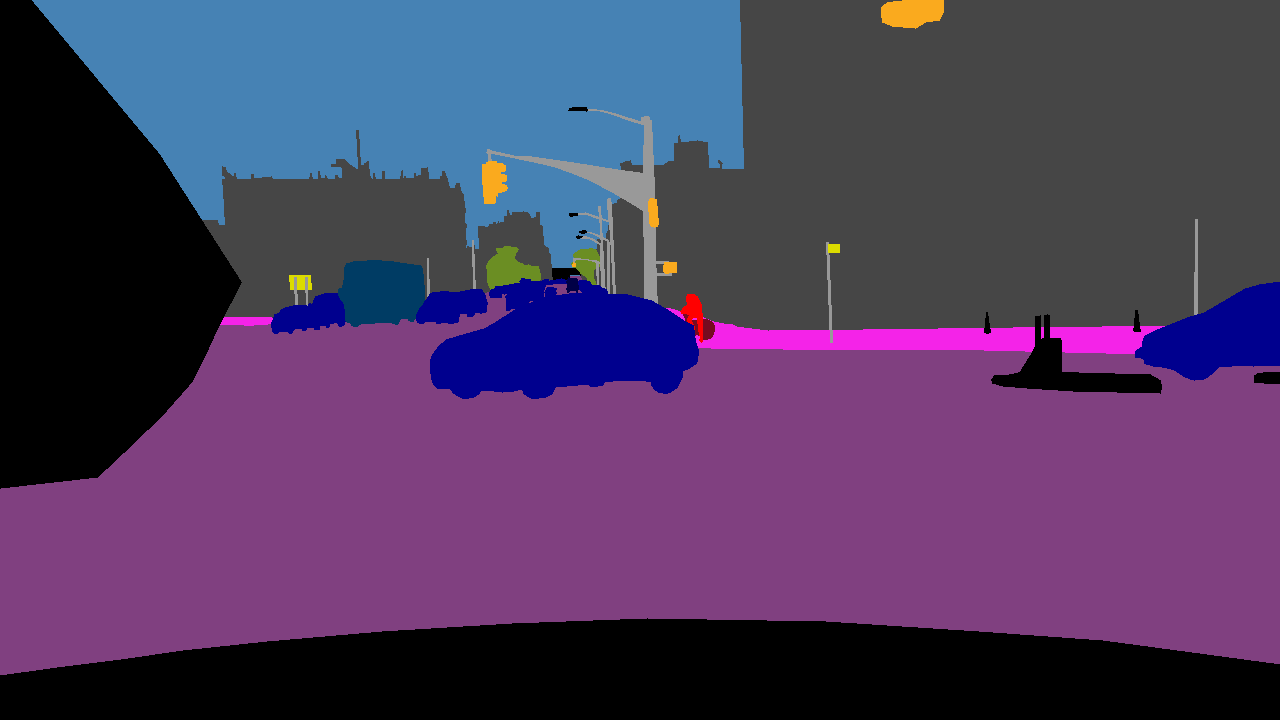}}
  \end{subfigure}
  \begin{subfigure}{0.19\textwidth}
    \raisebox{-\height}{\includegraphics[width=\textwidth, height=0.55\textwidth]{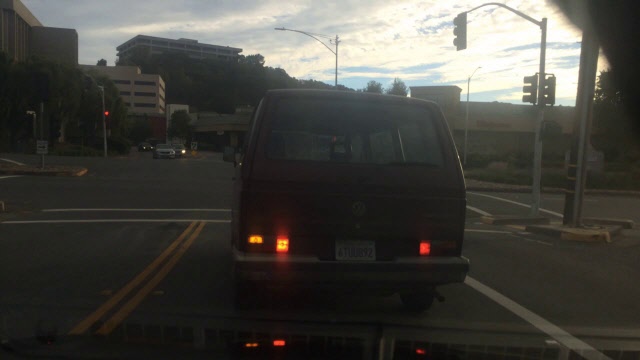}}
  \end{subfigure}
  \hfill
  \begin{subfigure}{0.19\textwidth}
    \raisebox{-\height}{\includegraphics[width=\textwidth, height=0.55\textwidth]{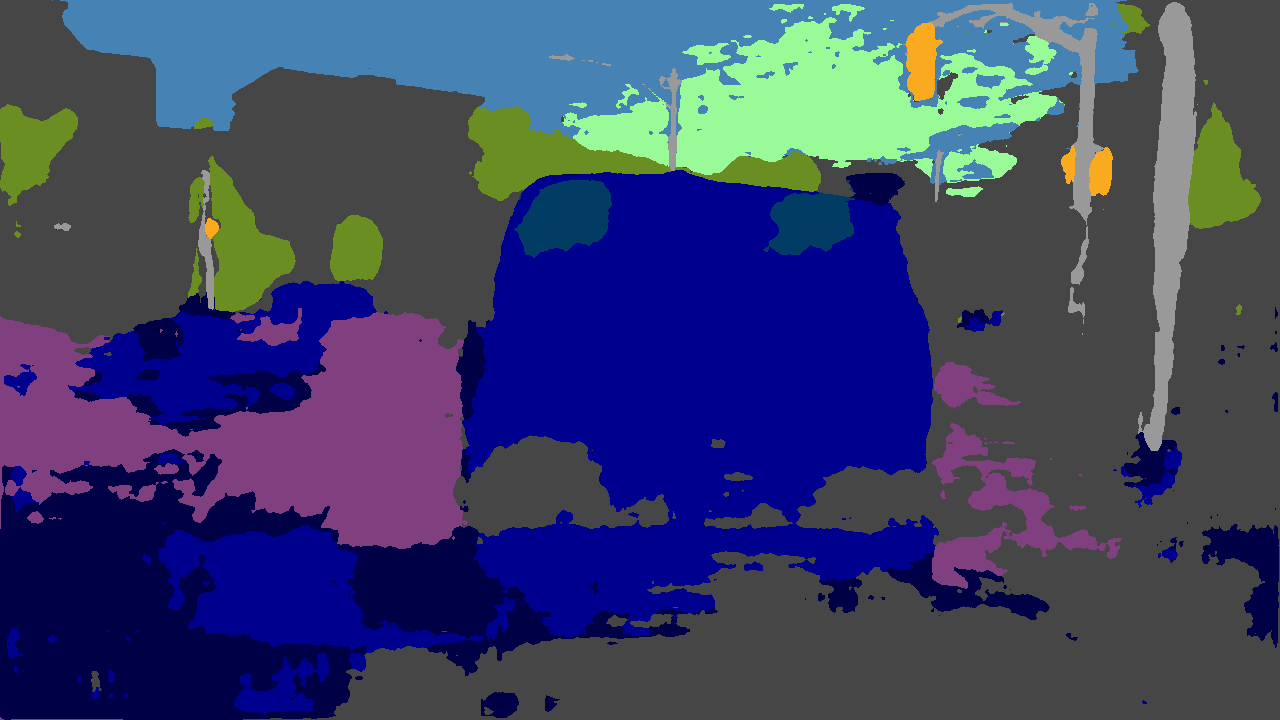}}
  \end{subfigure}
  \hfill
  \begin{subfigure}{0.19\textwidth}
    \raisebox{-\height}{\includegraphics[width=\textwidth, height=0.55\textwidth]{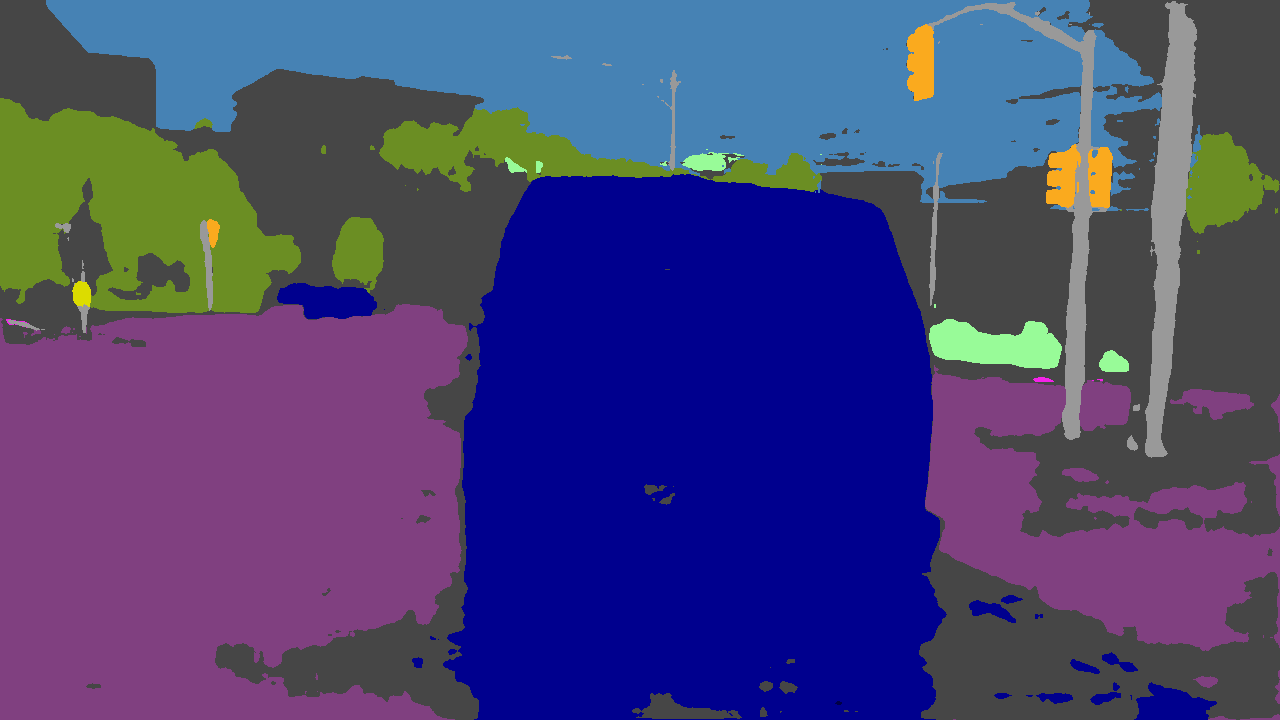}}
  \end{subfigure}
  \hfill
  \begin{subfigure}{0.19\textwidth}
    \raisebox{-\height}{\includegraphics[width=\textwidth, height=0.55\textwidth]{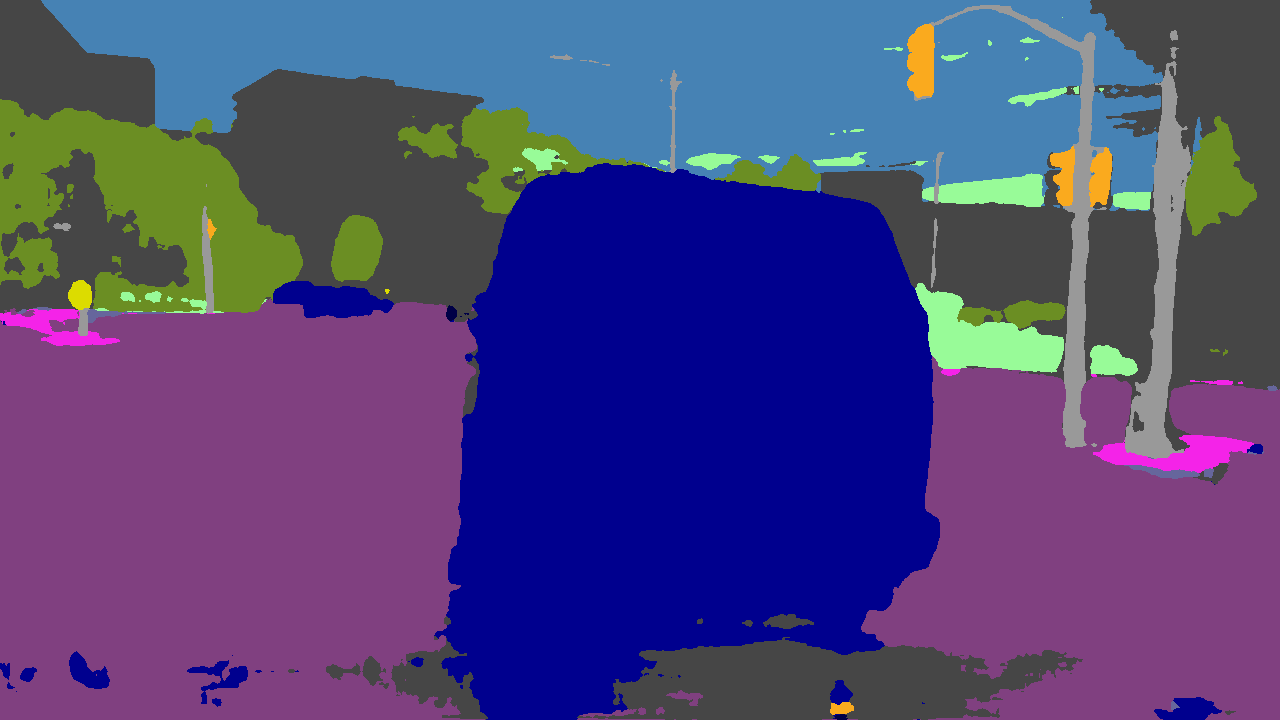}}
  \end{subfigure}
  \hfill
  \begin{subfigure}{0.19\textwidth}
    \raisebox{-\height}{\includegraphics[width=\textwidth, height=0.55\textwidth]{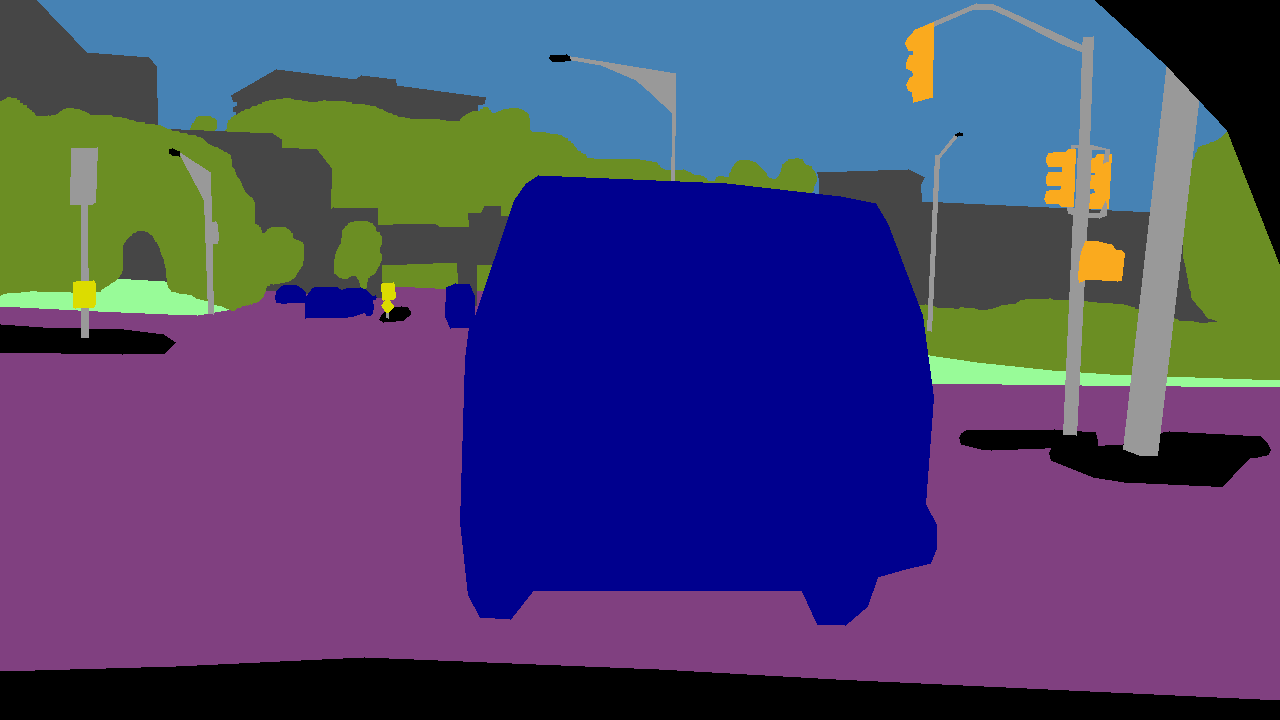}}
  \end{subfigure}
  \begin{subfigure}{0.19\textwidth}
    \raisebox{-\height}{\includegraphics[width=\textwidth, height=0.55\textwidth]{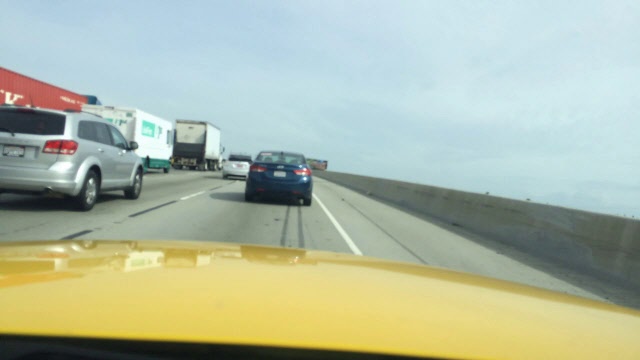}}
  \end{subfigure}
  \hfill
  \begin{subfigure}{0.19\textwidth}
    \raisebox{-\height}{\includegraphics[width=\textwidth, height=0.55\textwidth]{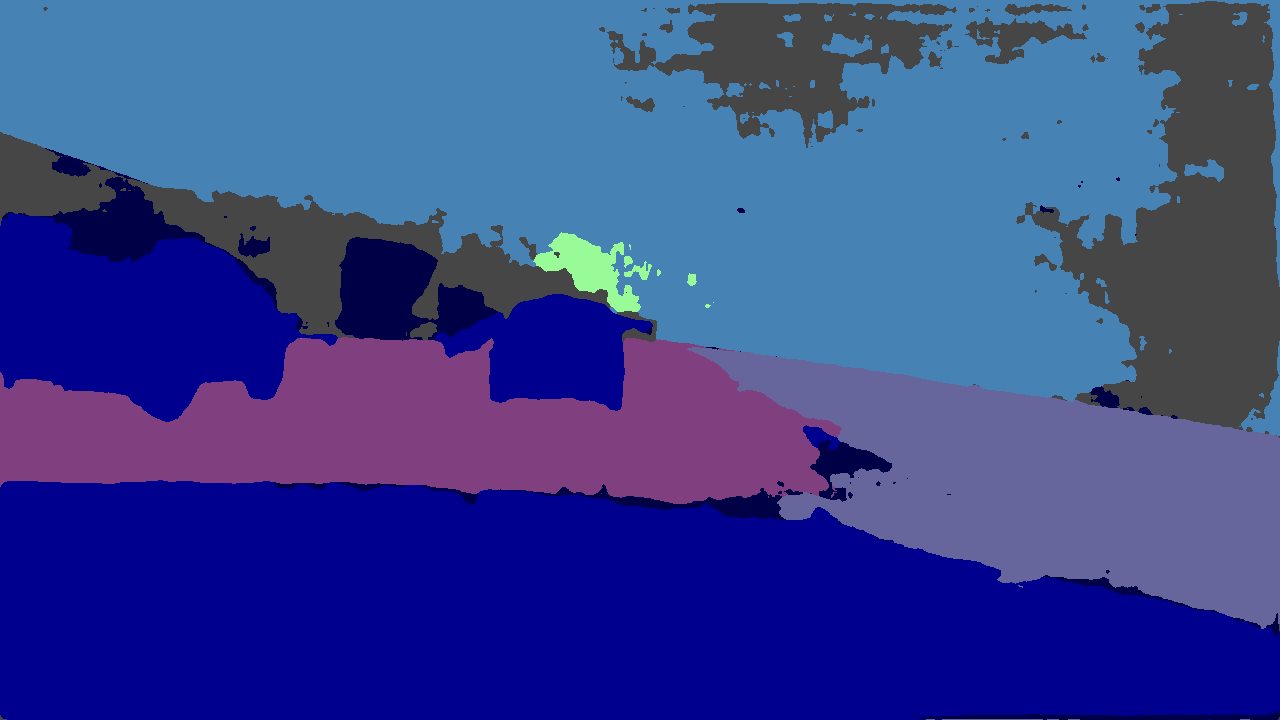}}
  \end{subfigure}
  \hfill
  \begin{subfigure}{0.19\textwidth}
    \raisebox{-\height}{\includegraphics[width=\textwidth, height=0.55\textwidth]{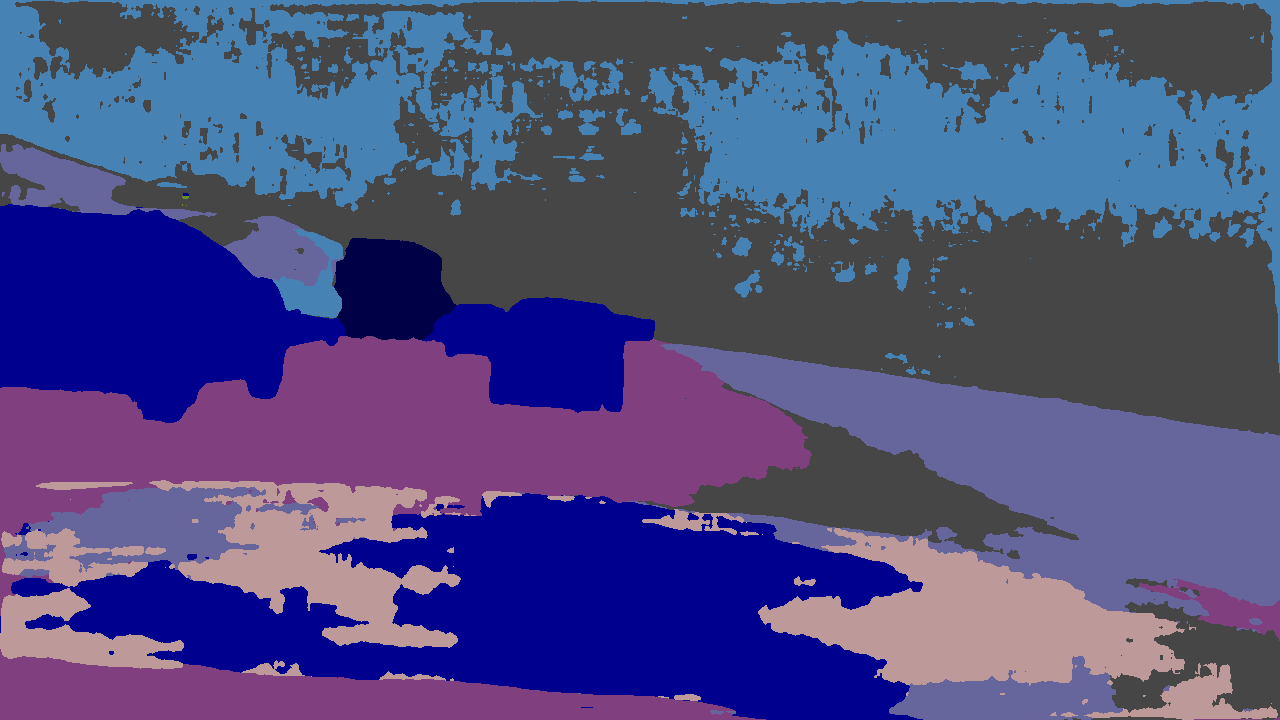}}
  \end{subfigure}
  \hfill
  \begin{subfigure}{0.19\textwidth}
    \raisebox{-\height}{\includegraphics[width=\textwidth, height=0.55\textwidth]{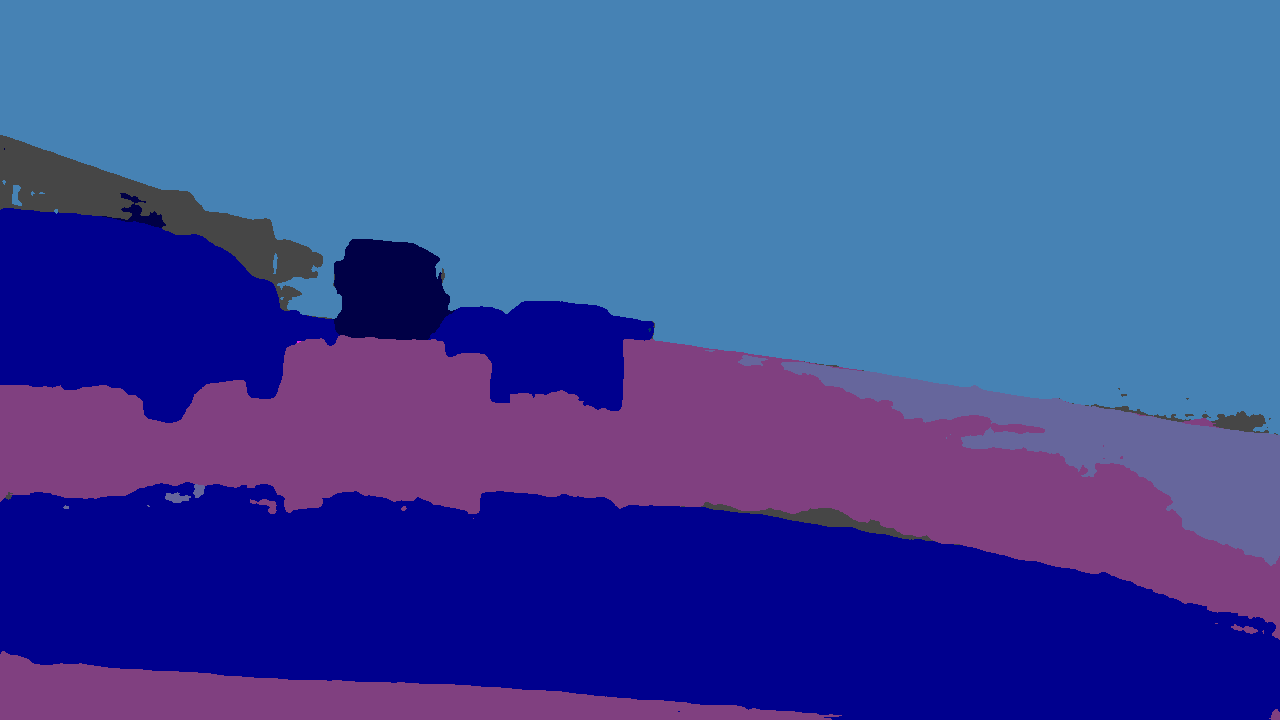}}
  \end{subfigure}
  \hfill
  \begin{subfigure}{0.19\textwidth}
    \raisebox{-\height}{\includegraphics[width=\textwidth, height=0.55\textwidth]{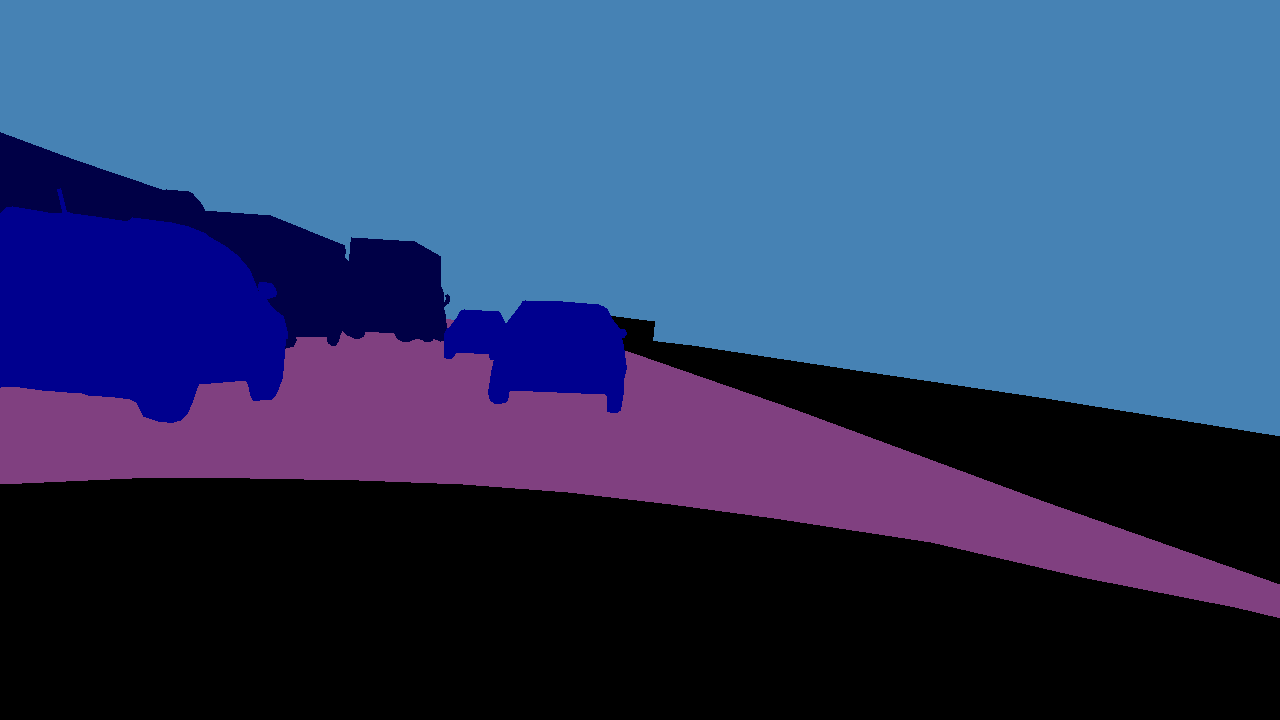}}
  \end{subfigure}
  \begin{subfigure}{0.19\textwidth}
    \raisebox{-\height}{\includegraphics[width=\textwidth, height=0.55\textwidth]{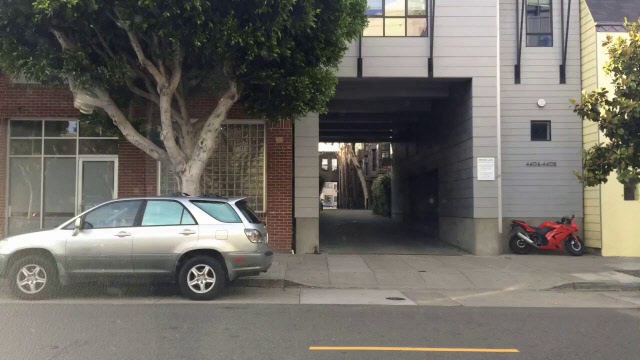}}
  \end{subfigure}
  \hfill
  \begin{subfigure}{0.19\textwidth}
    \raisebox{-\height}{\includegraphics[width=\textwidth, height=0.55\textwidth]{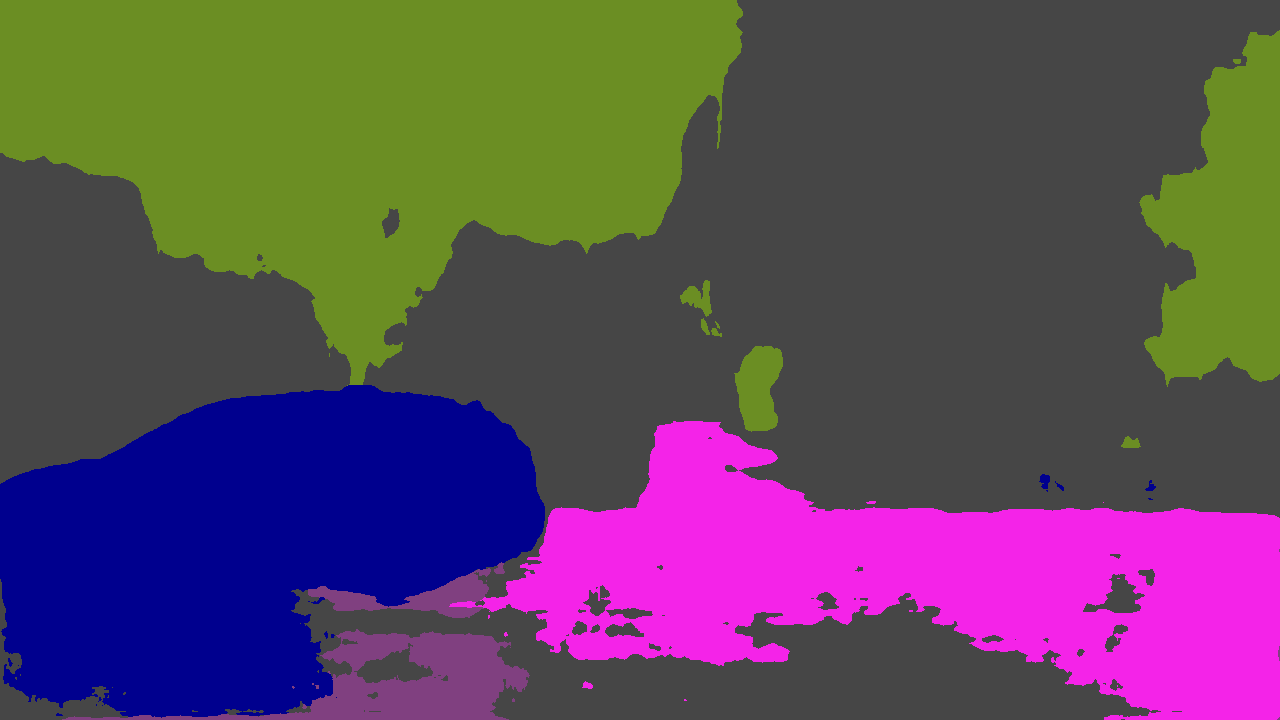}}
  \end{subfigure}
  \hfill
  \begin{subfigure}{0.19\textwidth}
    \raisebox{-\height}{\includegraphics[width=\textwidth, height=0.55\textwidth]{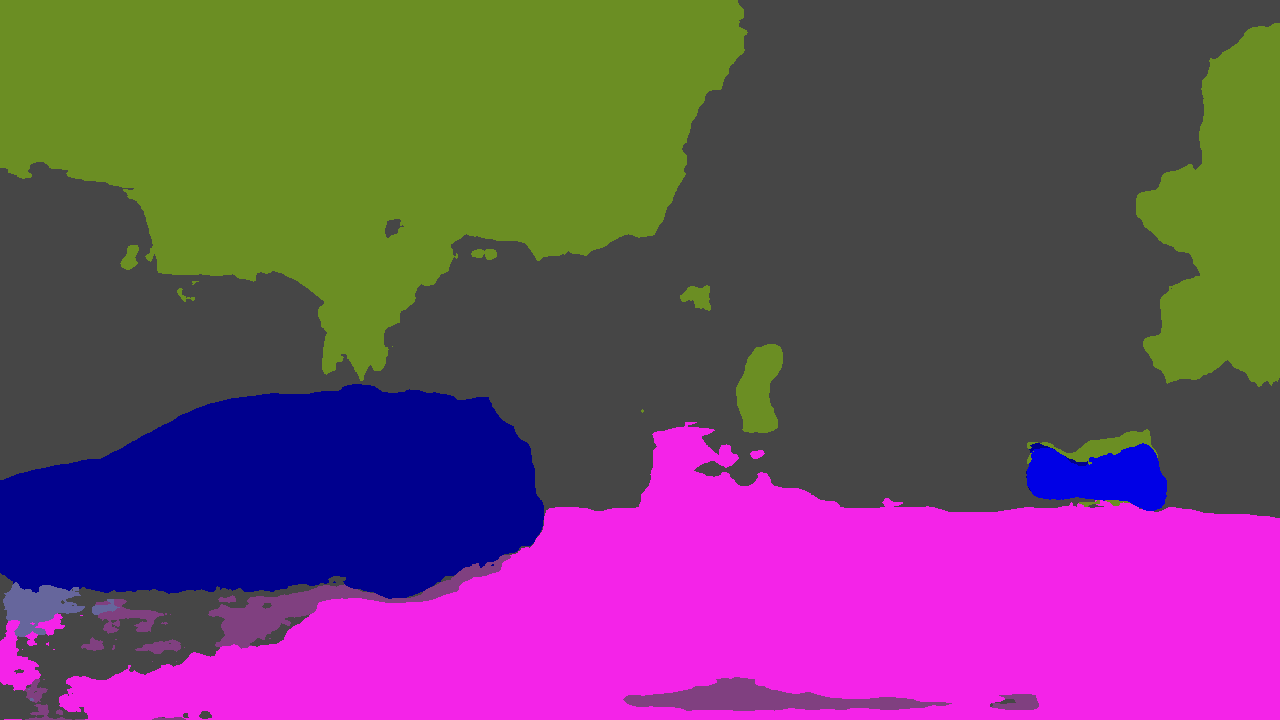}}
  \end{subfigure}
  \hfill
  \begin{subfigure}{0.19\textwidth}
    \raisebox{-\height}{\includegraphics[width=\textwidth, height=0.55\textwidth]{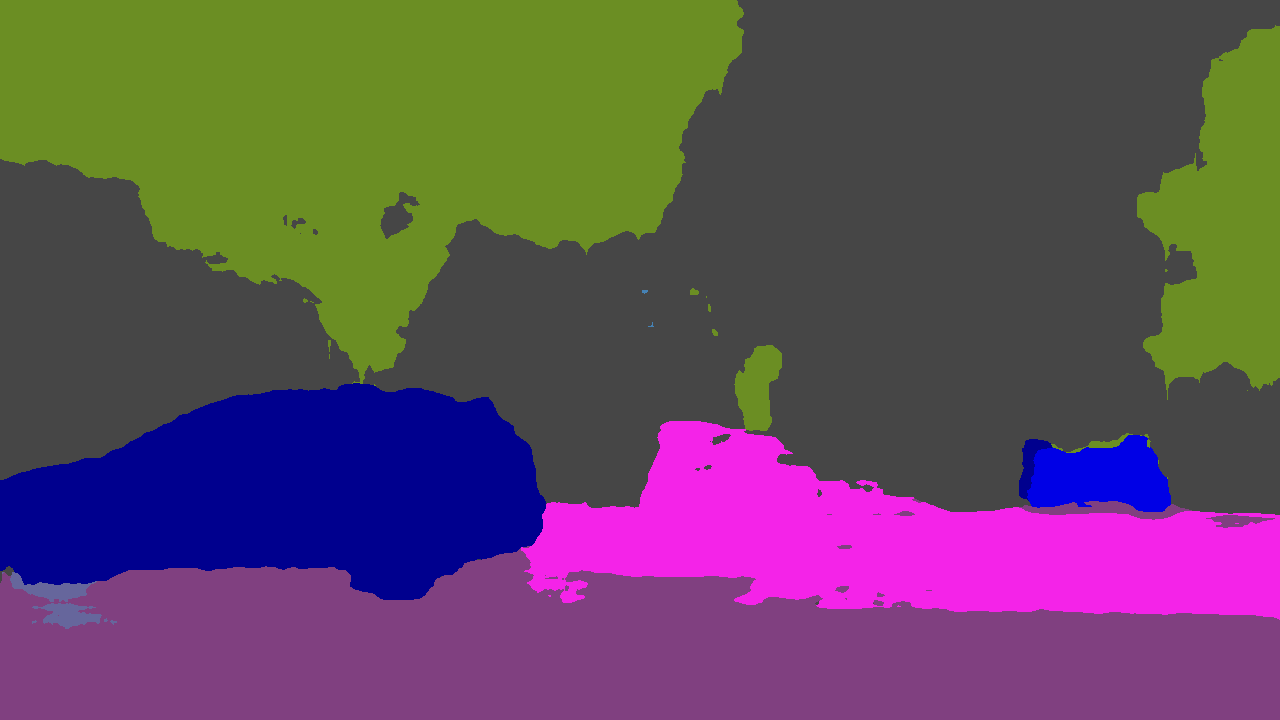}}
  \end{subfigure}
  \hfill
  \begin{subfigure}{0.19\textwidth}
    \raisebox{-\height}{\includegraphics[width=\textwidth, height=0.55\textwidth]{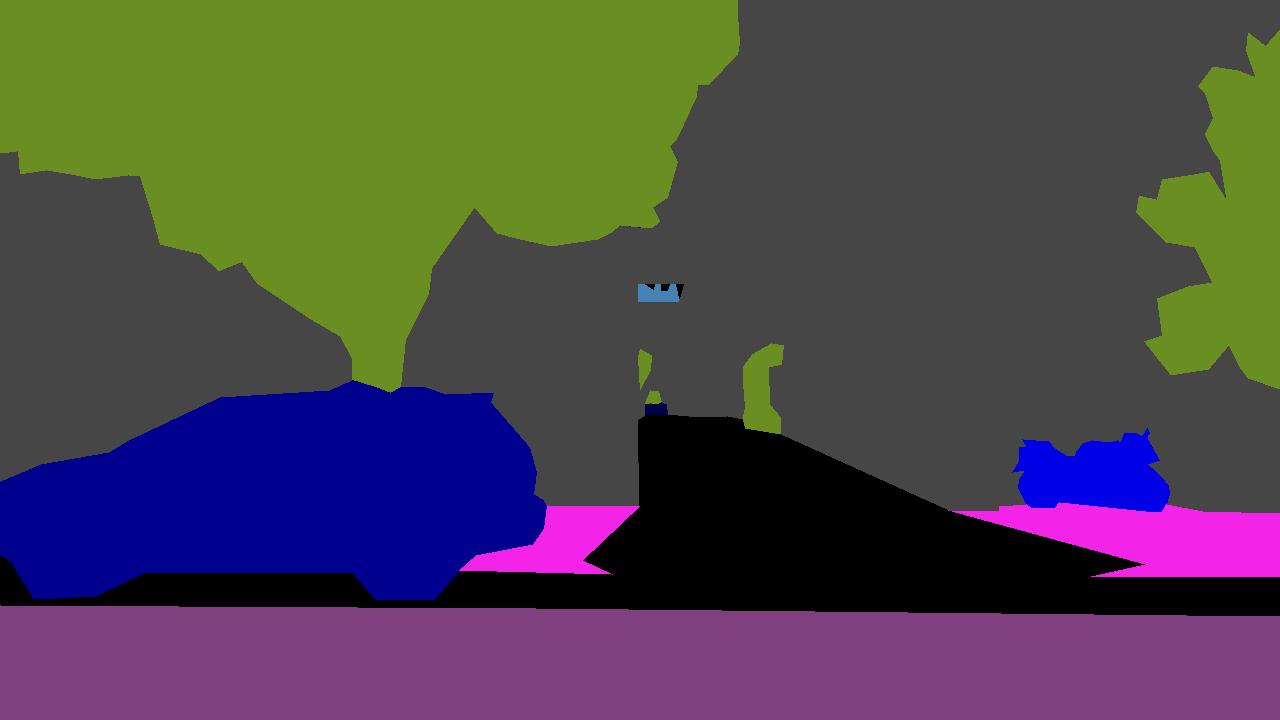}}
  \end{subfigure}
  \begin{subfigure}{0.19\textwidth}
    \raisebox{-\height}{\includegraphics[width=\textwidth, height=0.55\textwidth]{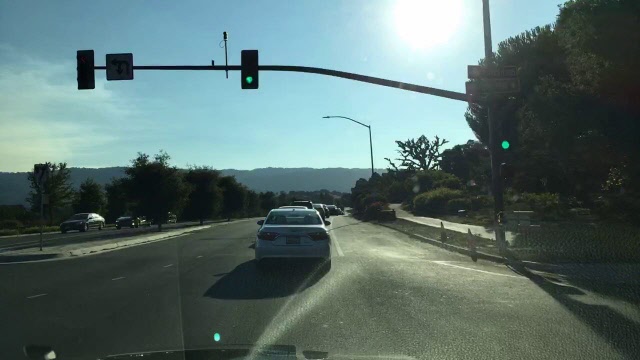}}
    \caption*{Unseen domain image}
    \label{fig:compare_r50_bdd_img}
  \end{subfigure}
  \hfill
  \begin{subfigure}{0.19\textwidth}
    \raisebox{-\height}{\includegraphics[width=\textwidth, height=0.55\textwidth]{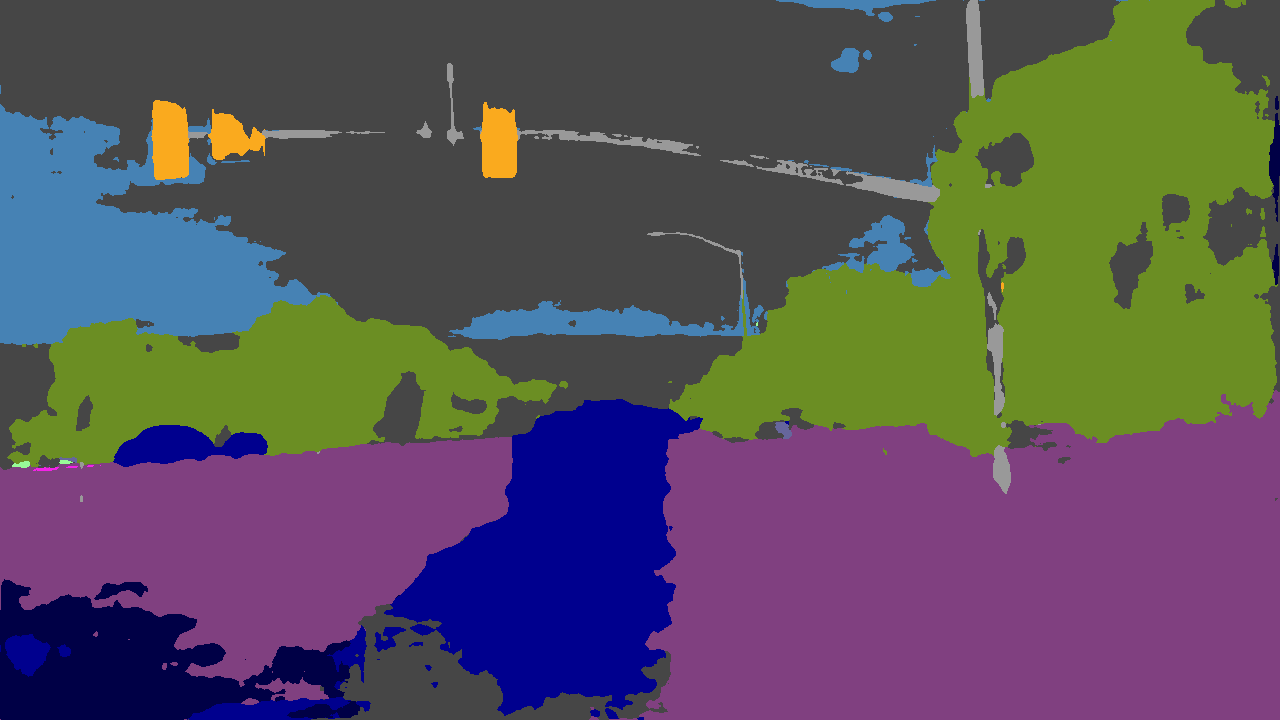}}
    \caption*{Baseline}
    \label{fig:compare_r50_bdd_base}
  \end{subfigure}
  \hfill
  \begin{subfigure}{0.19\textwidth}
    \raisebox{-\height}{\includegraphics[width=\textwidth, height=0.55\textwidth]{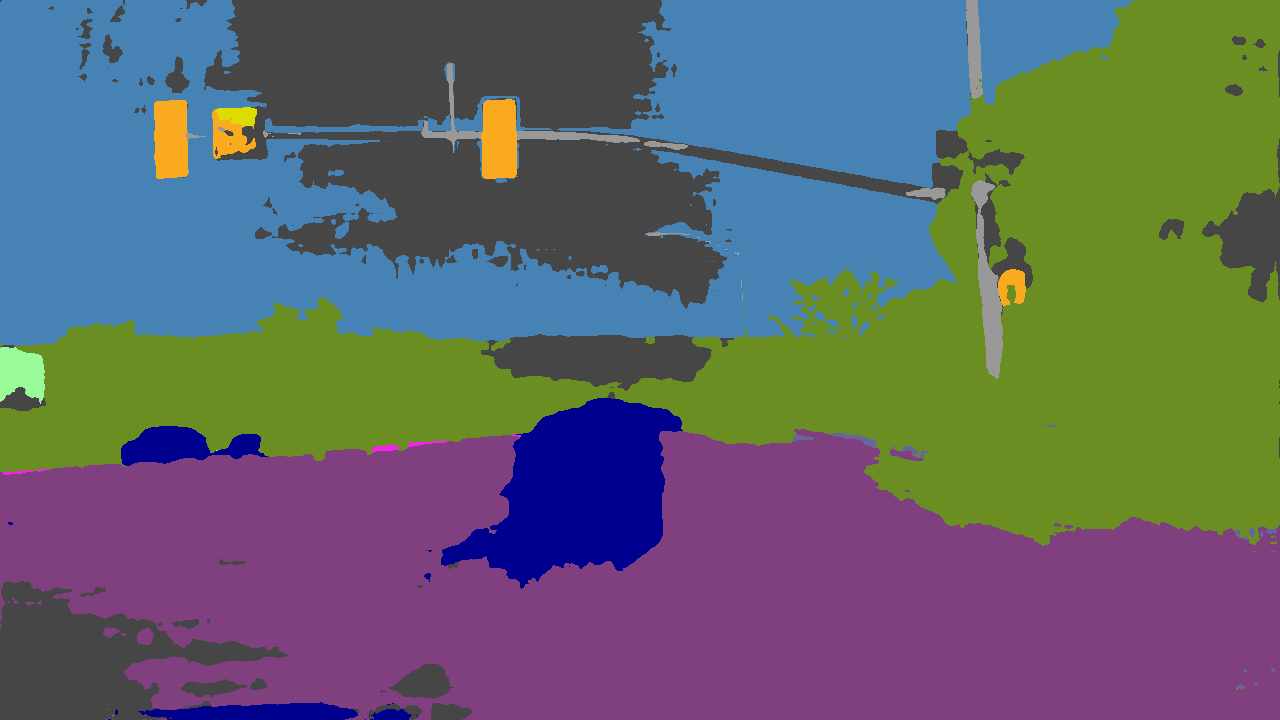}}
    \caption*{RobustNet}
    \label{fig:compare_r50_bdd_isw}
  \end{subfigure}
  \hfill
  \begin{subfigure}{0.19\textwidth}
    \raisebox{-\height}{\includegraphics[width=\textwidth, height=0.55\textwidth]{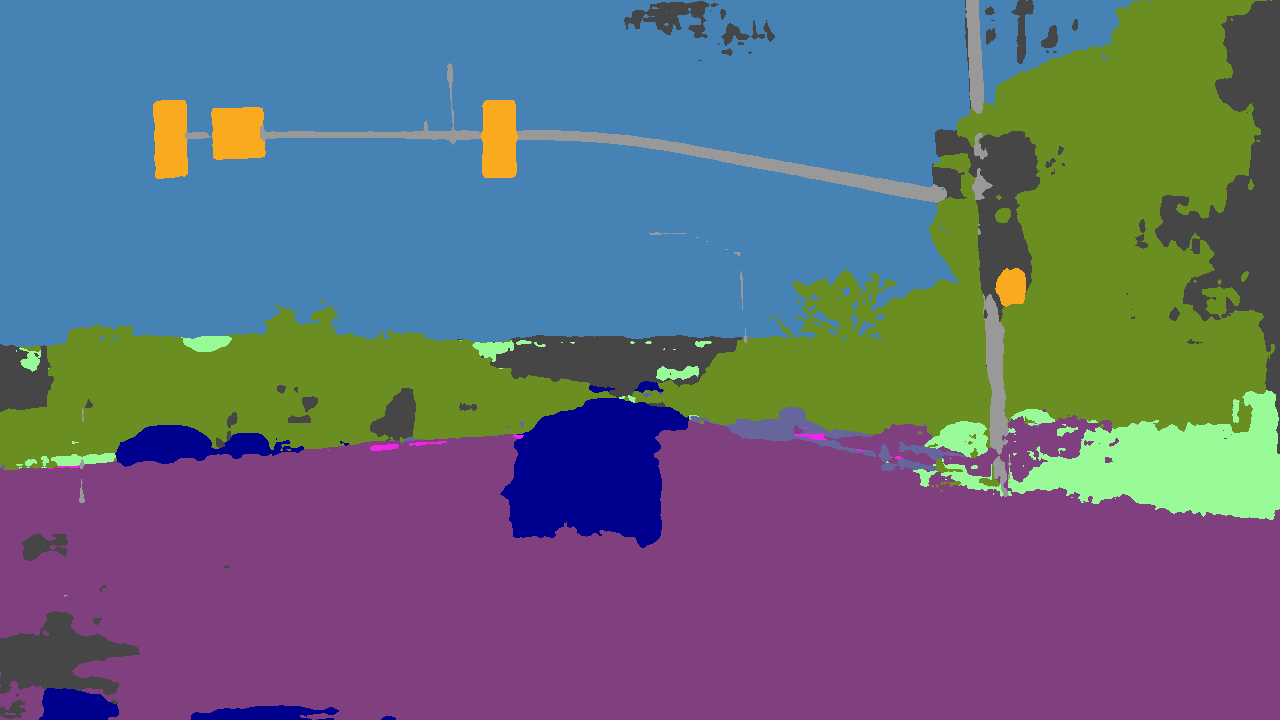}}
    \caption*{\textbf{Ours (WildNet)}}
    \label{fig:compare_r50_bdd_ours}
  \end{subfigure}
  \hfill
  \begin{subfigure}{0.19\textwidth}
    \raisebox{-\height}{\includegraphics[width=\textwidth, height=0.55\textwidth]{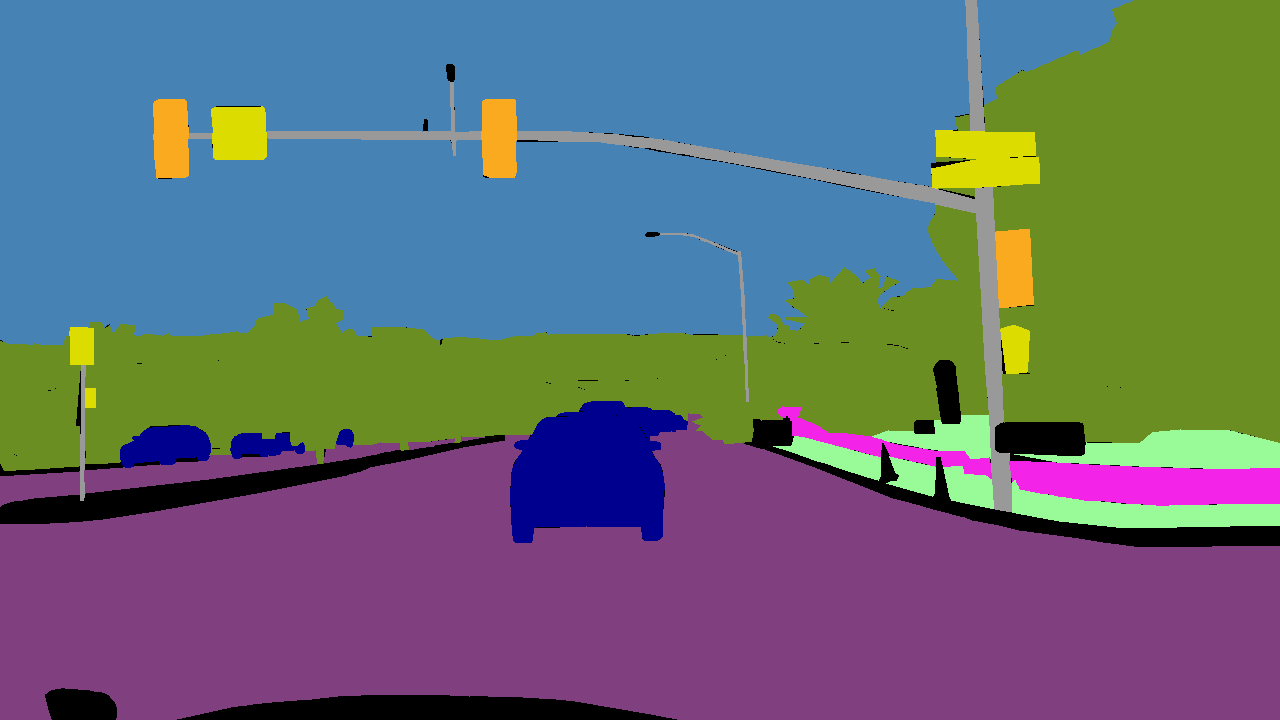}}
    \caption*{Ground truth}
    \label{fig:compare_r50_bdd_gt}
  \end{subfigure}
  \caption{Semantic segmentation results on unseen domain images in BDD100K with the models trained on GTAV.
  }
  \vspace{+1.0em}
  \label{fig:compare_r50_bdd_img_base_isw_ours}
\end{figure*}

\begin{figure*}
  \centering
  \begin{subfigure}{0.19\textwidth}
    \raisebox{-\height}{\includegraphics[width=\textwidth, height=0.55\textwidth]{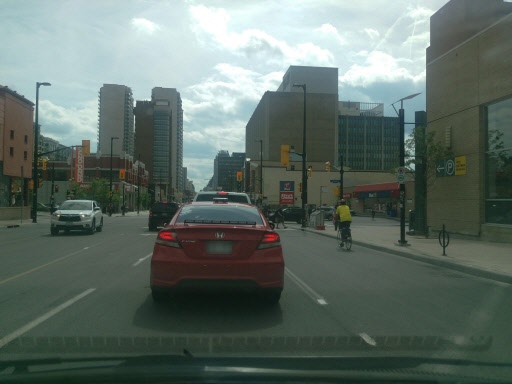}}
  \end{subfigure}
  \hfill
  \begin{subfigure}{0.19\textwidth}
    \raisebox{-\height}{\includegraphics[width=\textwidth, height=0.55\textwidth]{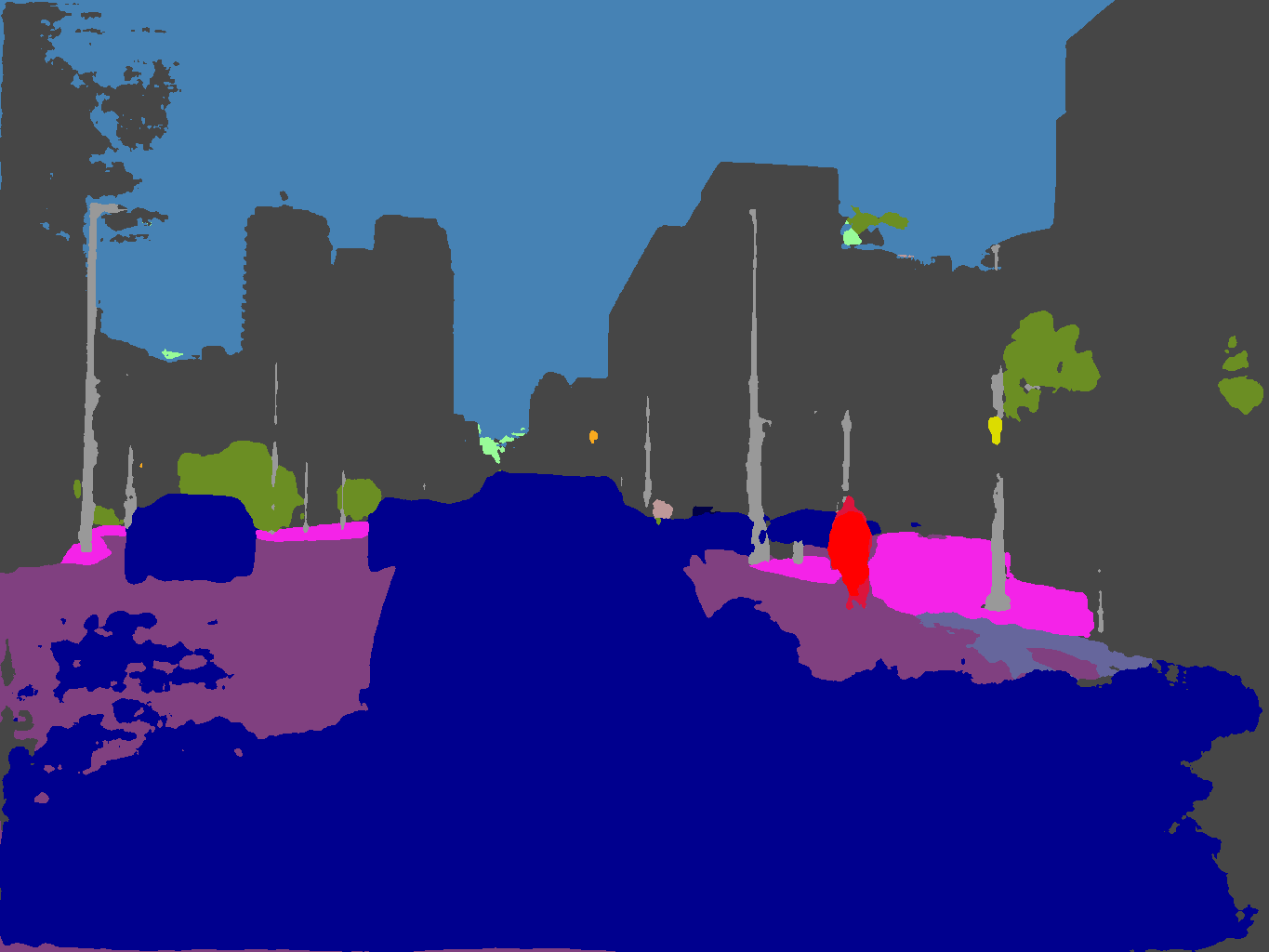}}
  \end{subfigure}
  \hfill
  \begin{subfigure}{0.19\textwidth}
    \raisebox{-\height}{\includegraphics[width=\textwidth, height=0.55\textwidth]{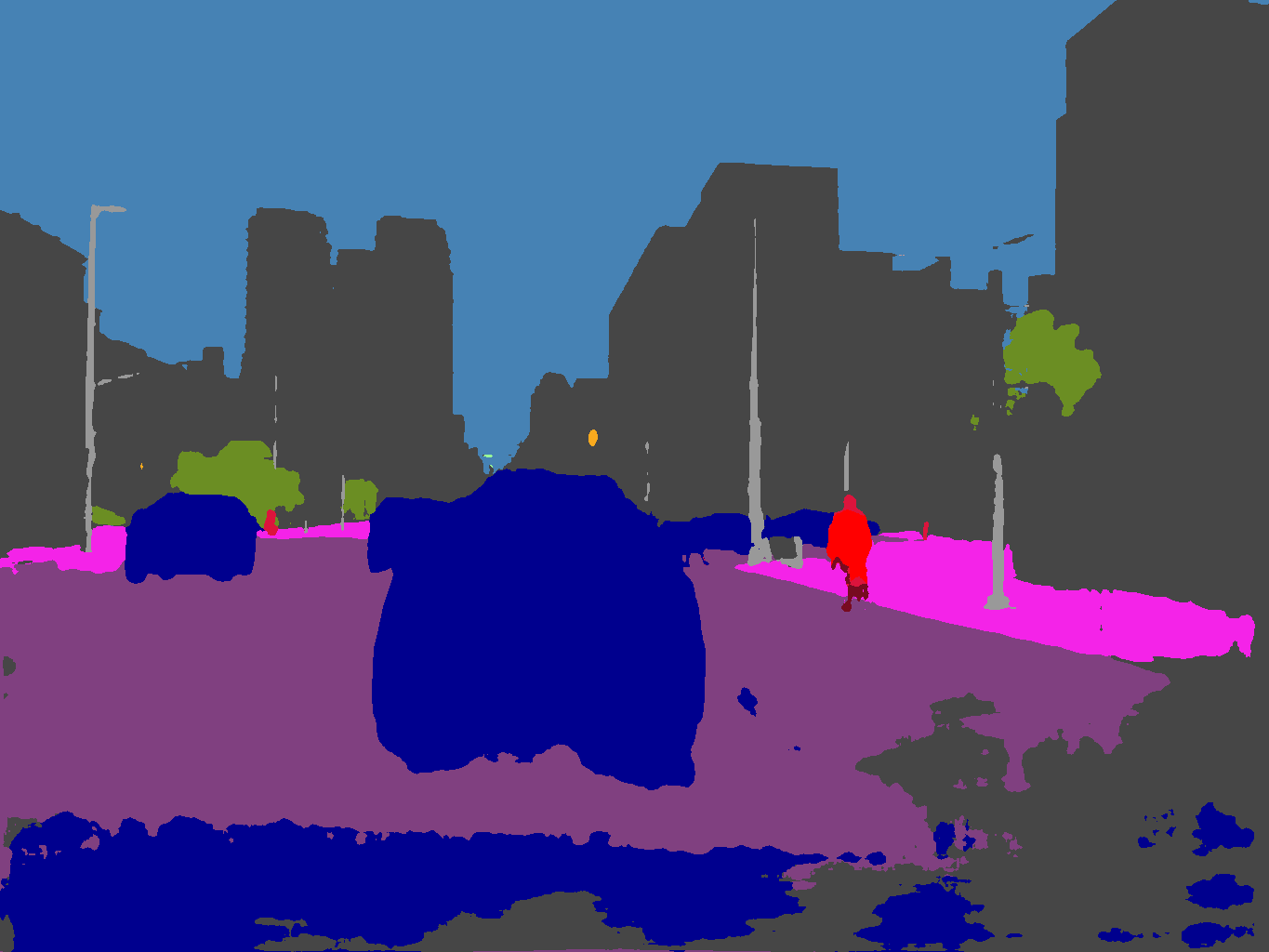}}
  \end{subfigure}
  \hfill
  \begin{subfigure}{0.19\textwidth}
    \raisebox{-\height}{\includegraphics[width=\textwidth, height=0.55\textwidth]{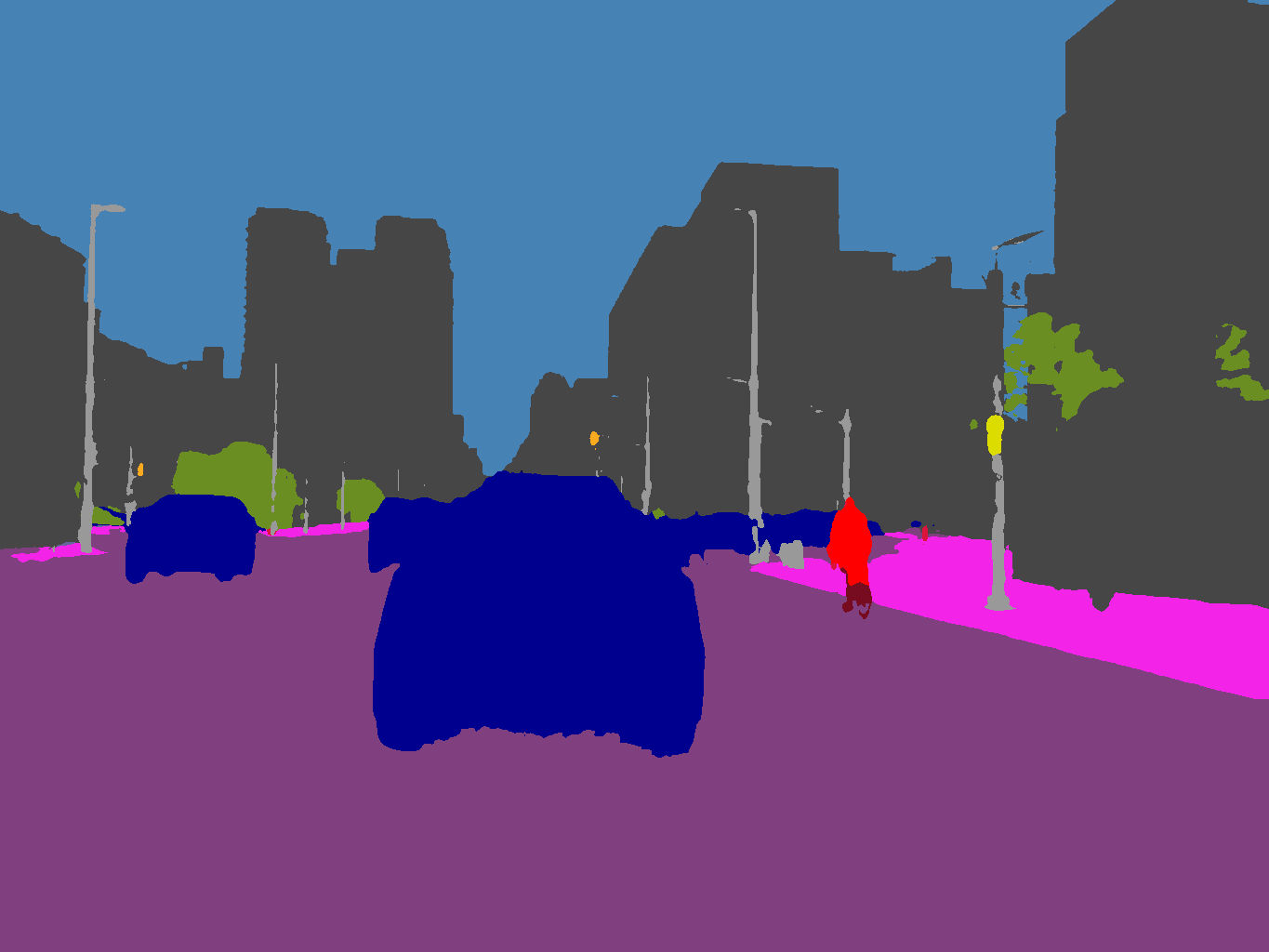}}
  \end{subfigure}
  \hfill
  \begin{subfigure}{0.19\textwidth}
    \raisebox{-\height}{\includegraphics[width=\textwidth, height=0.55\textwidth]{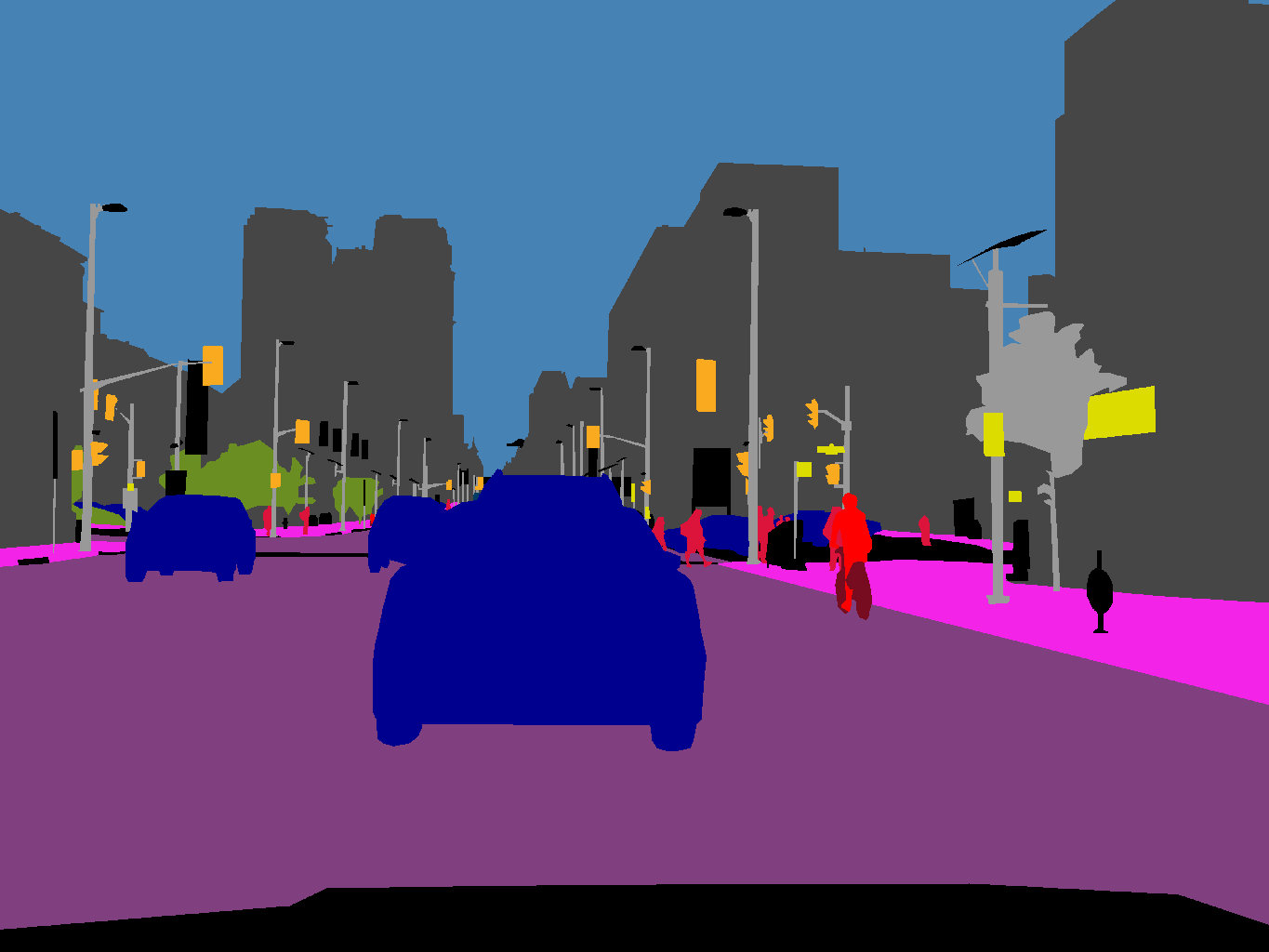}}
  \end{subfigure}
  \begin{subfigure}{0.19\textwidth}
    \raisebox{-\height}{\includegraphics[width=\textwidth, height=0.55\textwidth]{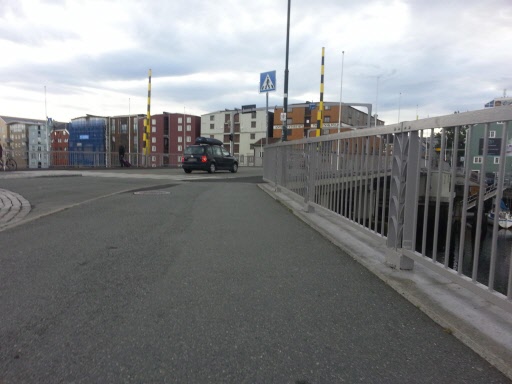}}
  \end{subfigure}
  \hfill
  \begin{subfigure}{0.19\textwidth}
    \raisebox{-\height}{\includegraphics[width=\textwidth, height=0.55\textwidth]{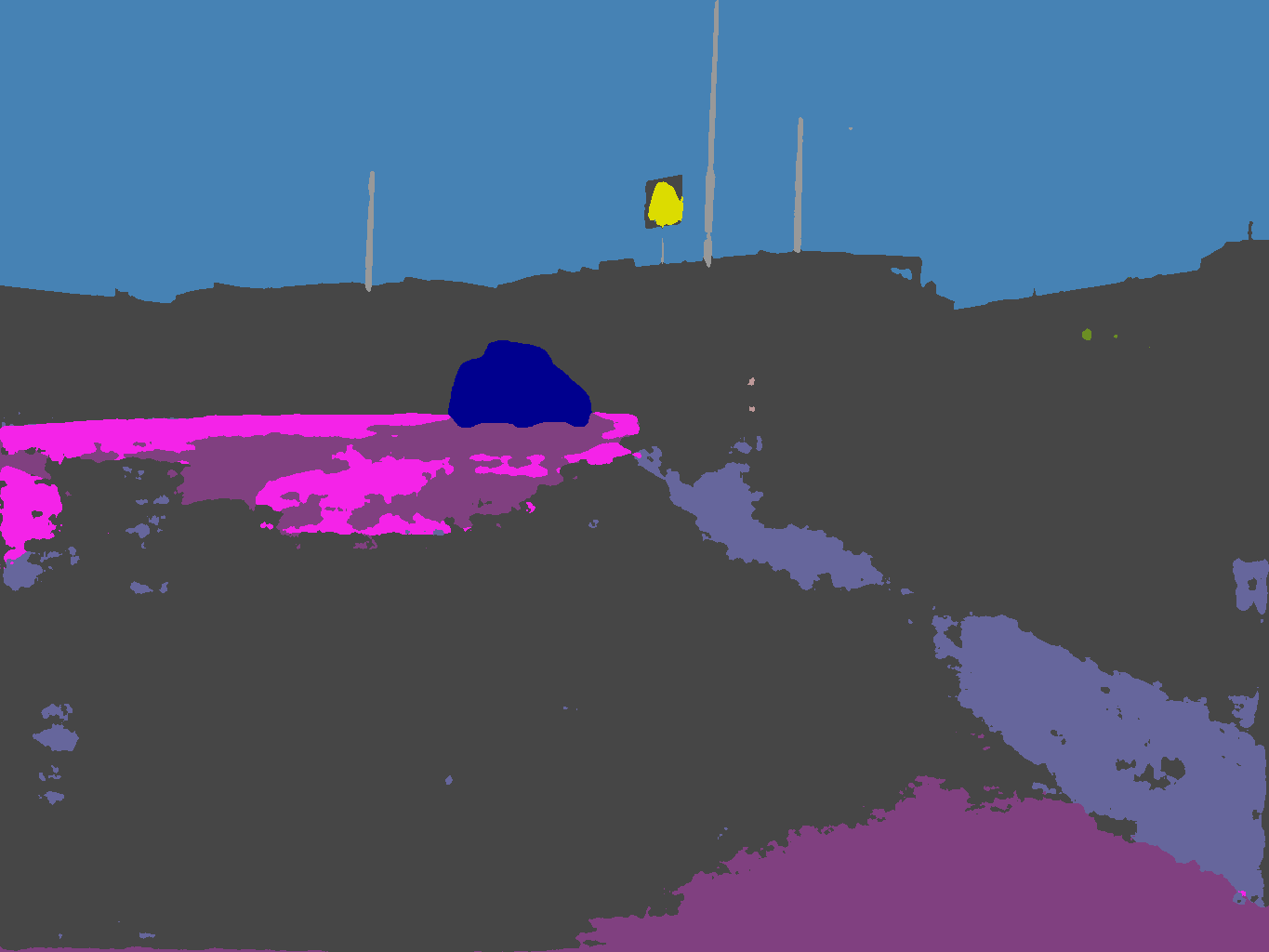}}
  \end{subfigure}
  \hfill
  \begin{subfigure}{0.19\textwidth}
    \raisebox{-\height}{\includegraphics[width=\textwidth, height=0.55\textwidth]{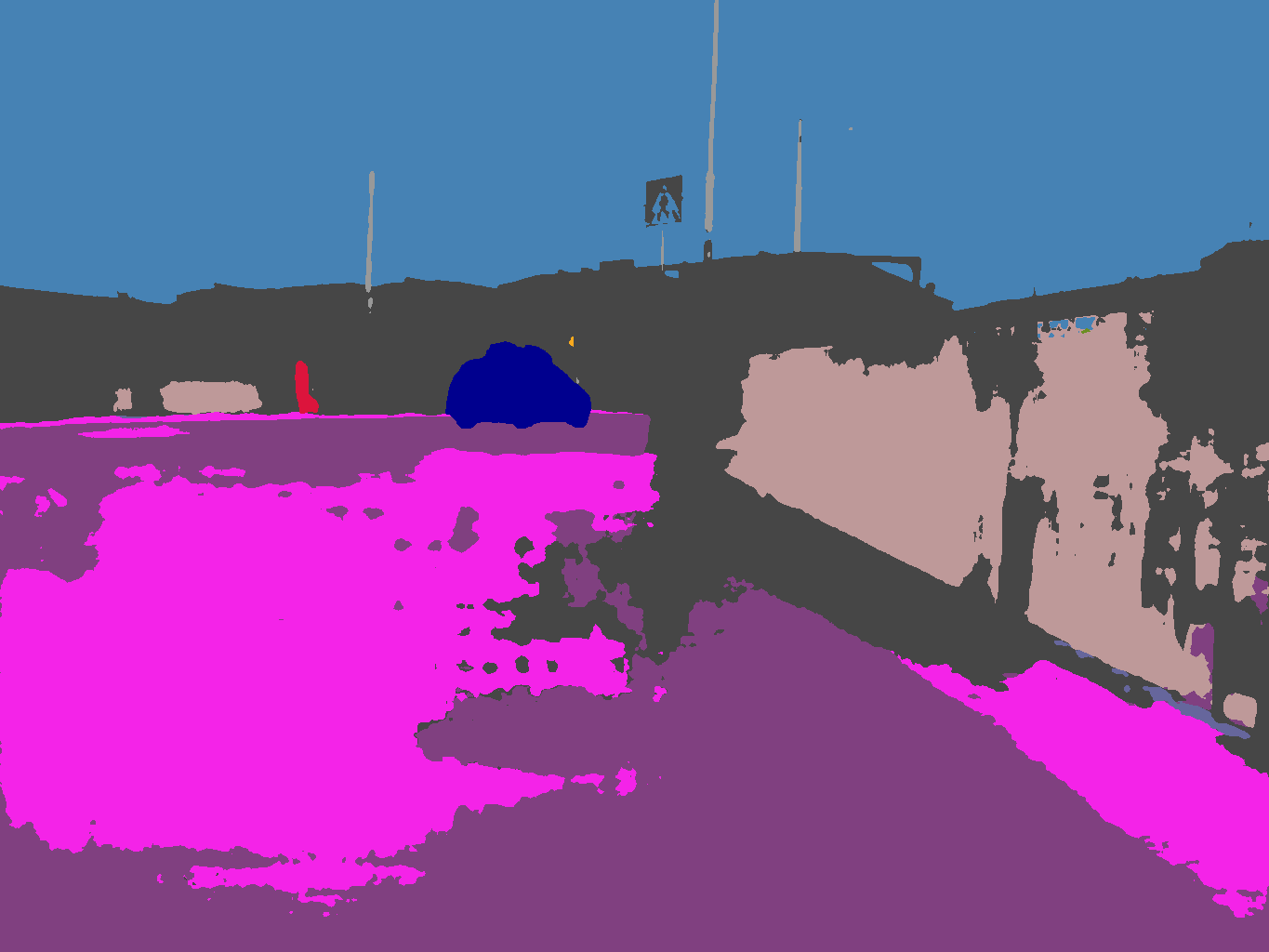}}
  \end{subfigure}
  \hfill
  \begin{subfigure}{0.19\textwidth}
    \raisebox{-\height}{\includegraphics[width=\textwidth, height=0.55\textwidth]{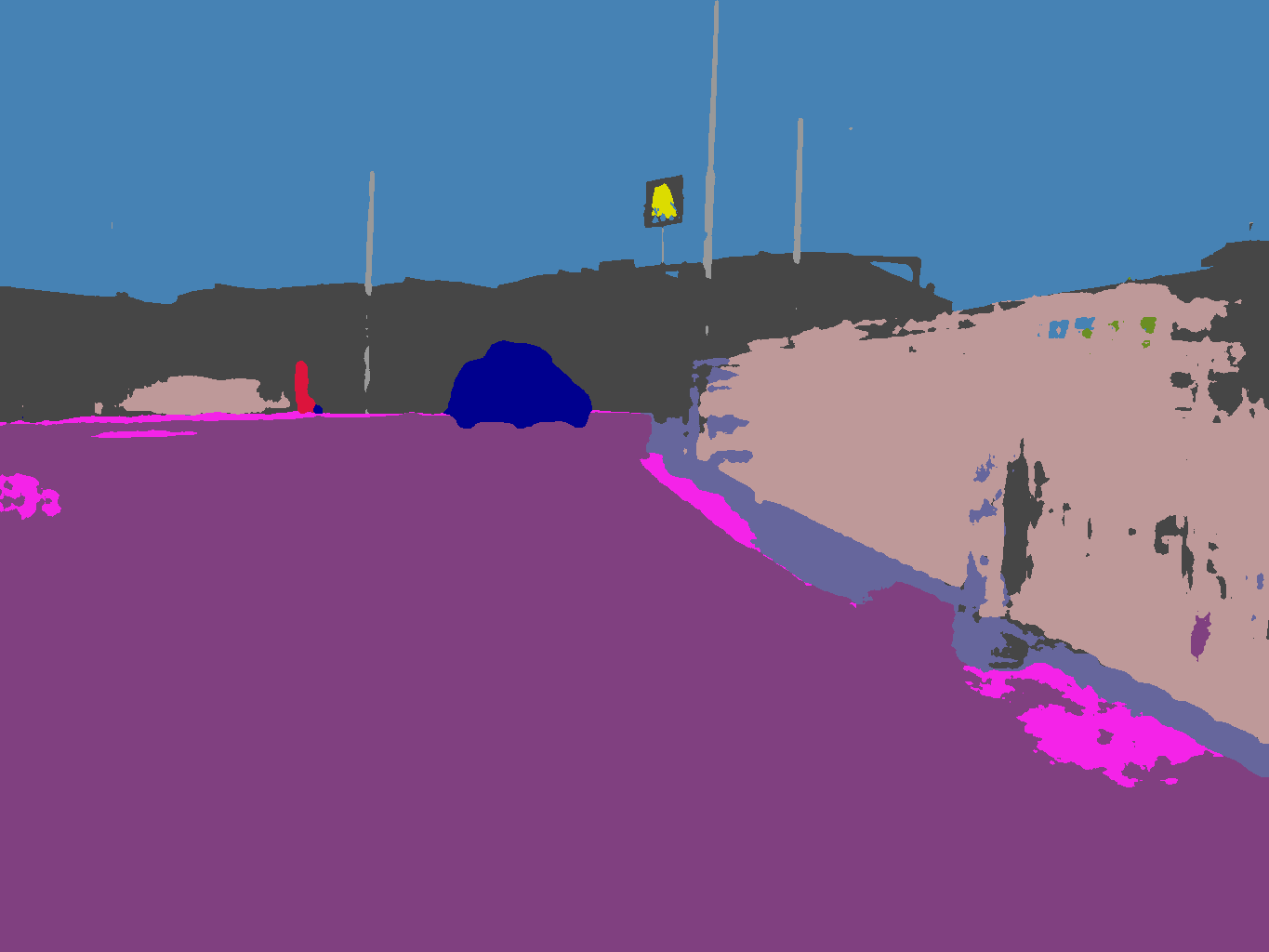}}
  \end{subfigure}
  \hfill
  \begin{subfigure}{0.19\textwidth}
    \raisebox{-\height}{\includegraphics[width=\textwidth, height=0.55\textwidth]{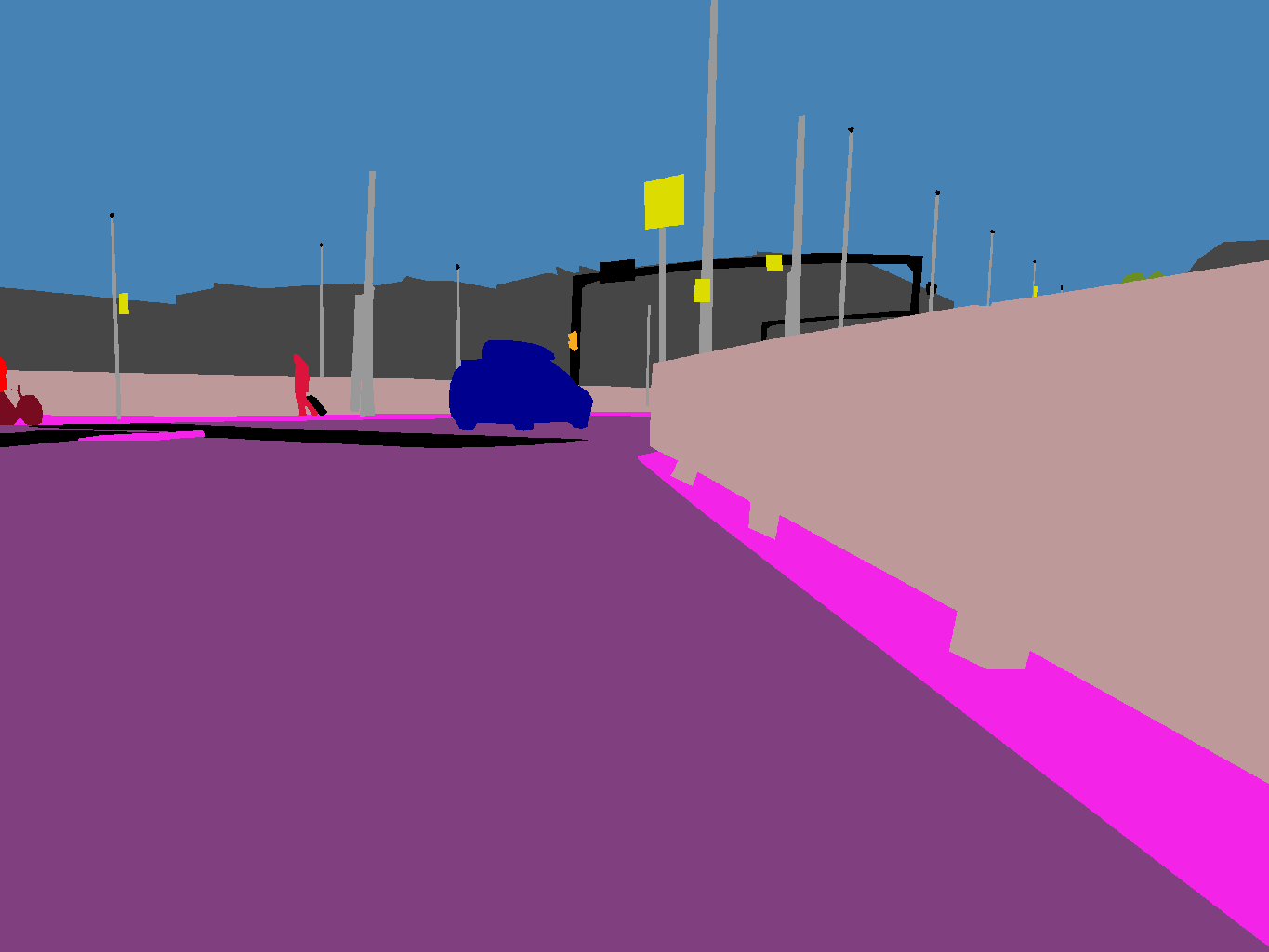}}
  \end{subfigure}
  \begin{subfigure}{0.19\textwidth}
    \raisebox{-\height}{\includegraphics[width=\textwidth, height=0.55\textwidth]{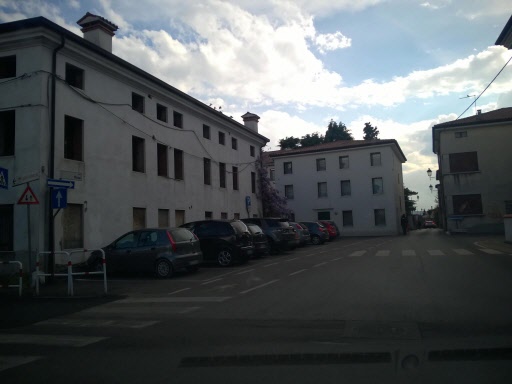}}
  \end{subfigure}
  \hfill
  \begin{subfigure}{0.19\textwidth}
    \raisebox{-\height}{\includegraphics[width=\textwidth, height=0.55\textwidth]{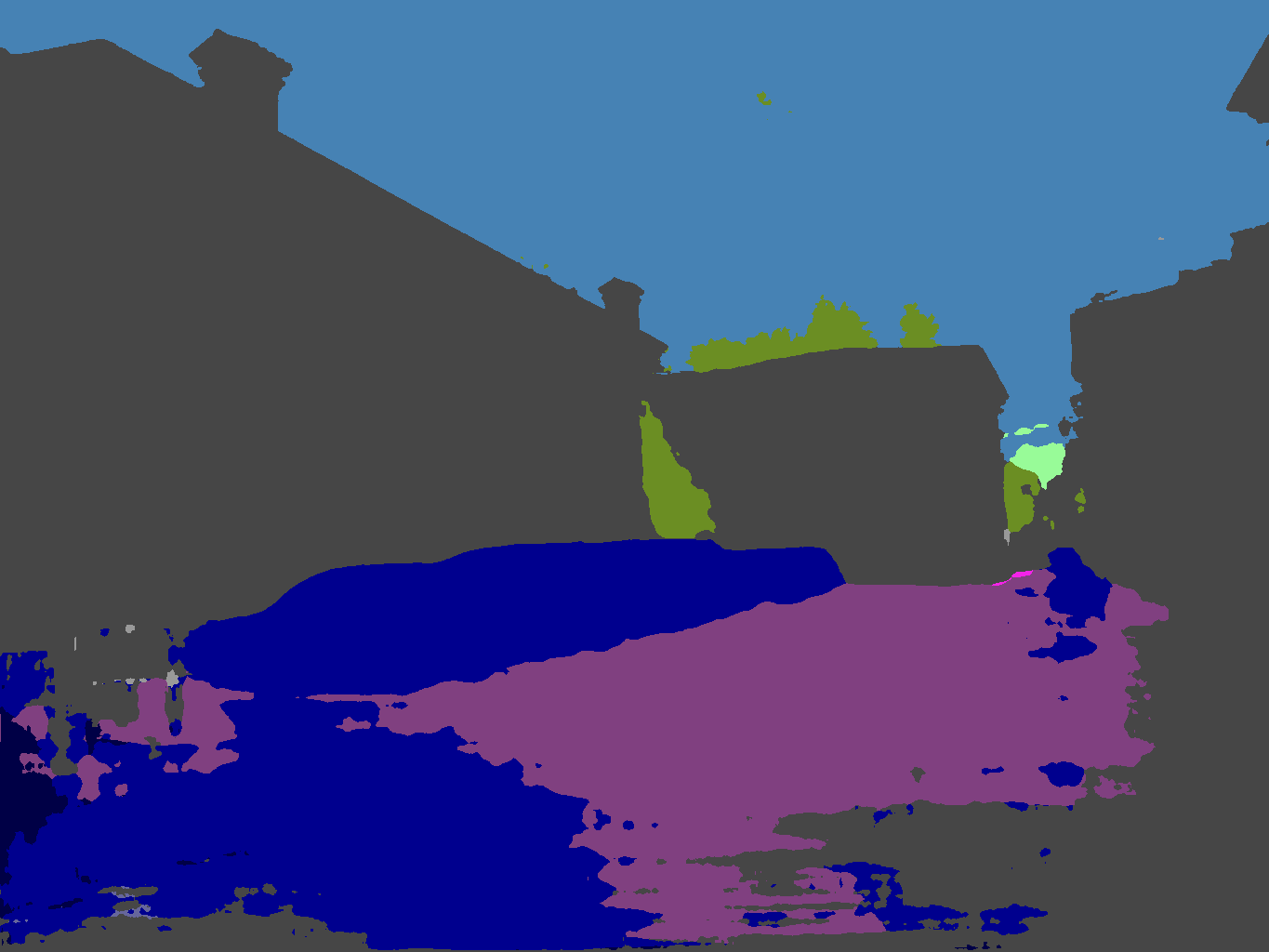}}
  \end{subfigure}
  \hfill
  \begin{subfigure}{0.19\textwidth}
    \raisebox{-\height}{\includegraphics[width=\textwidth, height=0.55\textwidth]{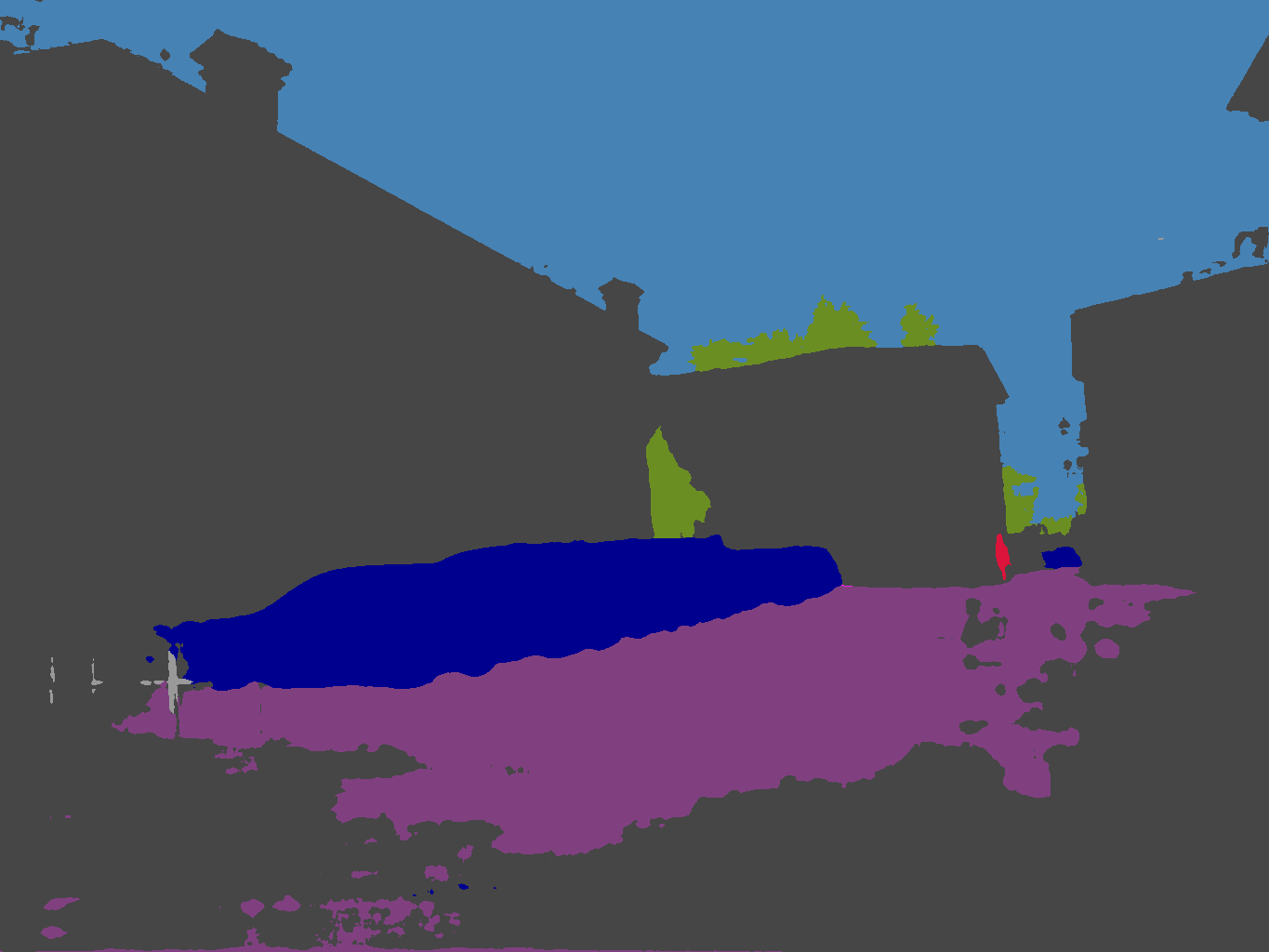}}
  \end{subfigure}
  \hfill
  \begin{subfigure}{0.19\textwidth}
    \raisebox{-\height}{\includegraphics[width=\textwidth, height=0.55\textwidth]{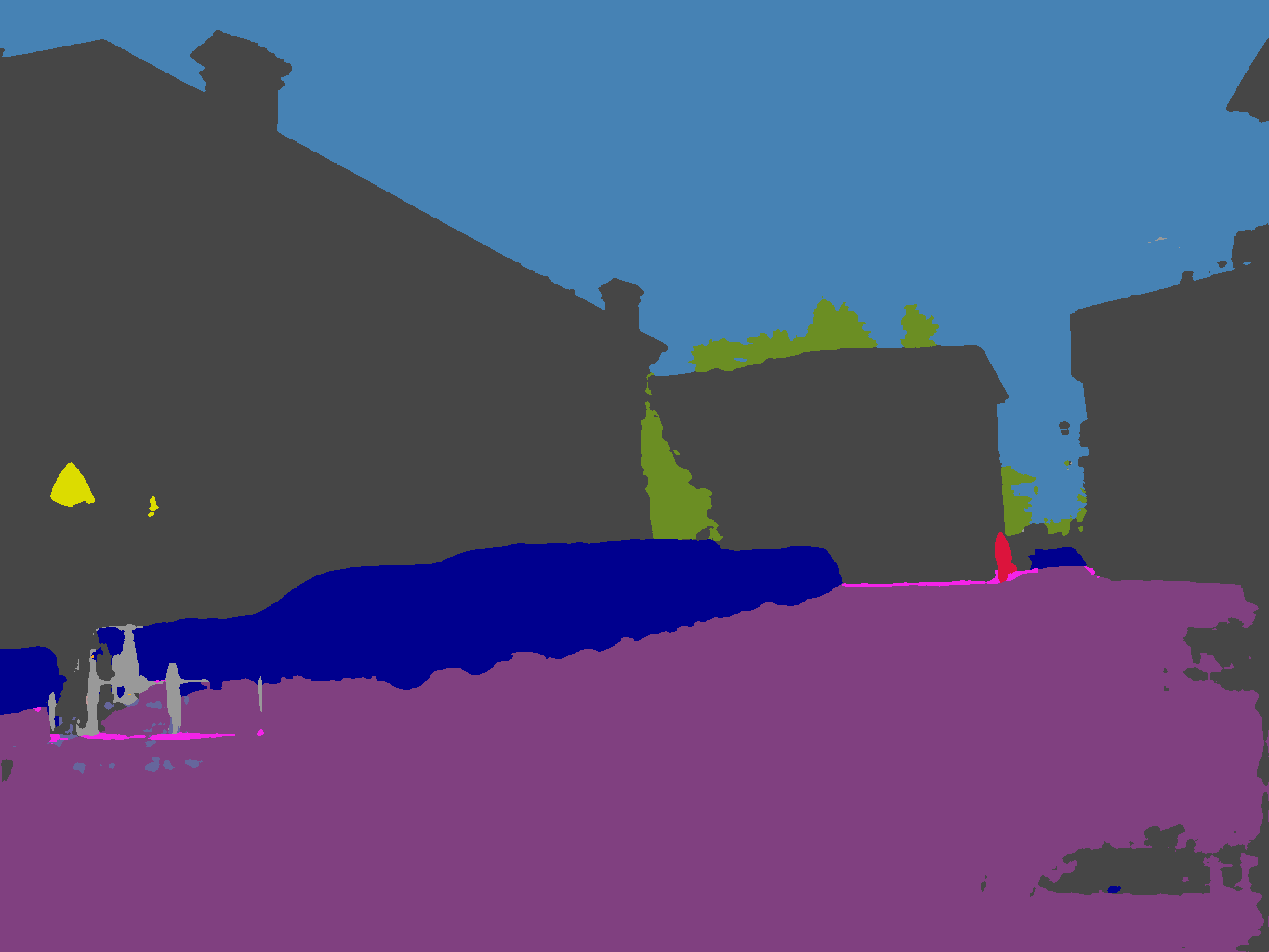}}
  \end{subfigure}
  \hfill
  \begin{subfigure}{0.19\textwidth}
    \raisebox{-\height}{\includegraphics[width=\textwidth, height=0.55\textwidth]{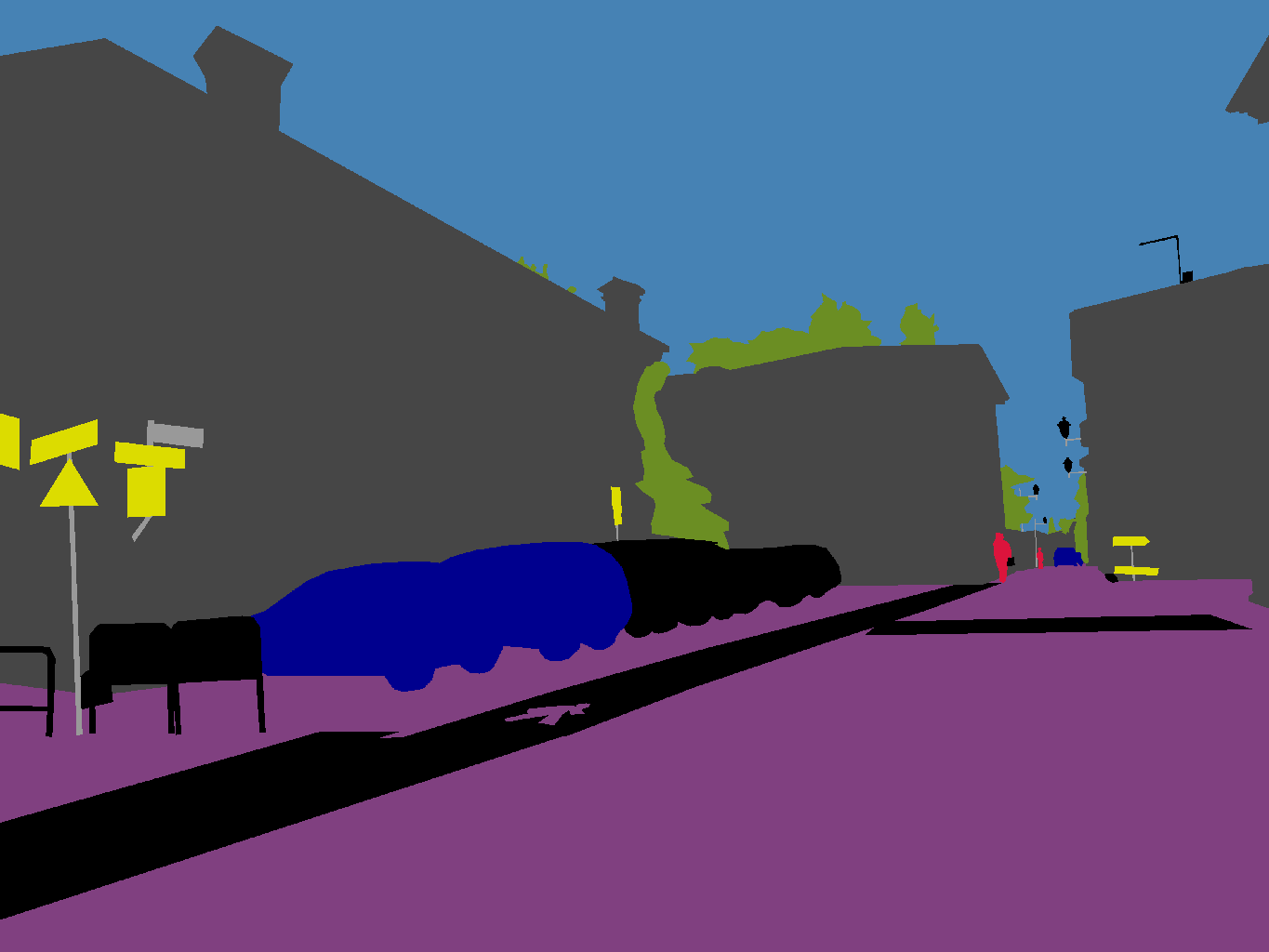}}
  \end{subfigure}
  \begin{subfigure}{0.19\textwidth}
    \raisebox{-\height}{\includegraphics[width=\textwidth, height=0.55\textwidth]{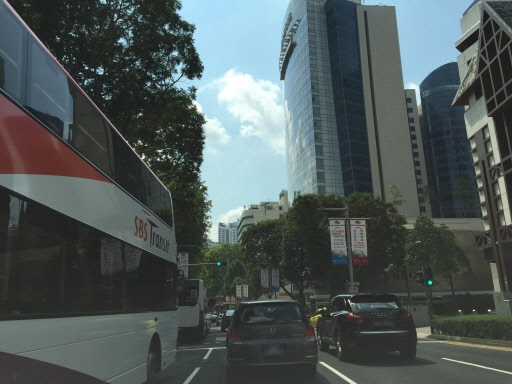}}
  \end{subfigure}
  \hfill
  \begin{subfigure}{0.19\textwidth}
    \raisebox{-\height}{\includegraphics[width=\textwidth, height=0.55\textwidth]{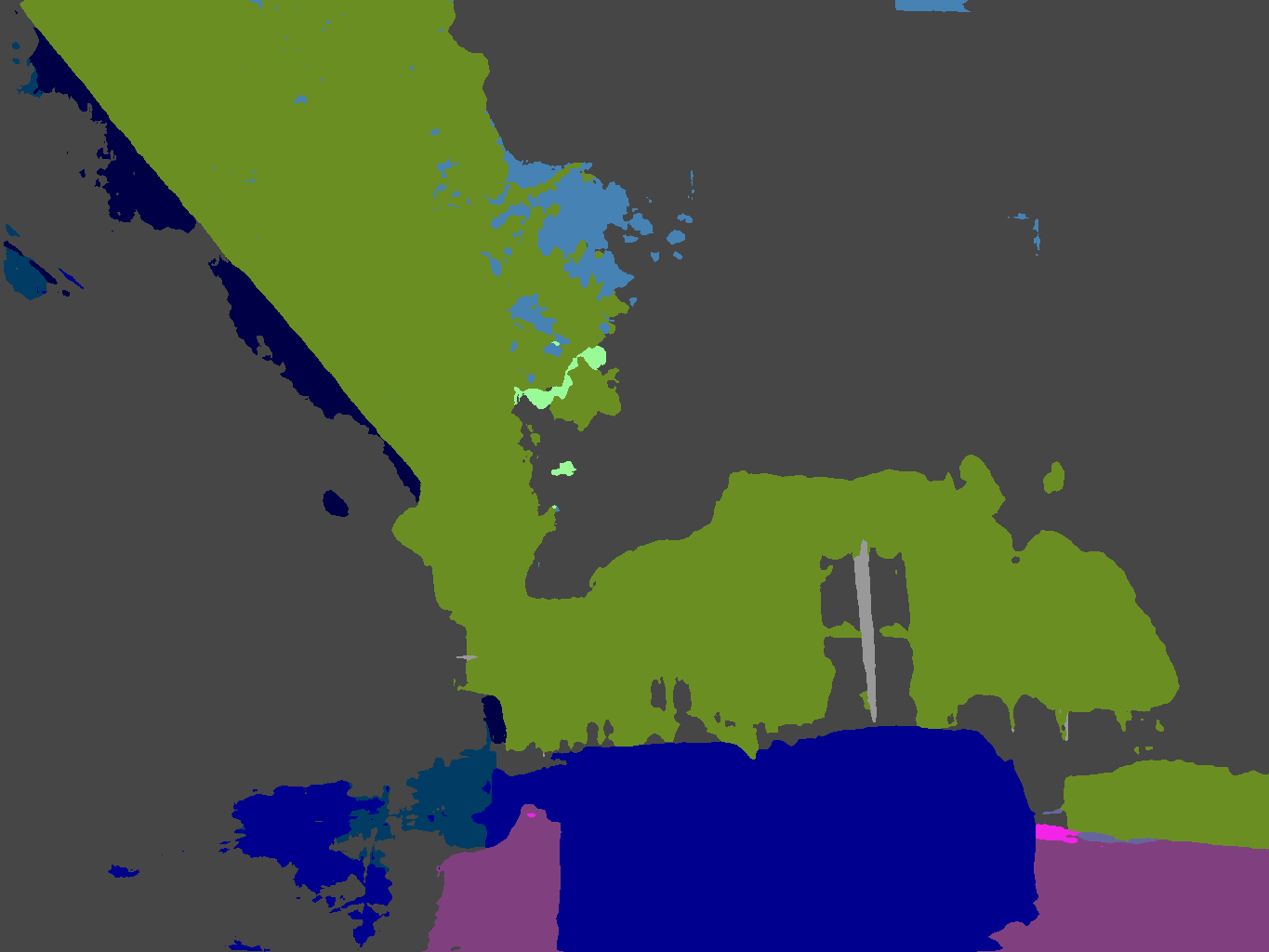}}
  \end{subfigure}
  \hfill
  \begin{subfigure}{0.19\textwidth}
    \raisebox{-\height}{\includegraphics[width=\textwidth, height=0.55\textwidth]{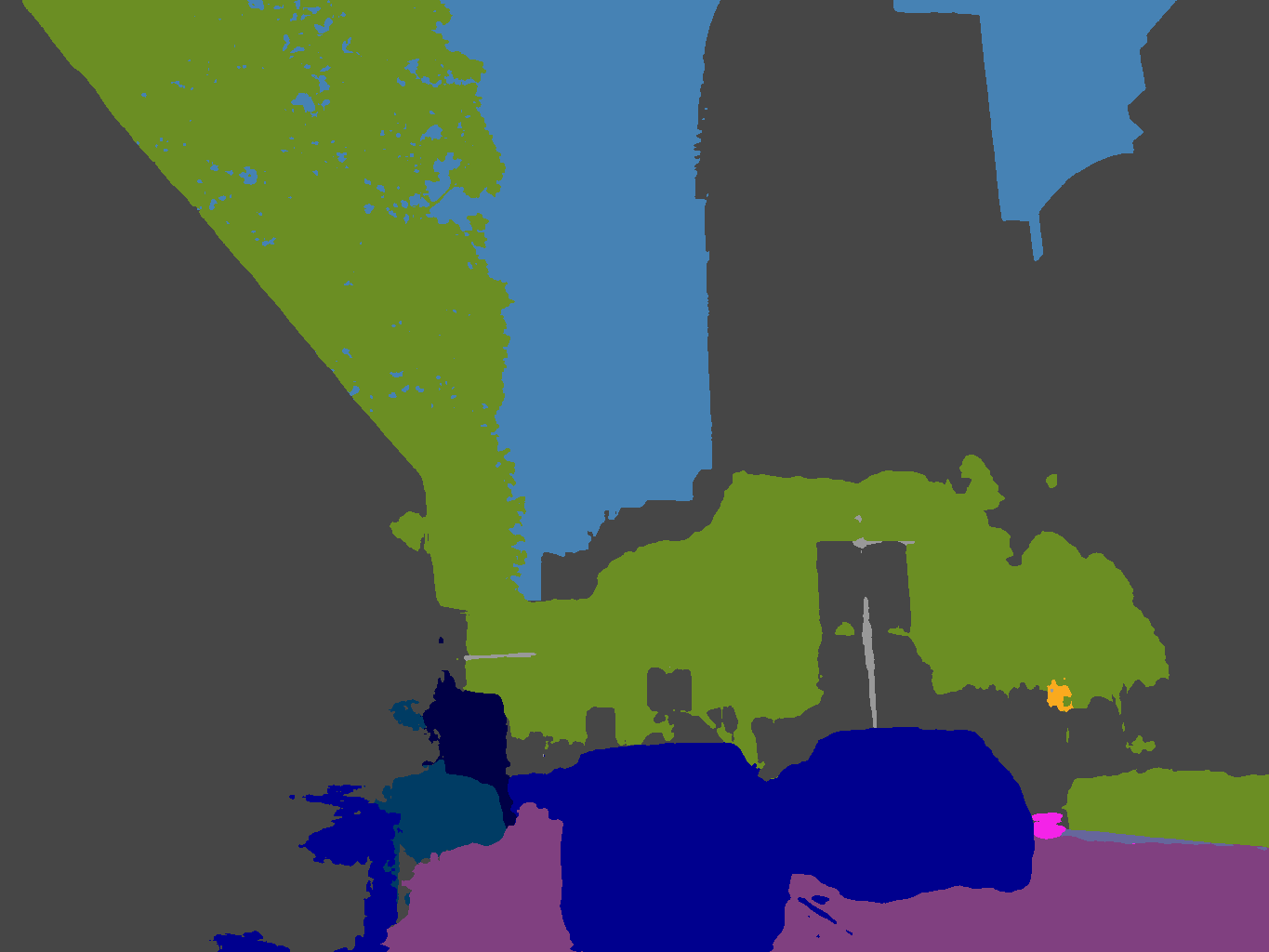}}
  \end{subfigure}
  \hfill
  \begin{subfigure}{0.19\textwidth}
    \raisebox{-\height}{\includegraphics[width=\textwidth, height=0.55\textwidth]{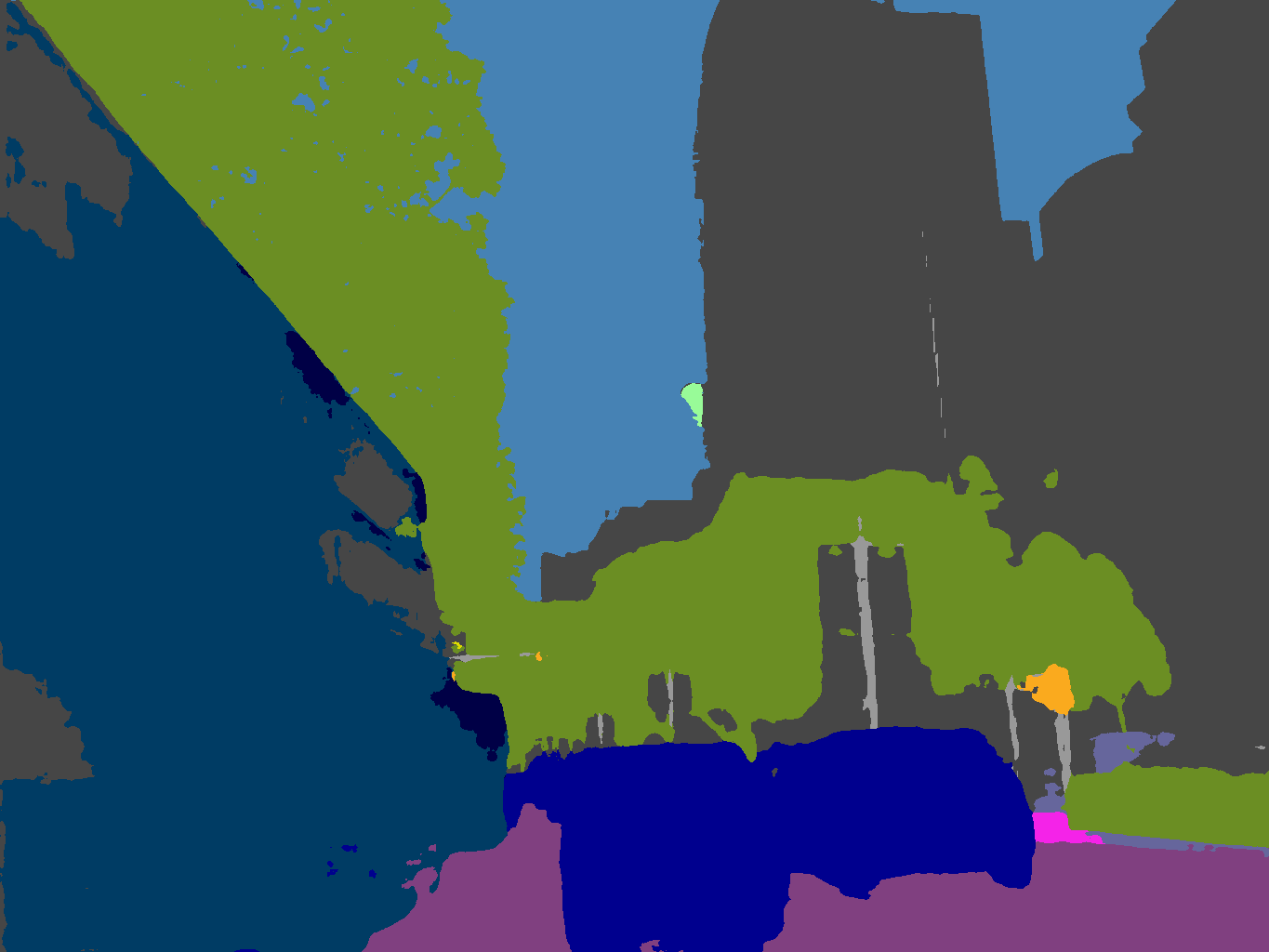}}
  \end{subfigure}
  \hfill
  \begin{subfigure}{0.19\textwidth}
    \raisebox{-\height}{\includegraphics[width=\textwidth, height=0.55\textwidth]{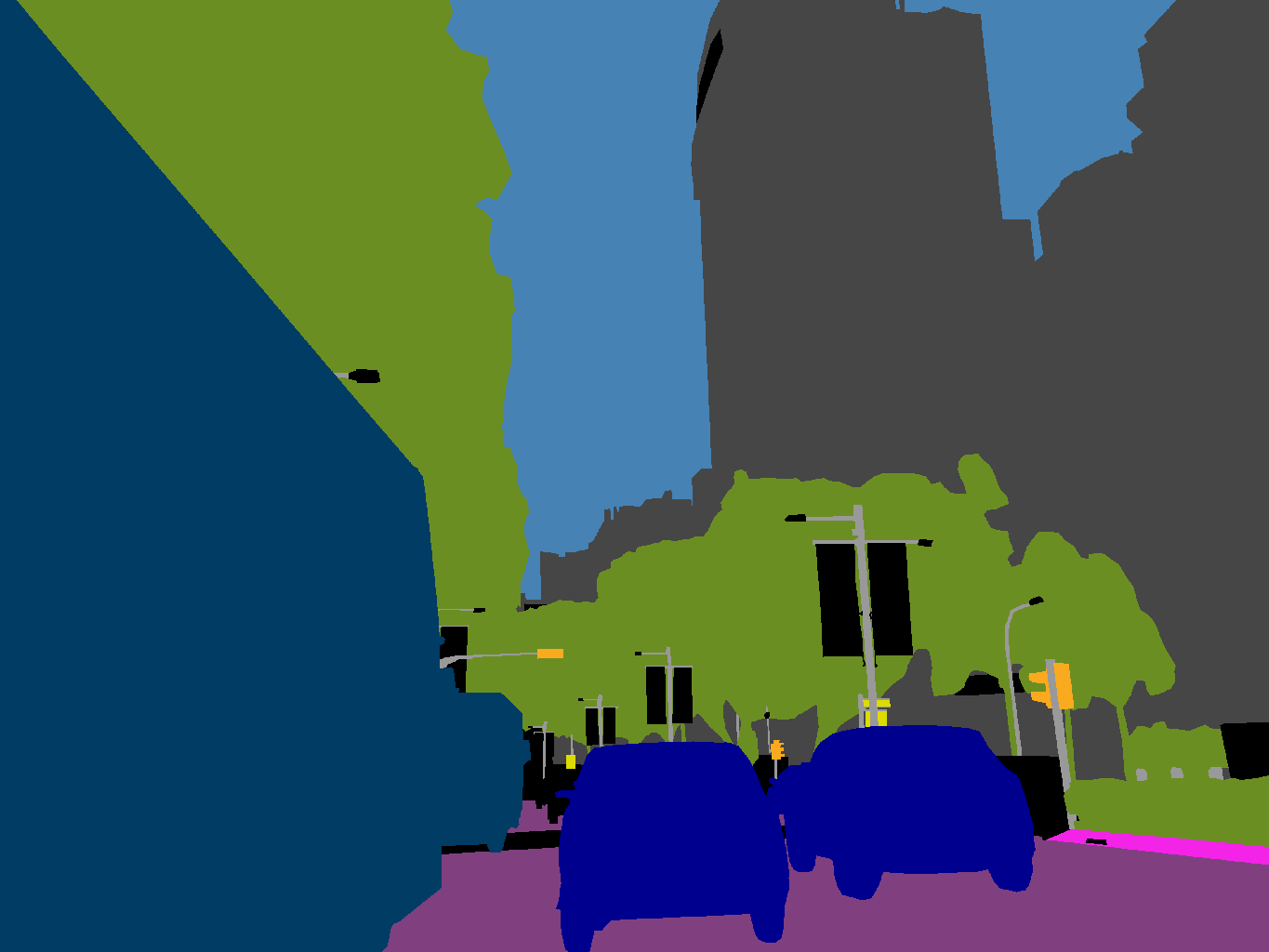}}
  \end{subfigure}
  \begin{subfigure}{0.19\textwidth}
    \raisebox{-\height}{\includegraphics[width=\textwidth, height=0.55\textwidth]{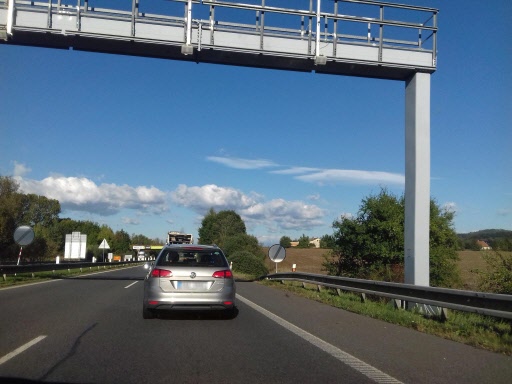}}
    \caption*{Unseen domain image}
    \label{fig:compare_r50_map_img}
  \end{subfigure}
  \hfill
  \begin{subfigure}{0.19\textwidth}
    \raisebox{-\height}{\includegraphics[width=\textwidth, height=0.55\textwidth]{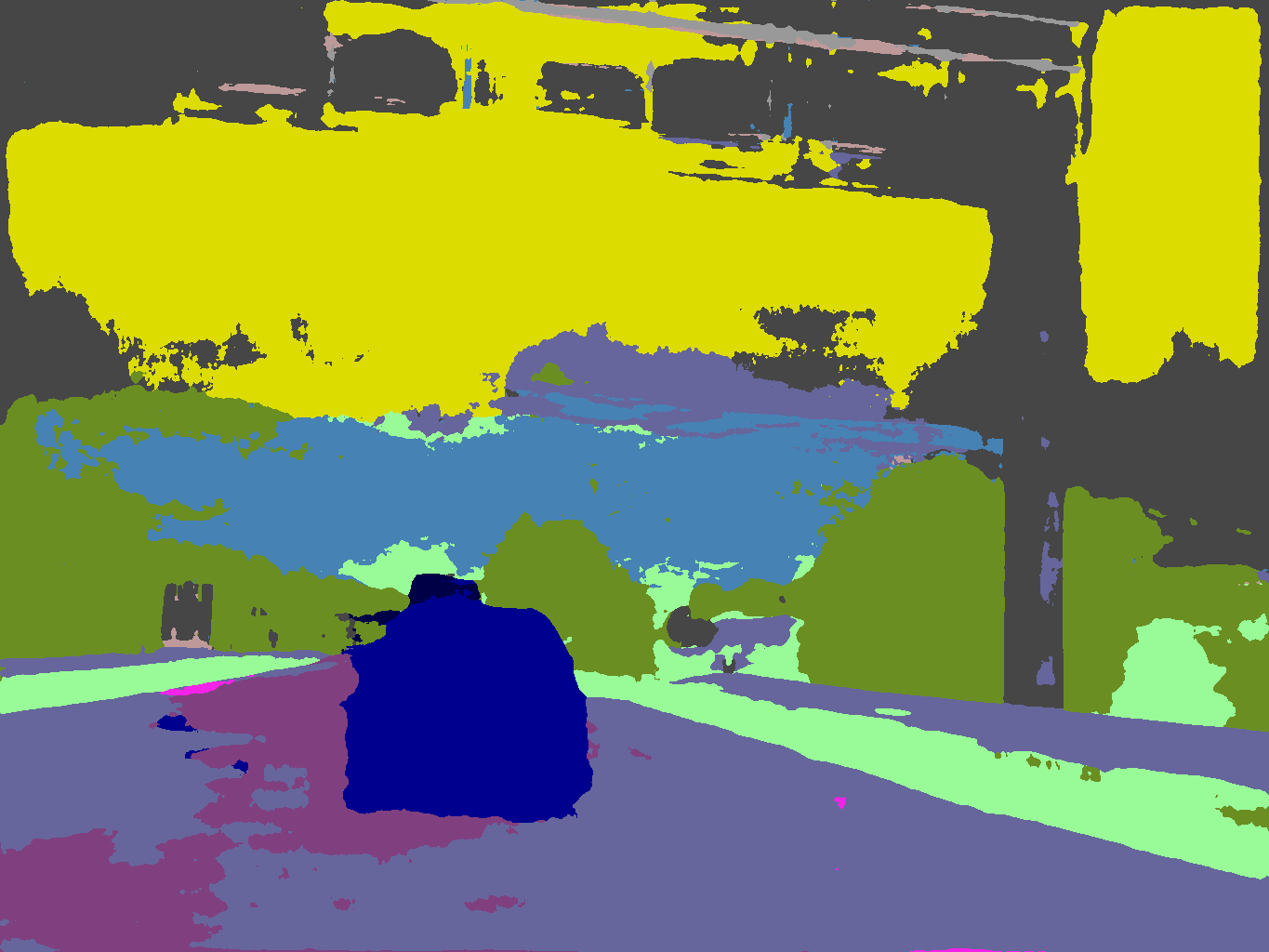}}
    \caption*{Baseline}
    \label{fig:compare_r50_map_base}
  \end{subfigure}
  \hfill
  \begin{subfigure}{0.19\textwidth}
    \raisebox{-\height}{\includegraphics[width=\textwidth, height=0.55\textwidth]{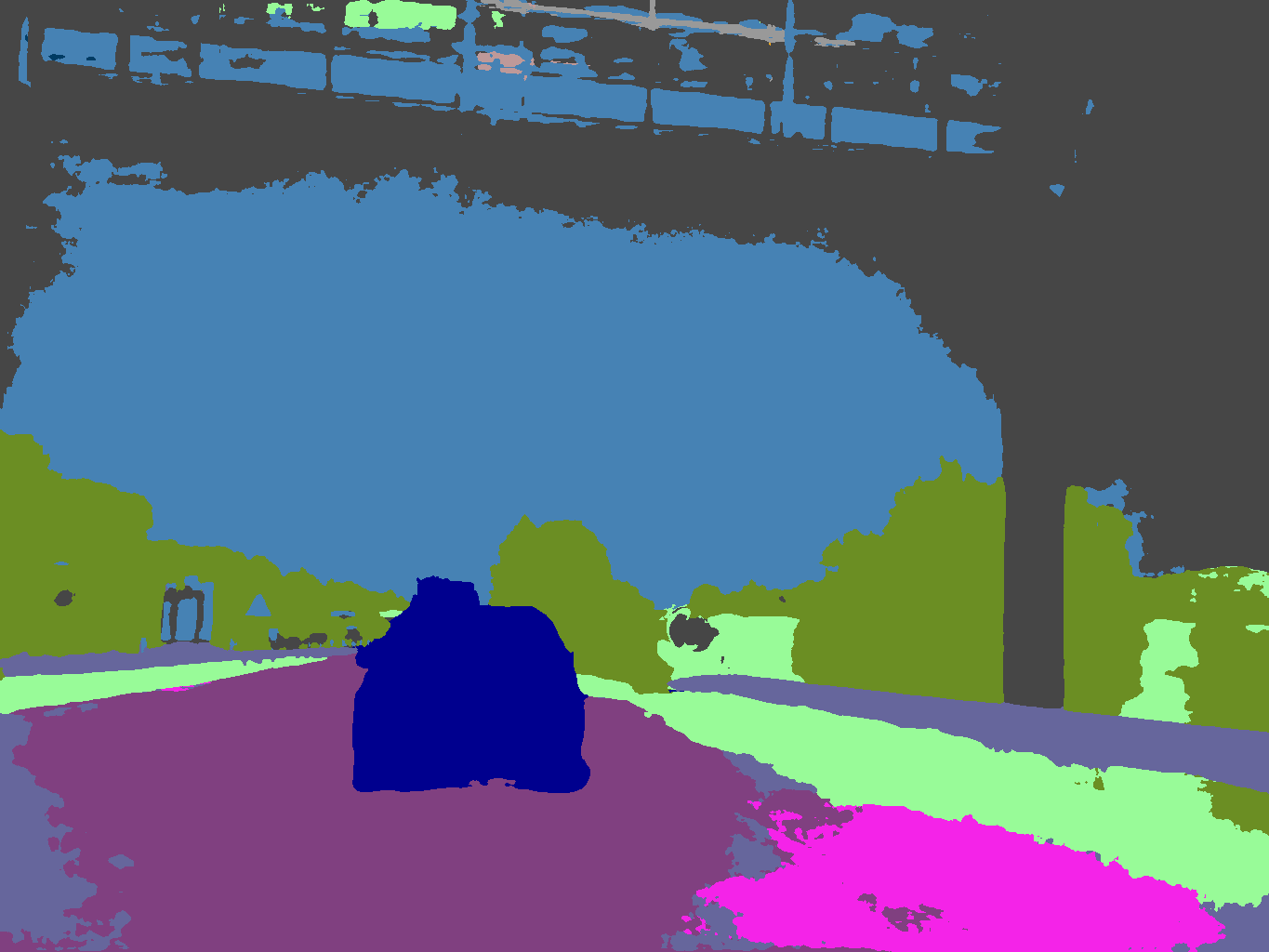}}
    \caption*{RobustNet}
    \label{fig:compare_r50_map_isw}
  \end{subfigure}
  \hfill
  \begin{subfigure}{0.19\textwidth}
    \raisebox{-\height}{\includegraphics[width=\textwidth, height=0.55\textwidth]{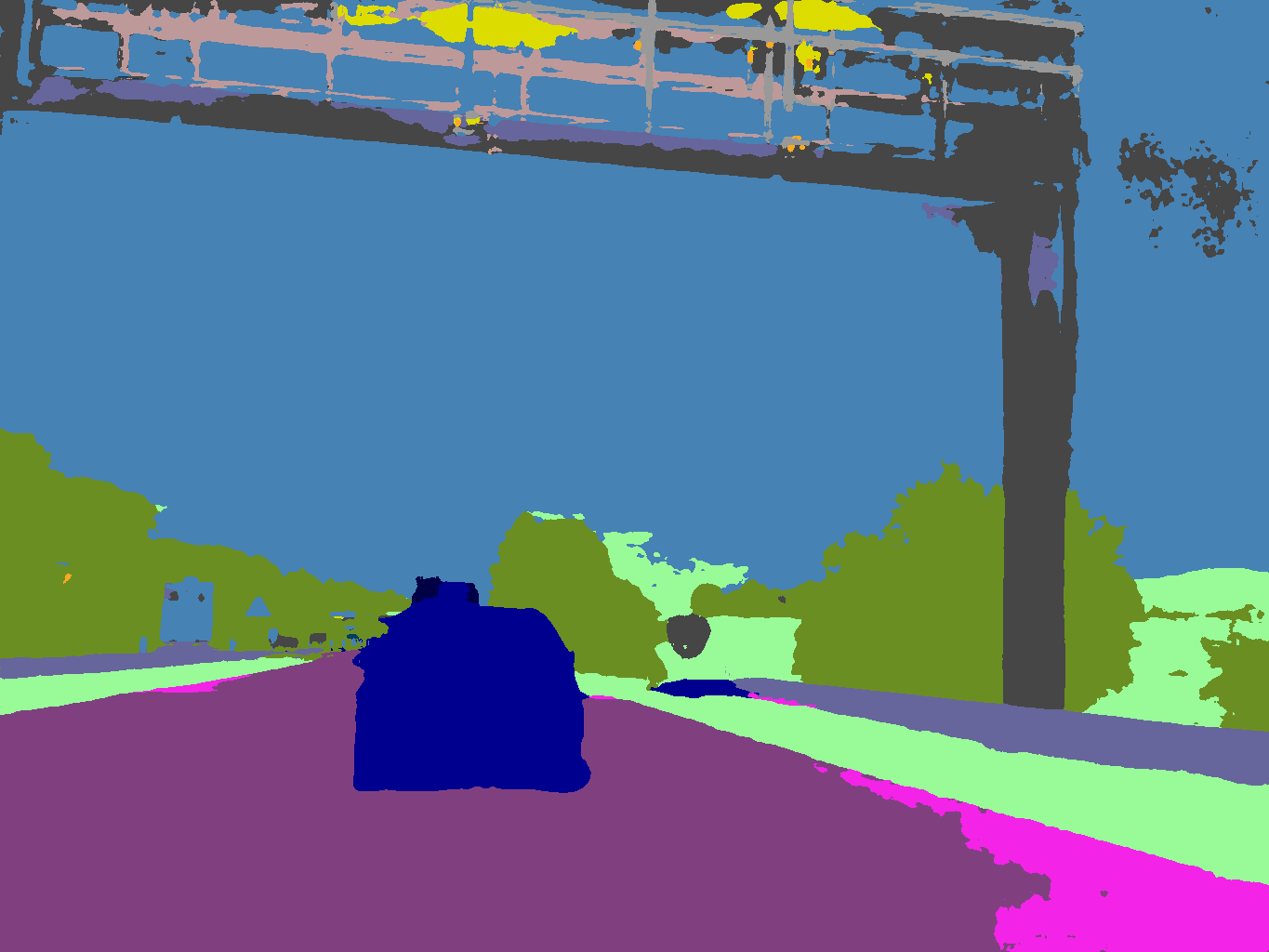}}
    \caption*{\textbf{Ours (WildNet)}}
    \label{fig:compare_r50_map_ours}
  \end{subfigure}
  \hfill
  \begin{subfigure}{0.19\textwidth}
    \raisebox{-\height}{\includegraphics[width=\textwidth, height=0.55\textwidth]{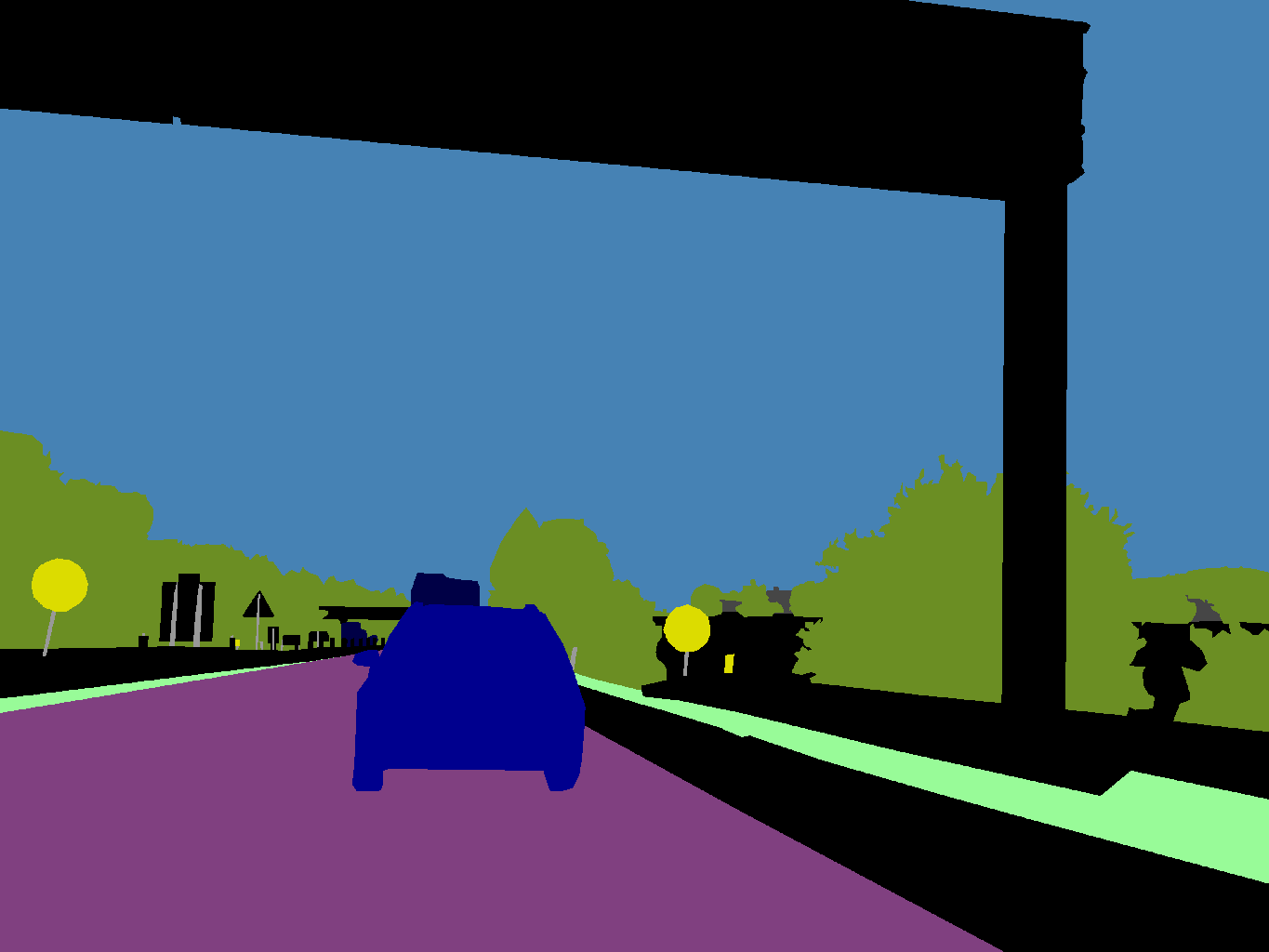}}
    \caption*{Ground truth}
    \label{fig:compare_r50_map_gt}
  \end{subfigure}
  \caption{Semantic segmentation results on unseen domain images in Mapillary with the models trained on GTAV.
  }
  \label{fig:compare_r50_map_img_base_isw_ours}
\end{figure*}

\begin{figure*}
  \centering
  \begin{subfigure}{0.19\textwidth}
    \raisebox{-\height}{\includegraphics[width=\textwidth, height=0.55\textwidth]{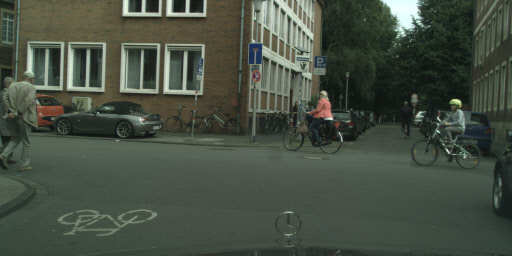}}
  \end{subfigure}
  \hfill
  \begin{subfigure}{0.19\textwidth}
    \raisebox{-\height}{\includegraphics[width=\textwidth, height=0.55\textwidth]{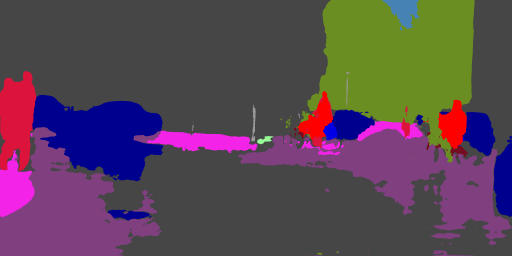}}
  \end{subfigure}
  \hfill
  \begin{subfigure}{0.19\textwidth}
    \raisebox{-\height}{\includegraphics[width=\textwidth, height=0.55\textwidth]{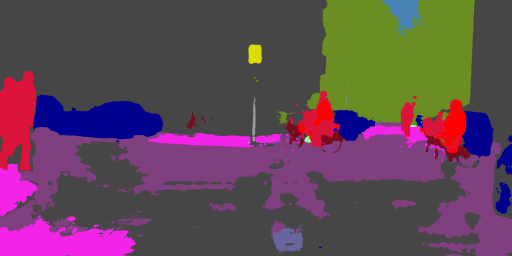}}
  \end{subfigure}
  \hfill
  \begin{subfigure}{0.19\textwidth}
    \raisebox{-\height}{\includegraphics[width=\textwidth, height=0.55\textwidth]{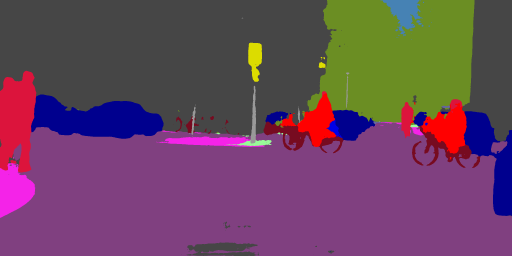}}
  \end{subfigure}
  \hfill
  \begin{subfigure}{0.19\textwidth}
    \raisebox{-\height}{\includegraphics[width=\textwidth, height=0.55\textwidth]{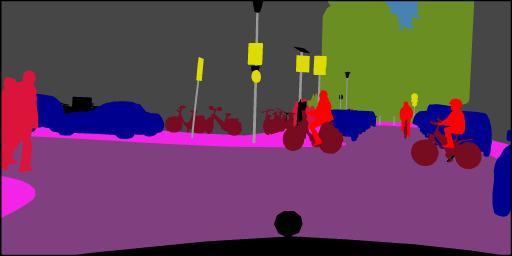}}
  \end{subfigure}
  \begin{subfigure}{0.19\textwidth}
    \raisebox{-\height}{\includegraphics[width=\textwidth, height=0.55\textwidth]{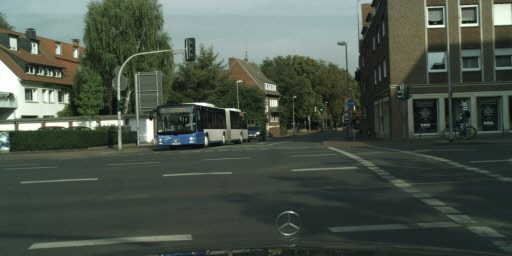}}
  \end{subfigure}
  \hfill
  \begin{subfigure}{0.19\textwidth}
    \raisebox{-\height}{\includegraphics[width=\textwidth, height=0.55\textwidth]{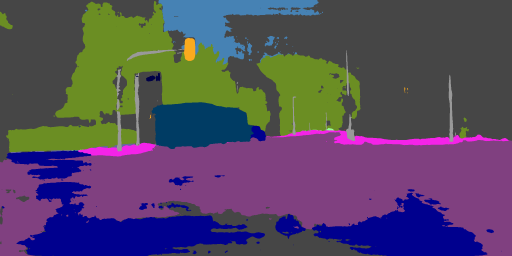}}
  \end{subfigure}
  \hfill
  \begin{subfigure}{0.19\textwidth}
    \raisebox{-\height}{\includegraphics[width=\textwidth, height=0.55\textwidth]{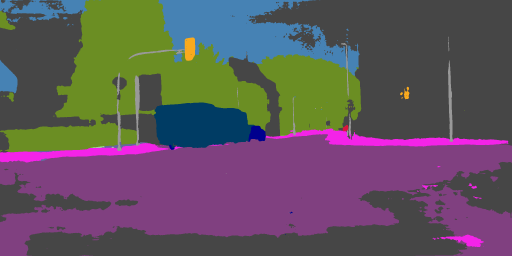}}
  \end{subfigure}
  \hfill
  \begin{subfigure}{0.19\textwidth}
    \raisebox{-\height}{\includegraphics[width=\textwidth, height=0.55\textwidth]{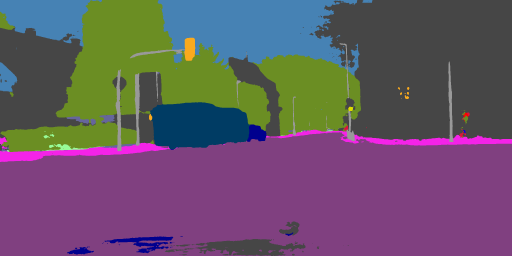}}
  \end{subfigure}
  \hfill
  \begin{subfigure}{0.19\textwidth}
    \raisebox{-\height}{\includegraphics[width=\textwidth, height=0.55\textwidth]{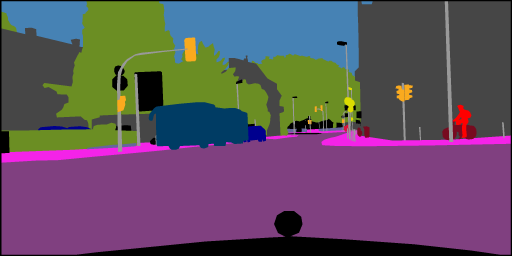}}
  \end{subfigure}
  \begin{subfigure}{0.19\textwidth}
    \raisebox{-\height}{\includegraphics[width=\textwidth, height=0.55\textwidth]{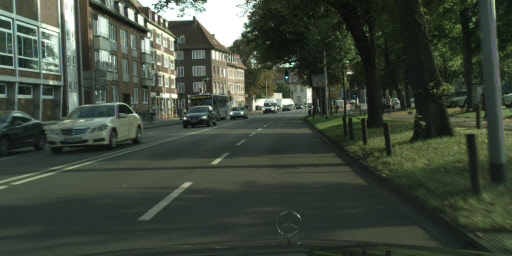}}
  \end{subfigure}
  \hfill
  \begin{subfigure}{0.19\textwidth}
    \raisebox{-\height}{\includegraphics[width=\textwidth, height=0.55\textwidth]{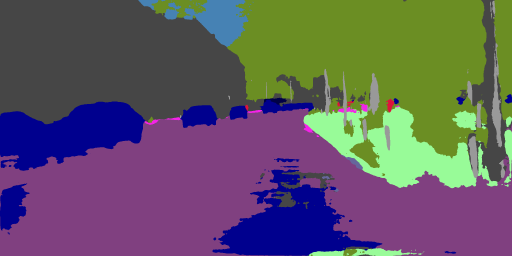}}
  \end{subfigure}
  \hfill
  \begin{subfigure}{0.19\textwidth}
    \raisebox{-\height}{\includegraphics[width=\textwidth, height=0.55\textwidth]{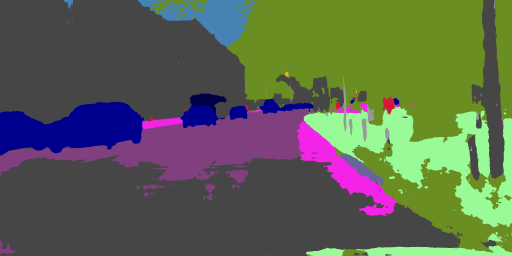}}
  \end{subfigure}
  \hfill
  \begin{subfigure}{0.19\textwidth}
    \raisebox{-\height}{\includegraphics[width=\textwidth, height=0.55\textwidth]{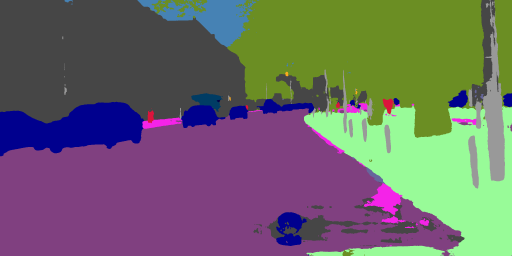}}
  \end{subfigure}
  \hfill
  \begin{subfigure}{0.19\textwidth}
    \raisebox{-\height}{\includegraphics[width=\textwidth, height=0.55\textwidth]{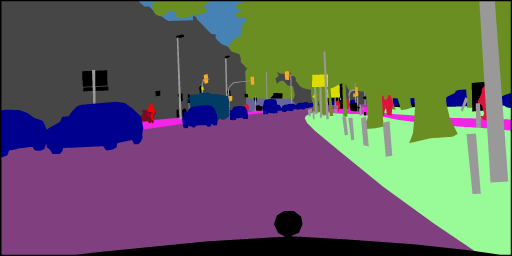}}
  \end{subfigure}
  \begin{subfigure}{0.19\textwidth}
    \raisebox{-\height}{\includegraphics[width=\textwidth, height=0.55\textwidth]{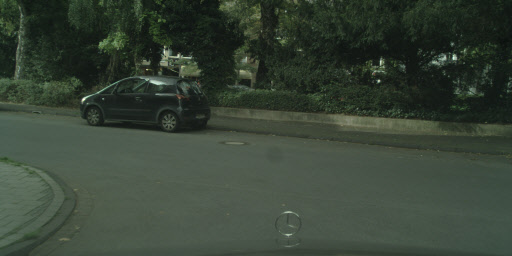}}
  \end{subfigure}
  \hfill
  \begin{subfigure}{0.19\textwidth}
    \raisebox{-\height}{\includegraphics[width=\textwidth, height=0.55\textwidth]{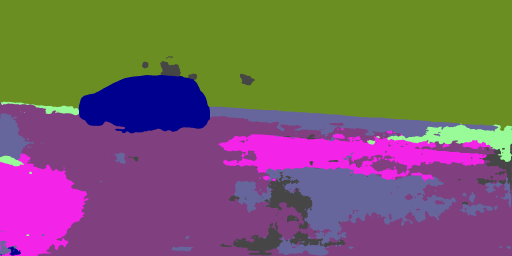}}
  \end{subfigure}
  \hfill
  \begin{subfigure}{0.19\textwidth}
    \raisebox{-\height}{\includegraphics[width=\textwidth, height=0.55\textwidth]{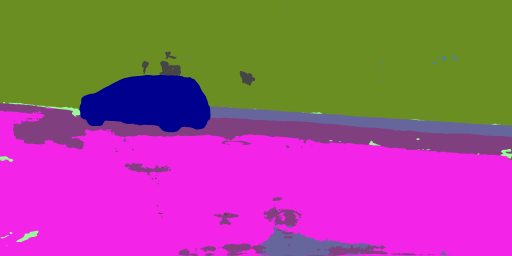}}
  \end{subfigure}
  \hfill
  \begin{subfigure}{0.19\textwidth}
    \raisebox{-\height}{\includegraphics[width=\textwidth, height=0.55\textwidth]{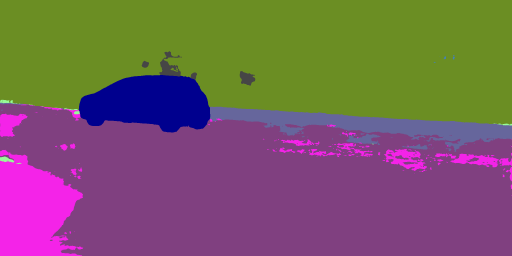}}
  \end{subfigure}
  \hfill
  \begin{subfigure}{0.19\textwidth}
    \raisebox{-\height}{\includegraphics[width=\textwidth, height=0.55\textwidth]{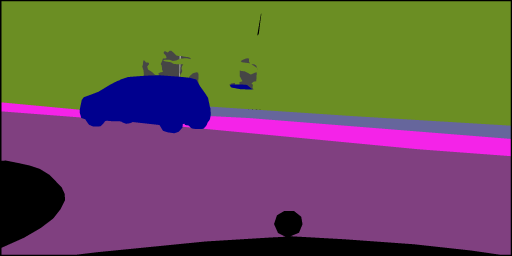}}
  \end{subfigure}
  \begin{subfigure}{0.19\textwidth}
    \raisebox{-\height}{\includegraphics[width=\textwidth, height=0.55\textwidth]{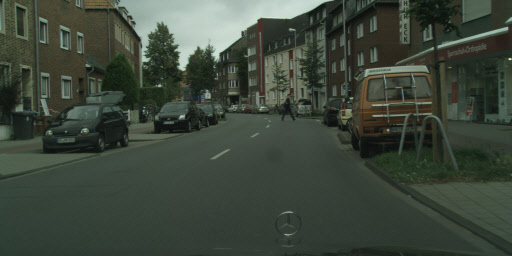}}
    \caption*{Unseen domain image}
    \label{fig:compare_r50_cty_img}
  \end{subfigure}
  \hfill
  \begin{subfigure}{0.19\textwidth}
    \raisebox{-\height}{\includegraphics[width=\textwidth, height=0.55\textwidth]{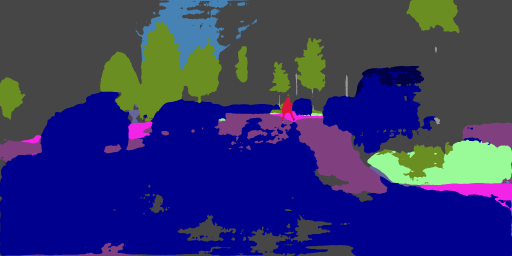}}
    \caption*{Baseline}
    \label{fig:compare_r50_cty_base}
  \end{subfigure}
  \hfill
  \begin{subfigure}{0.19\textwidth}
    \raisebox{-\height}{\includegraphics[width=\textwidth, height=0.55\textwidth]{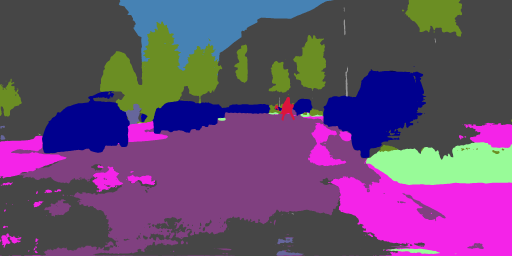}}
    \caption*{RobustNet}
    \label{fig:compare_r50_cty_isw}
  \end{subfigure}
  \hfill
  \begin{subfigure}{0.19\textwidth}
    \raisebox{-\height}{\includegraphics[width=\textwidth, height=0.55\textwidth]{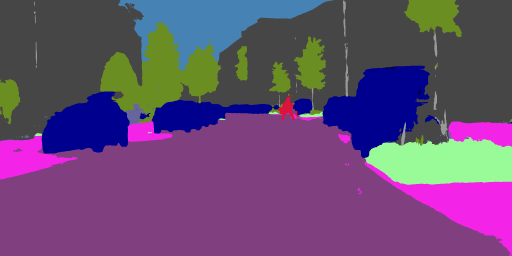}}
    \caption*{\textbf{Ours (WildNet)}}
    \label{fig:compare_r50_cty_ours}
  \end{subfigure}
  \hfill
  \begin{subfigure}{0.19\textwidth}
    \raisebox{-\height}{\includegraphics[width=\textwidth, height=0.55\textwidth]{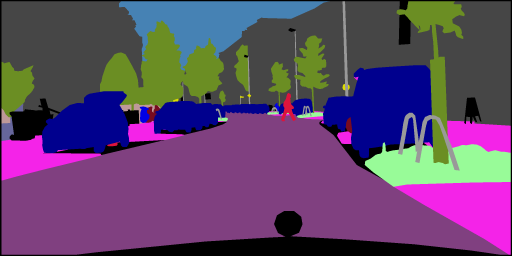}}
    \caption*{Ground truth}
    \label{fig:compare_r50_cty_gt}
  \end{subfigure}
  \caption{Semantic segmentation results on unseen domain images in Cityscapes with the models trained on GTAV.
  }
  \vspace{+1.0em}
  \label{fig:compare_r50_cty_img_base_isw_ours}
\end{figure*}

\begin{figure*}
  \centering
  \begin{subfigure}{0.19\textwidth}
    \raisebox{-\height}{\includegraphics[width=\textwidth, height=0.55\textwidth]{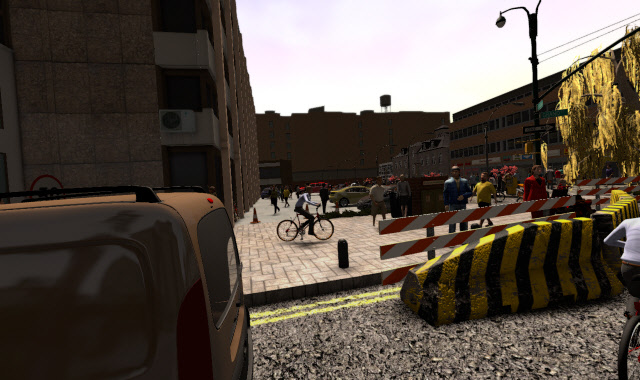}}
  \end{subfigure}
  \hfill
  \begin{subfigure}{0.19\textwidth}
    \raisebox{-\height}{\includegraphics[width=\textwidth, height=0.55\textwidth]{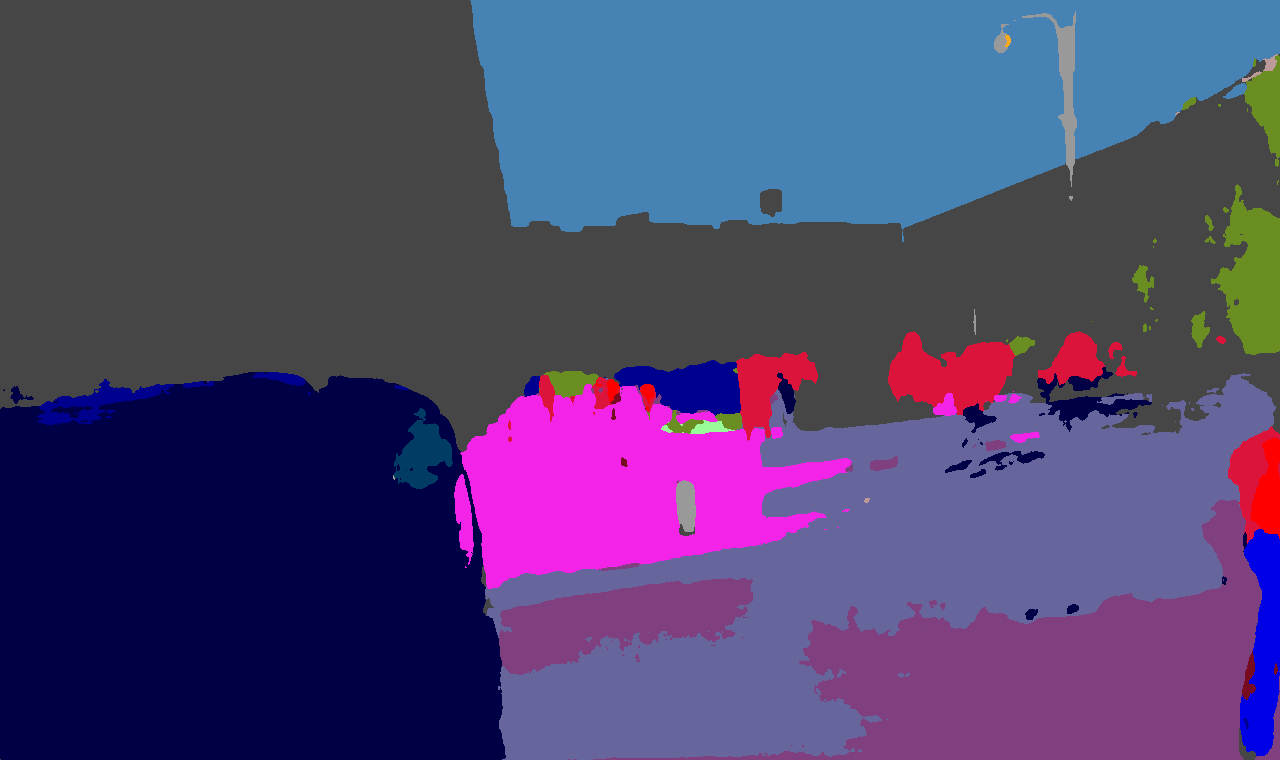}}
  \end{subfigure}
  \hfill
  \begin{subfigure}{0.19\textwidth}
    \raisebox{-\height}{\includegraphics[width=\textwidth, height=0.55\textwidth]{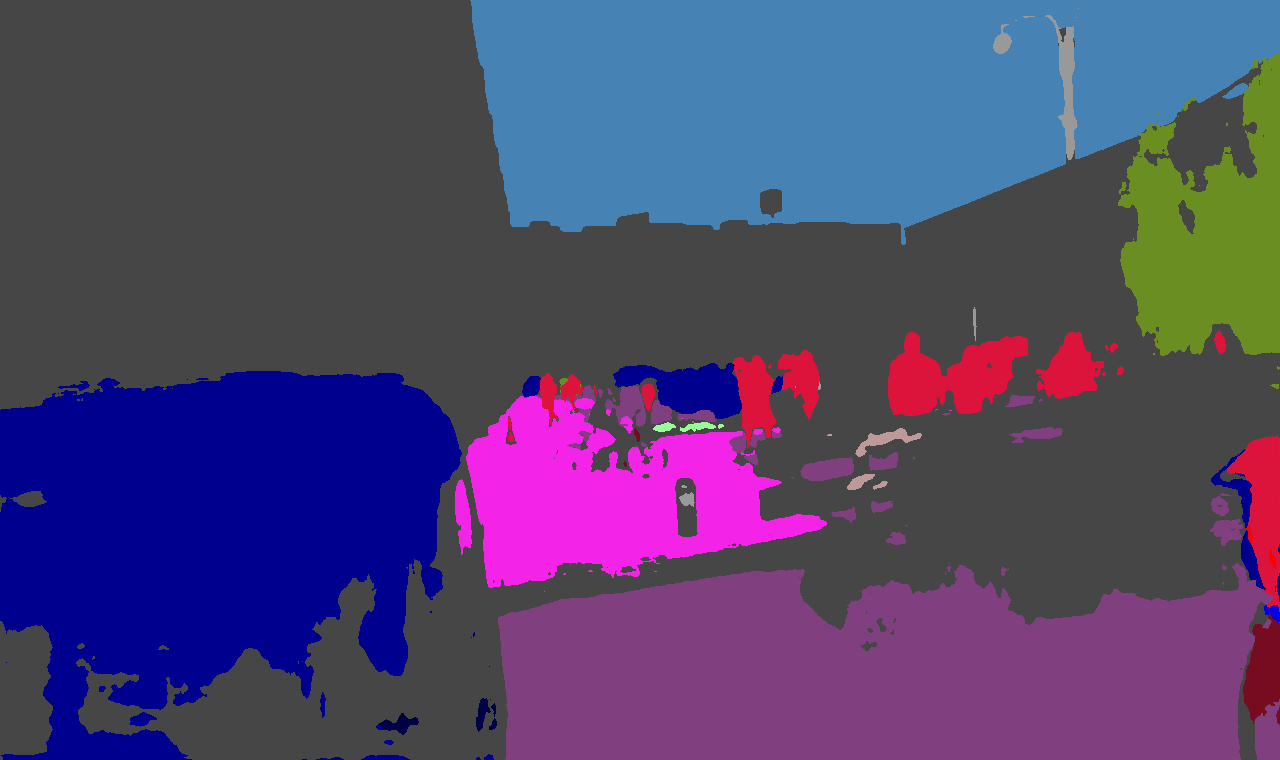}}
  \end{subfigure}
  \hfill
  \begin{subfigure}{0.19\textwidth}
    \raisebox{-\height}{\includegraphics[width=\textwidth, height=0.55\textwidth]{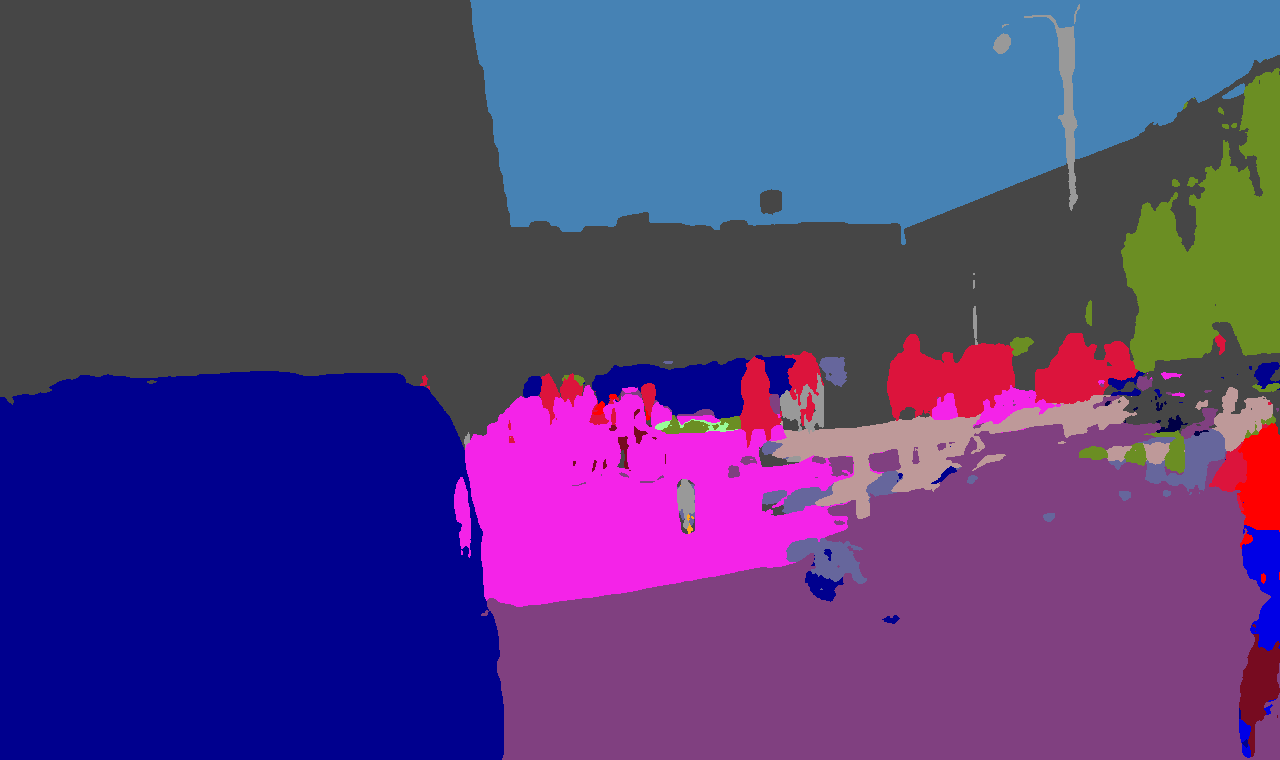}}
  \end{subfigure}
  \hfill
  \begin{subfigure}{0.19\textwidth}
    \raisebox{-\height}{\includegraphics[width=\textwidth, height=0.55\textwidth]{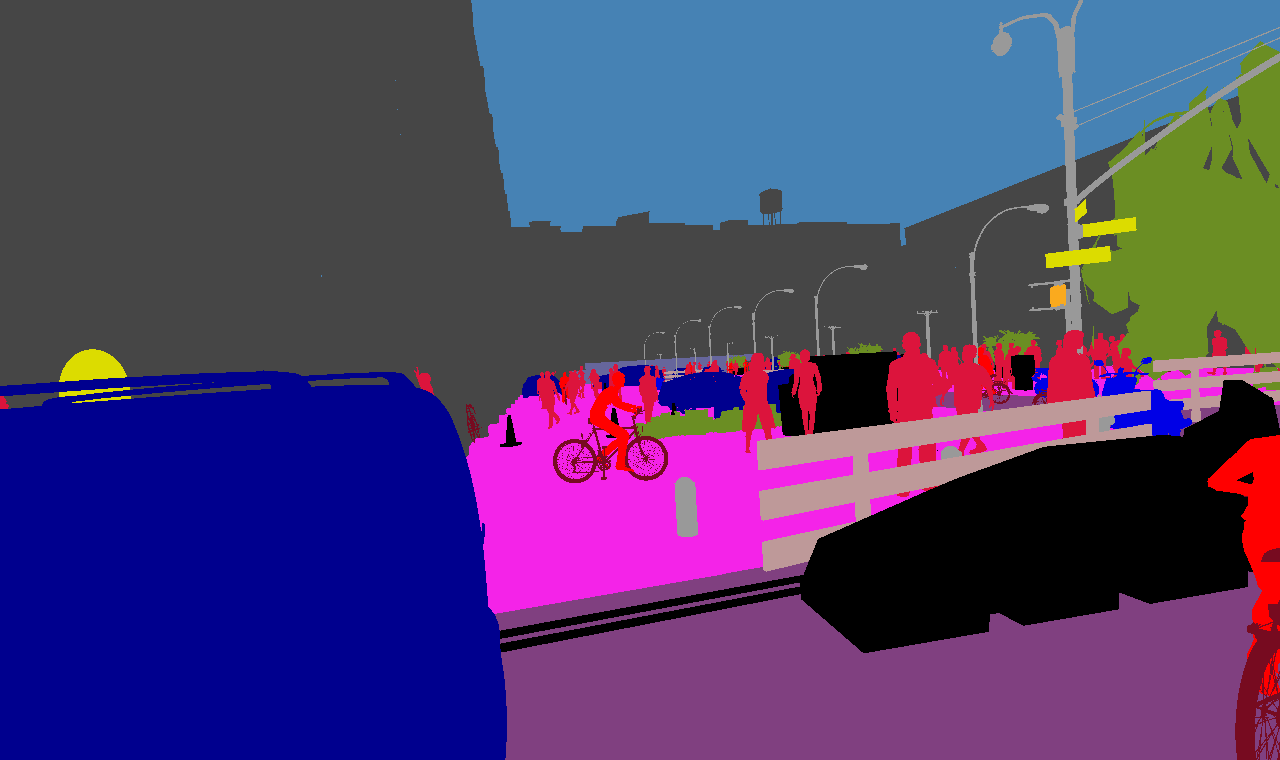}}
  \end{subfigure}
  \begin{subfigure}{0.19\textwidth}
    \raisebox{-\height}{\includegraphics[width=\textwidth, height=0.55\textwidth]{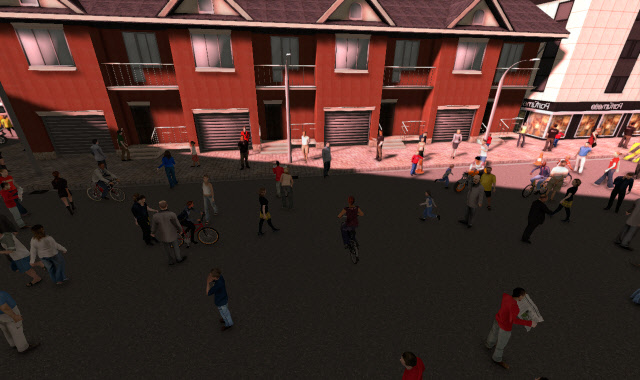}}
  \end{subfigure}
  \hfill
  \begin{subfigure}{0.19\textwidth}
    \raisebox{-\height}{\includegraphics[width=\textwidth, height=0.55\textwidth]{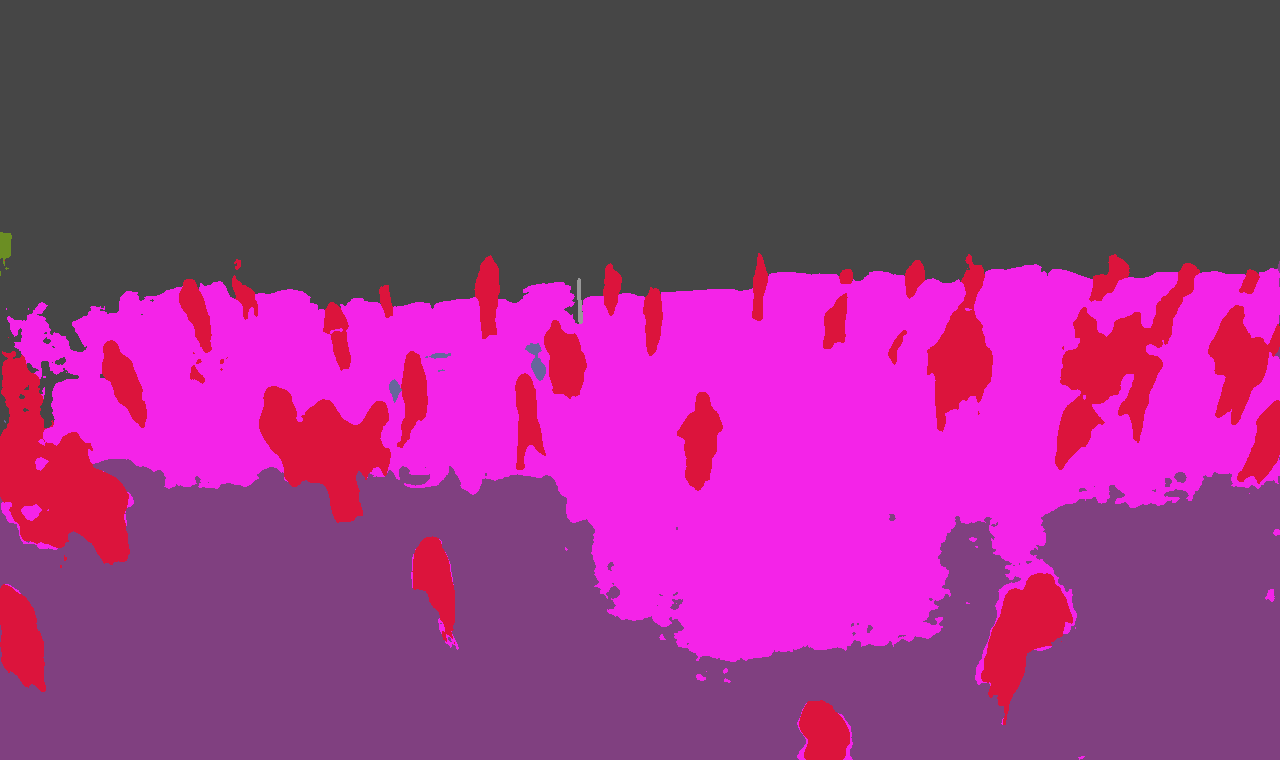}}
  \end{subfigure}
  \hfill
  \begin{subfigure}{0.19\textwidth}
    \raisebox{-\height}{\includegraphics[width=\textwidth, height=0.55\textwidth]{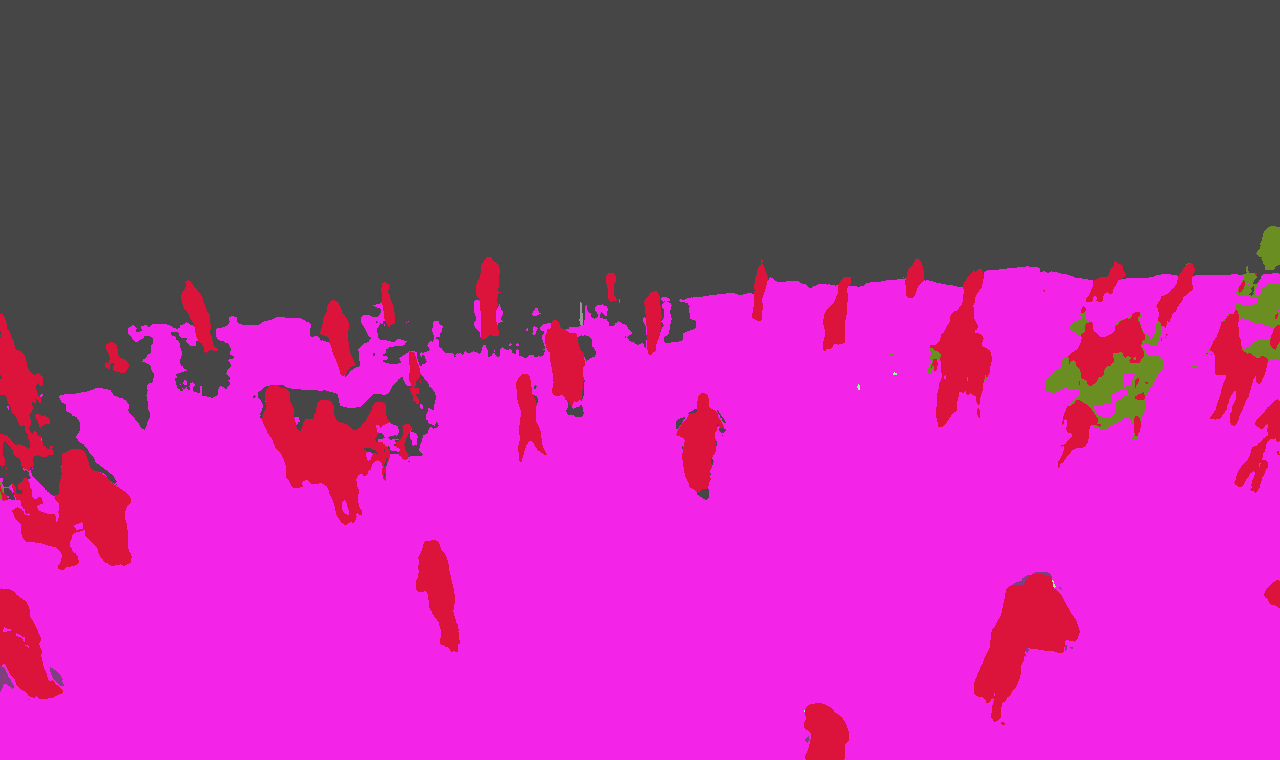}}
  \end{subfigure}
  \hfill
  \begin{subfigure}{0.19\textwidth}
    \raisebox{-\height}{\includegraphics[width=\textwidth, height=0.55\textwidth]{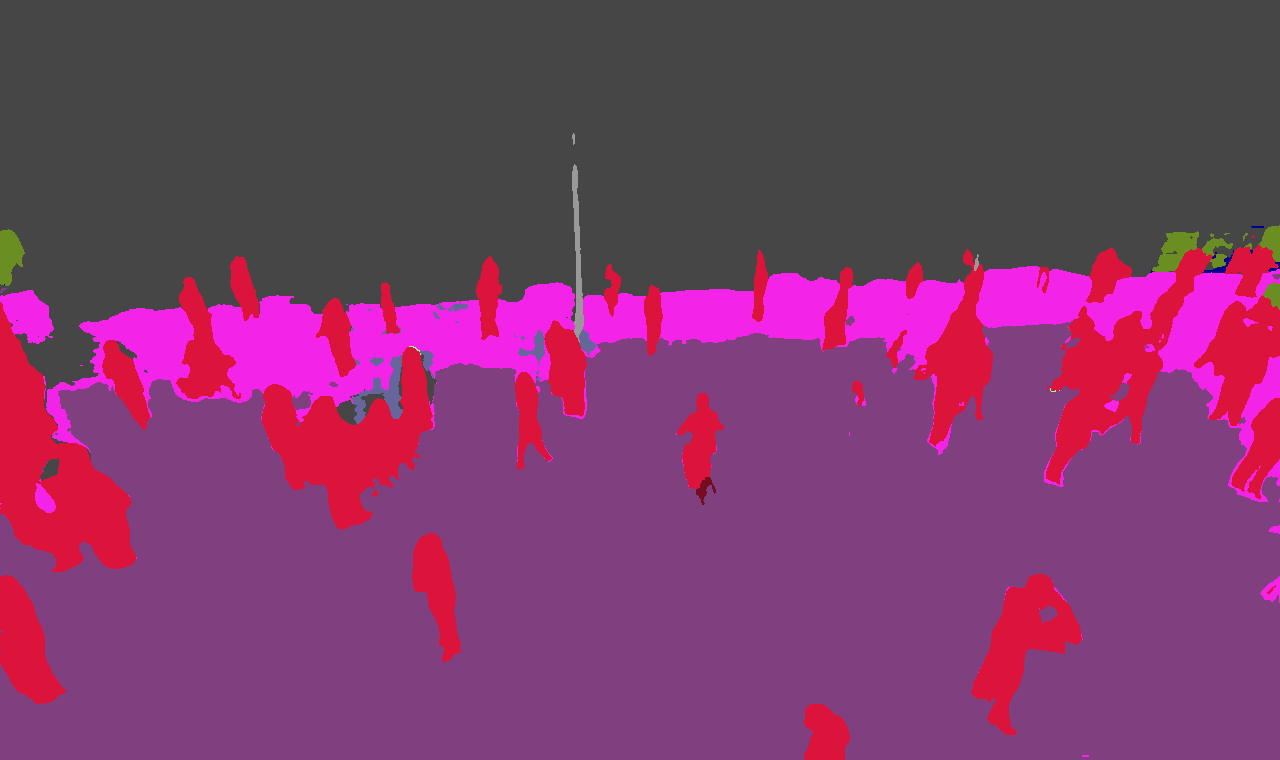}}
  \end{subfigure}
  \hfill
  \begin{subfigure}{0.19\textwidth}
    \raisebox{-\height}{\includegraphics[width=\textwidth, height=0.55\textwidth]{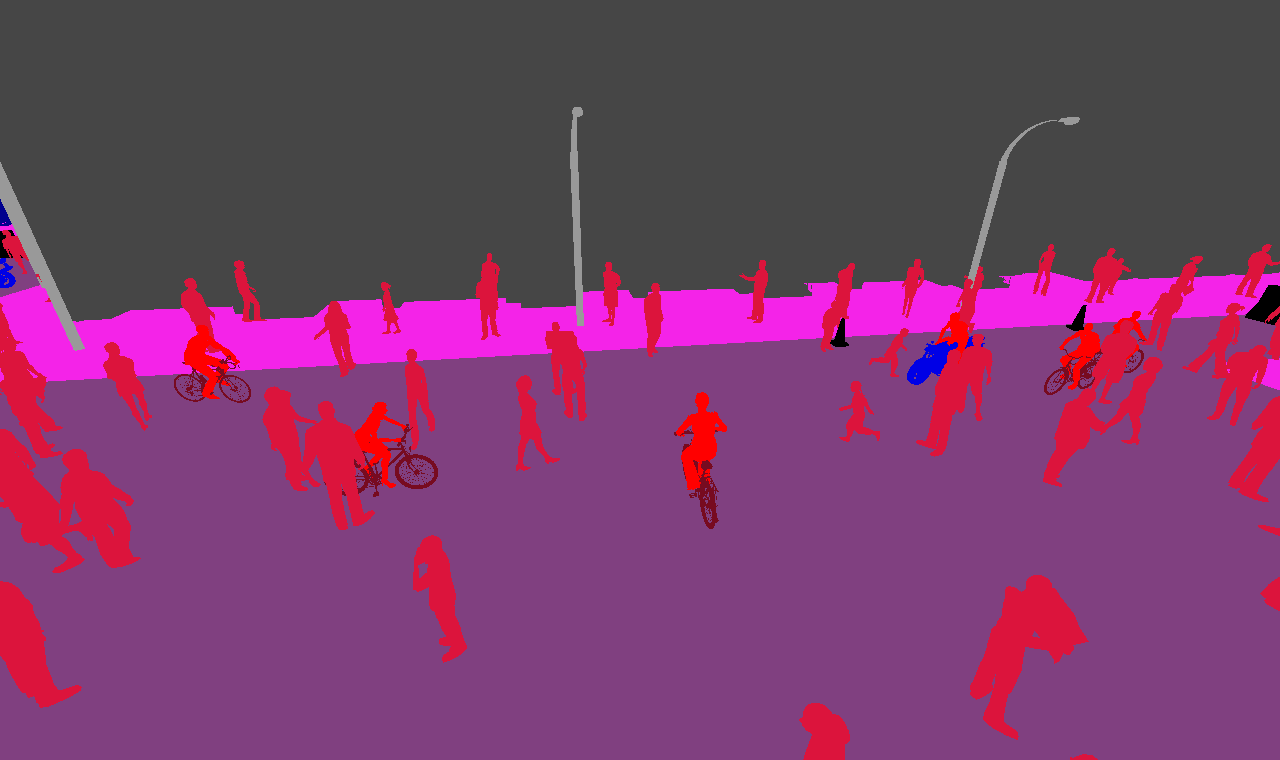}}
  \end{subfigure}
  \begin{subfigure}{0.19\textwidth}
    \raisebox{-\height}{\includegraphics[width=\textwidth, height=0.55\textwidth]{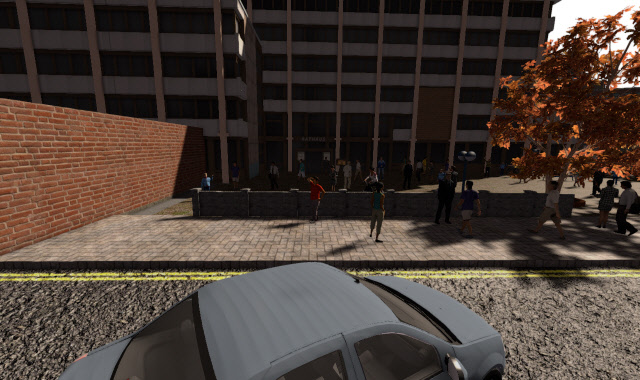}}
  \end{subfigure}
  \hfill
  \begin{subfigure}{0.19\textwidth}
    \raisebox{-\height}{\includegraphics[width=\textwidth, height=0.55\textwidth]{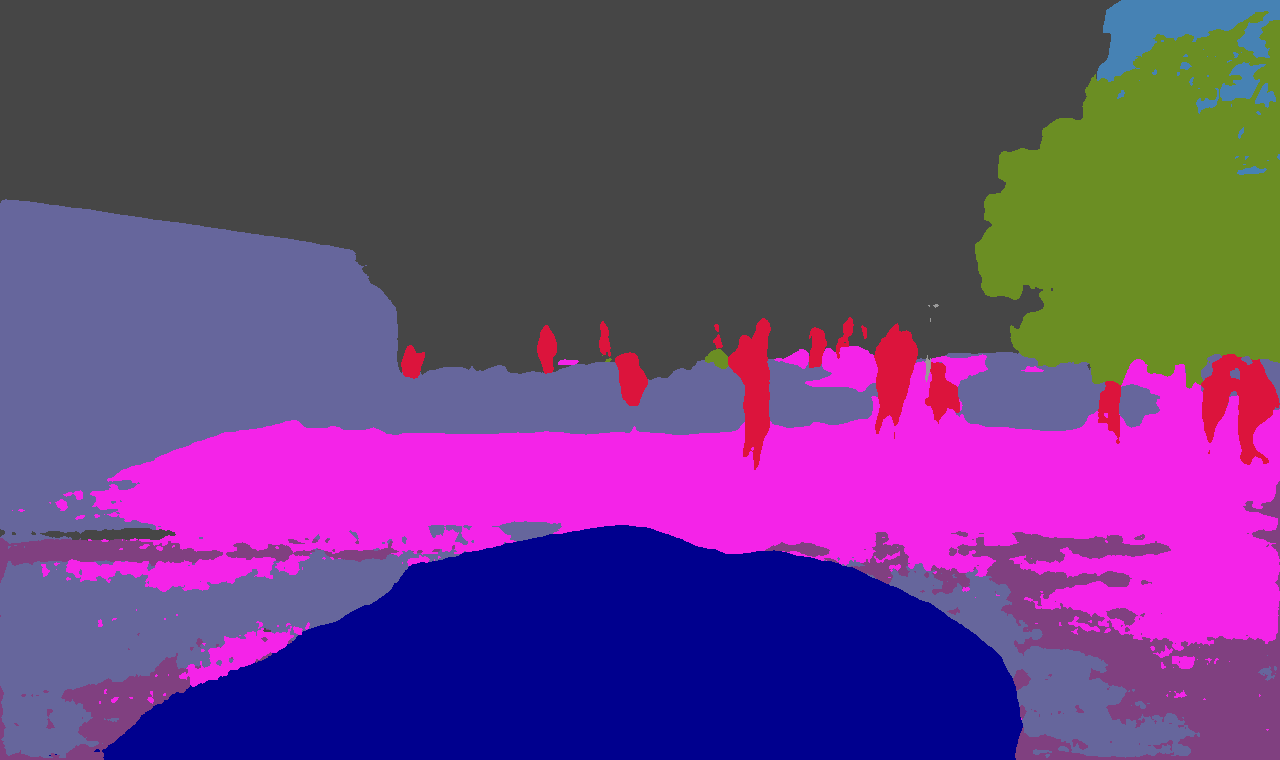}}
  \end{subfigure}
  \hfill
  \begin{subfigure}{0.19\textwidth}
    \raisebox{-\height}{\includegraphics[width=\textwidth, height=0.55\textwidth]{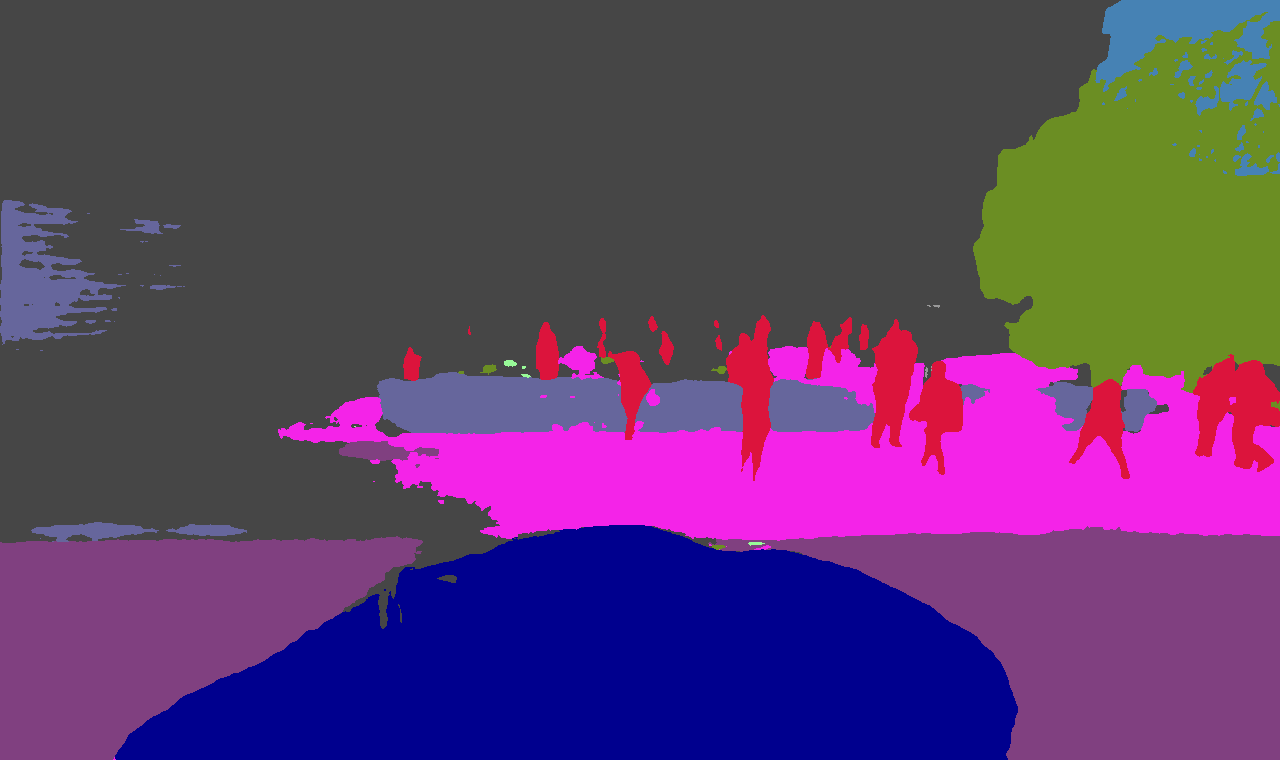}}
  \end{subfigure}
  \hfill
  \begin{subfigure}{0.19\textwidth}
    \raisebox{-\height}{\includegraphics[width=\textwidth, height=0.55\textwidth]{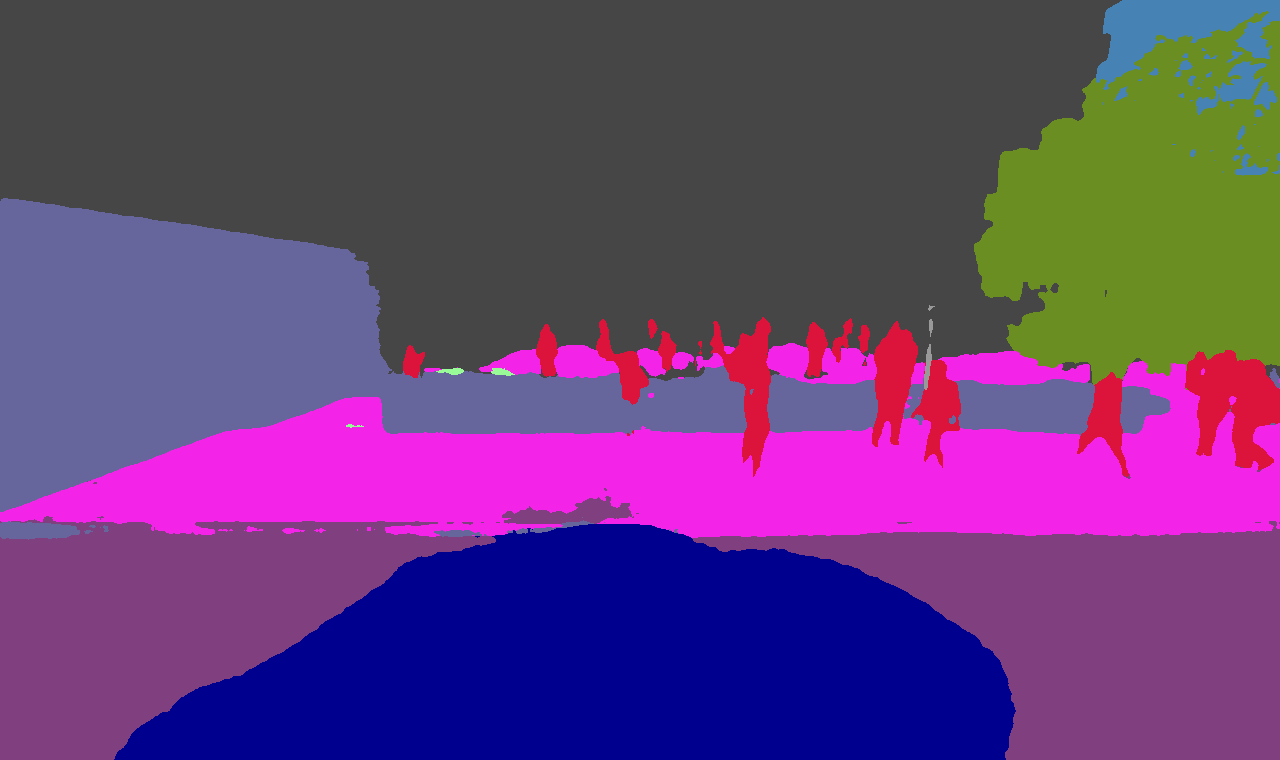}}
  \end{subfigure}
  \hfill
  \begin{subfigure}{0.19\textwidth}
    \raisebox{-\height}{\includegraphics[width=\textwidth, height=0.55\textwidth]{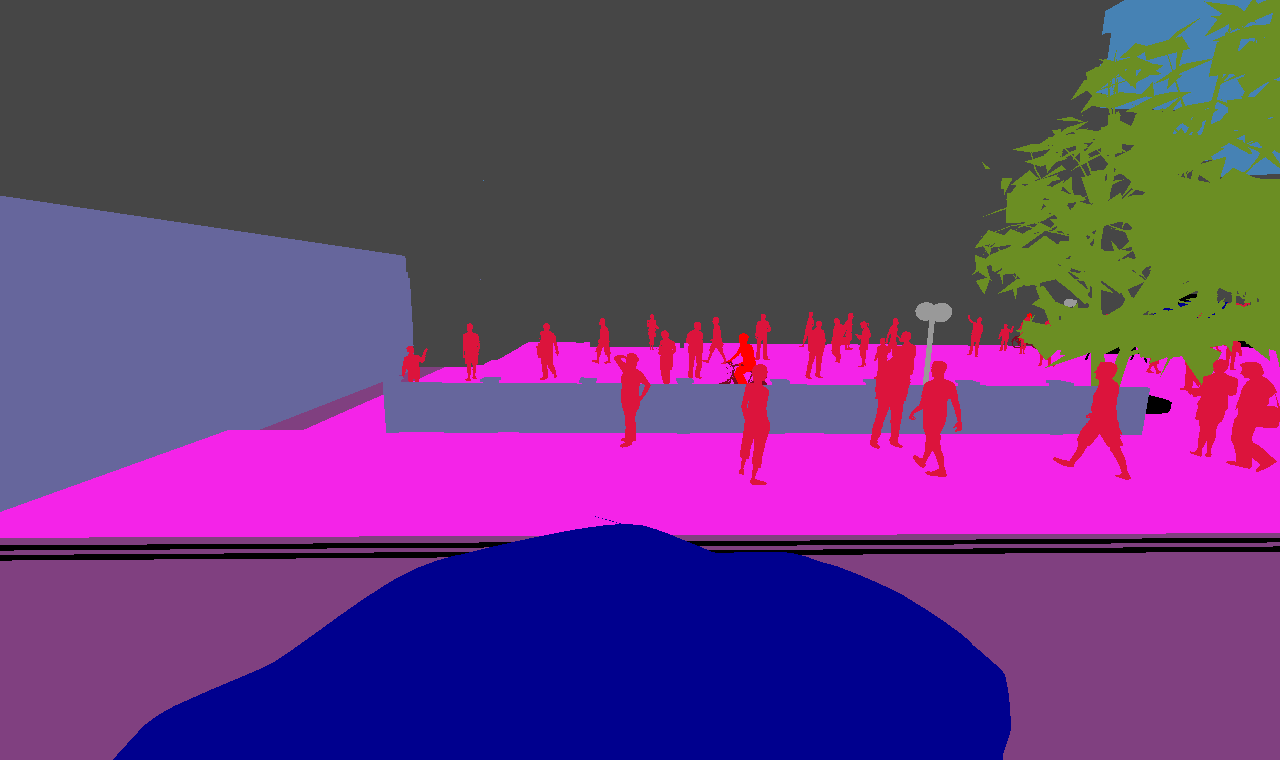}}
  \end{subfigure}
  \begin{subfigure}{0.19\textwidth}
    \raisebox{-\height}{\includegraphics[width=\textwidth, height=0.55\textwidth]{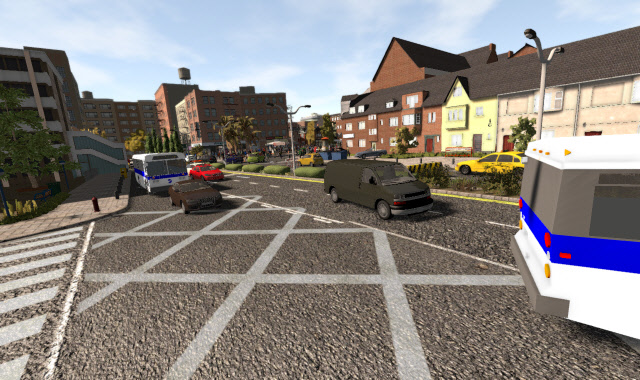}}
  \end{subfigure}
  \hfill
  \begin{subfigure}{0.19\textwidth}
    \raisebox{-\height}{\includegraphics[width=\textwidth, height=0.55\textwidth]{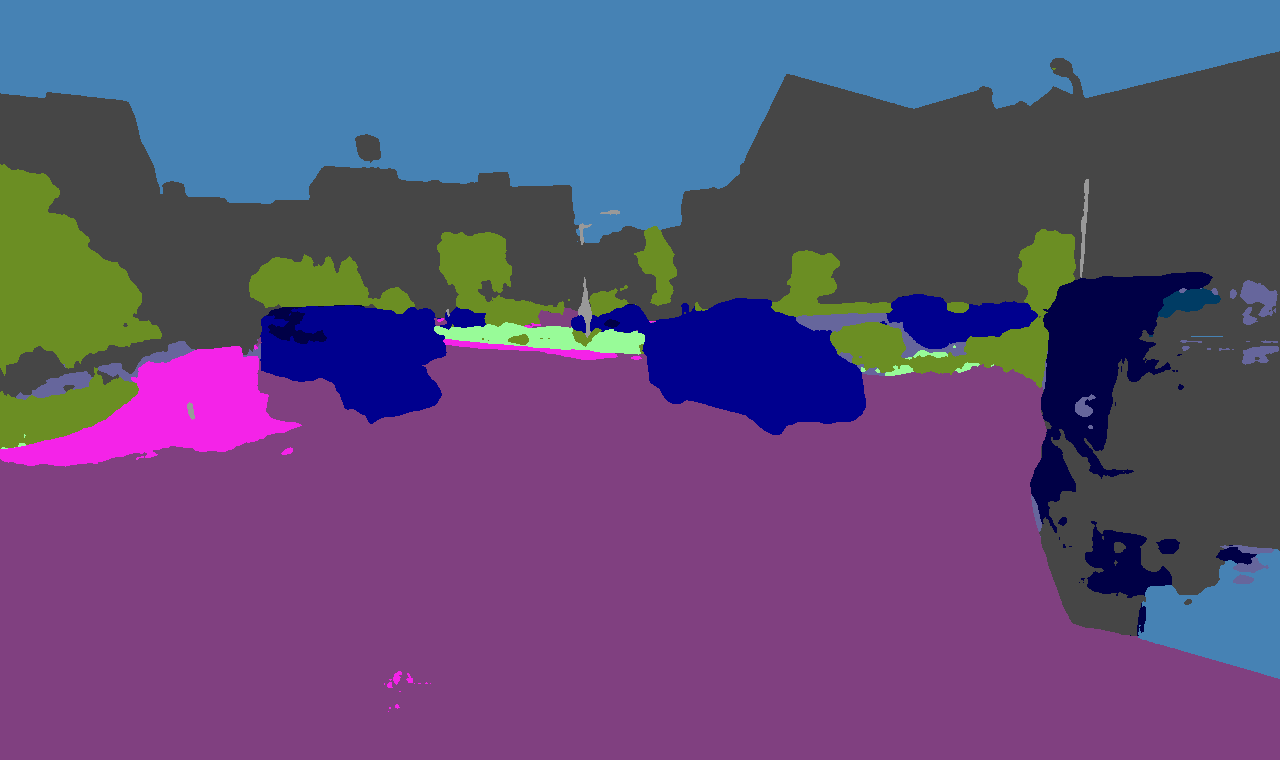}}
  \end{subfigure}
  \hfill
  \begin{subfigure}{0.19\textwidth}
    \raisebox{-\height}{\includegraphics[width=\textwidth, height=0.55\textwidth]{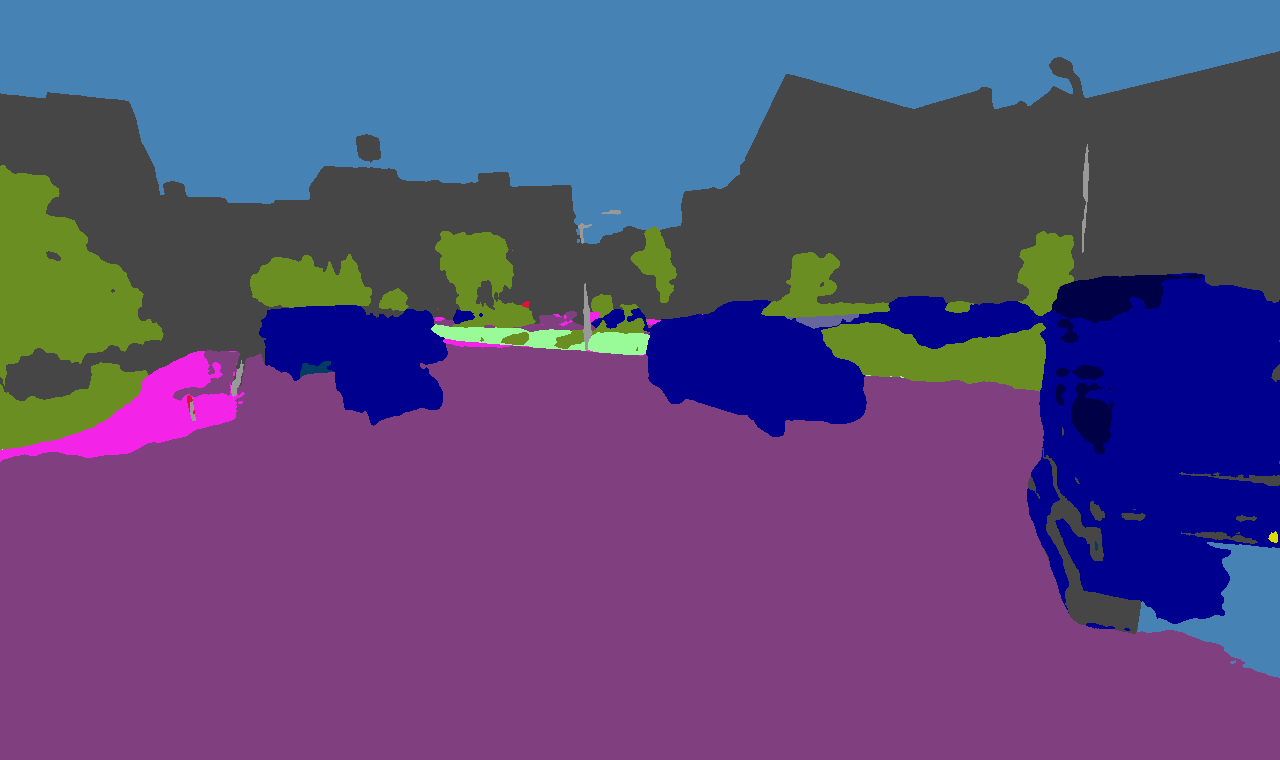}}
  \end{subfigure}
  \hfill
  \begin{subfigure}{0.19\textwidth}
    \raisebox{-\height}{\includegraphics[width=\textwidth, height=0.55\textwidth]{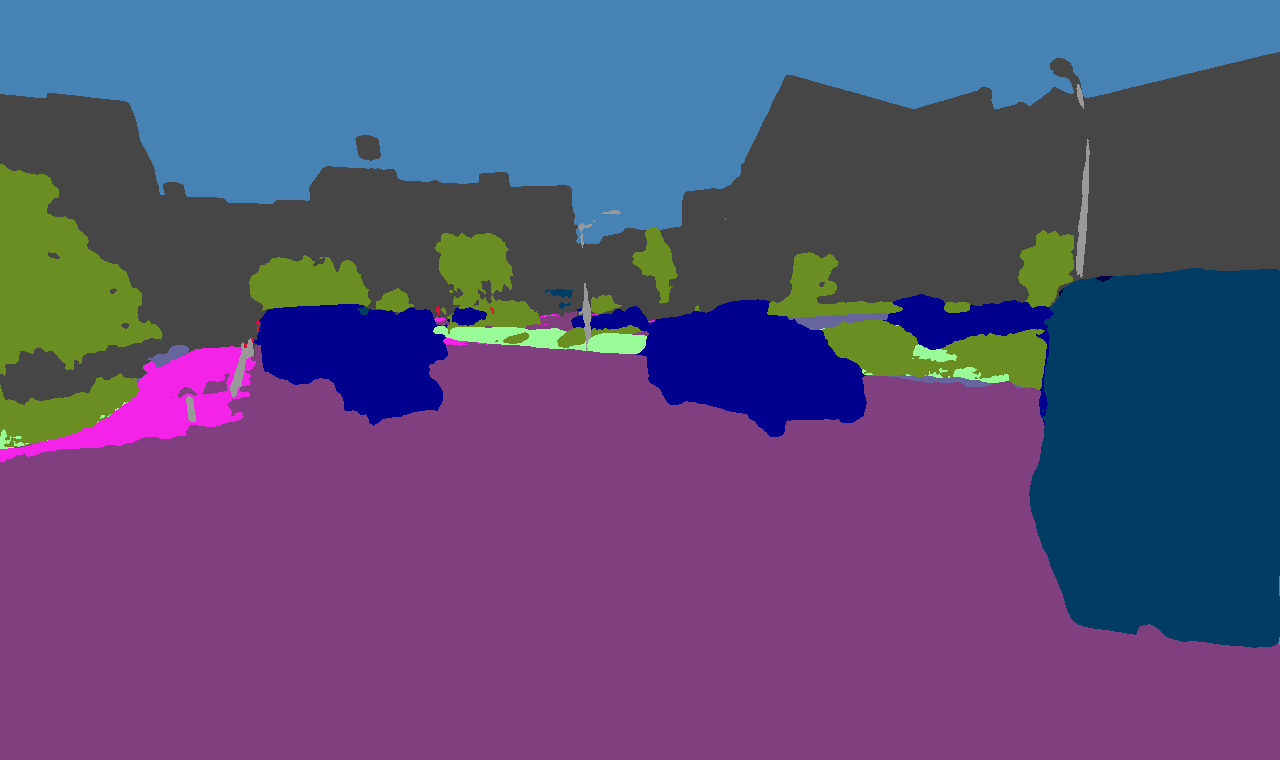}}
  \end{subfigure}
  \hfill
  \begin{subfigure}{0.19\textwidth}
    \raisebox{-\height}{\includegraphics[width=\textwidth, height=0.55\textwidth]{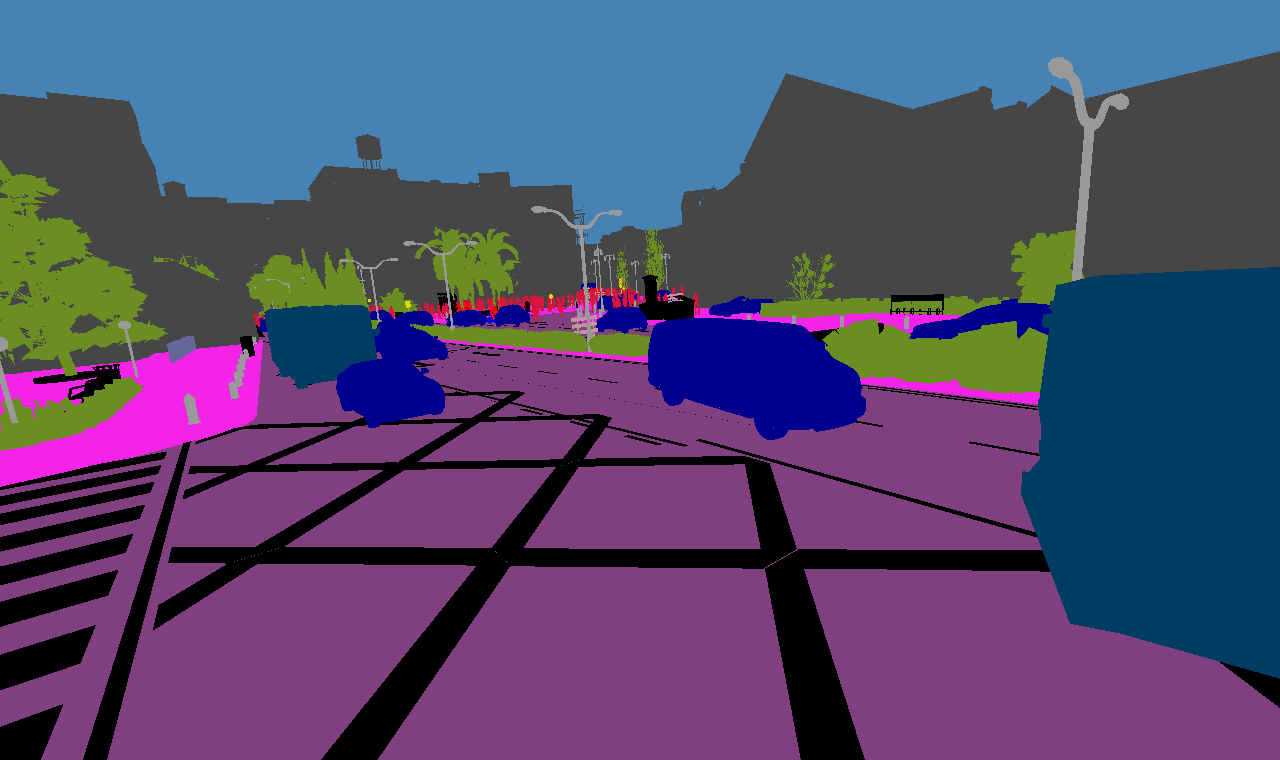}}
  \end{subfigure}
  \begin{subfigure}{0.19\textwidth}
    \raisebox{-\height}{\includegraphics[width=\textwidth, height=0.55\textwidth]{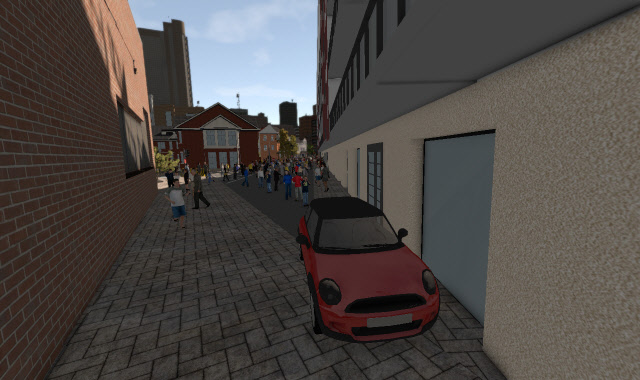}}
    \caption*{Unseen domain image}
    \label{fig:compare_r50_syn_img}
  \end{subfigure}
  \hfill
  \begin{subfigure}{0.19\textwidth}
    \raisebox{-\height}{\includegraphics[width=\textwidth, height=0.55\textwidth]{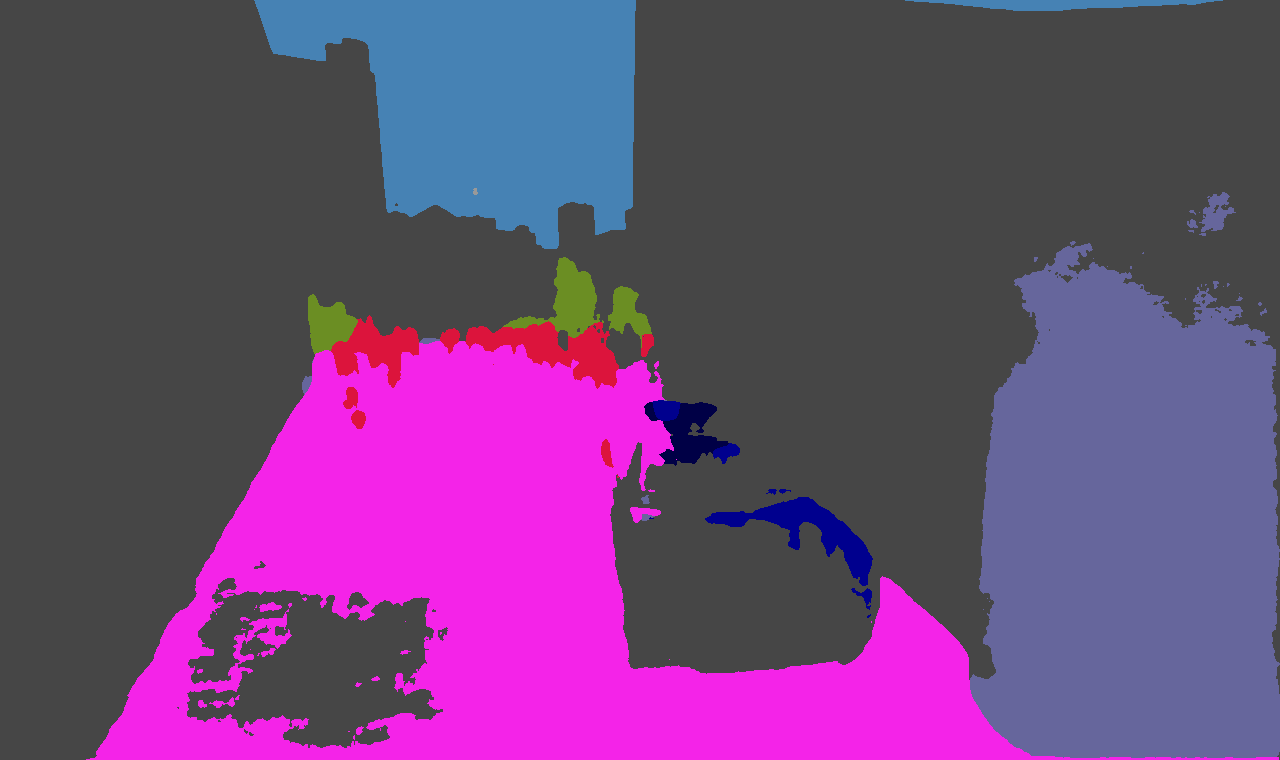}}
    \caption*{Baseline}
    \label{fig:compare_r50_syn_base}
  \end{subfigure}
  \hfill
  \begin{subfigure}{0.19\textwidth}
    \raisebox{-\height}{\includegraphics[width=\textwidth, height=0.55\textwidth]{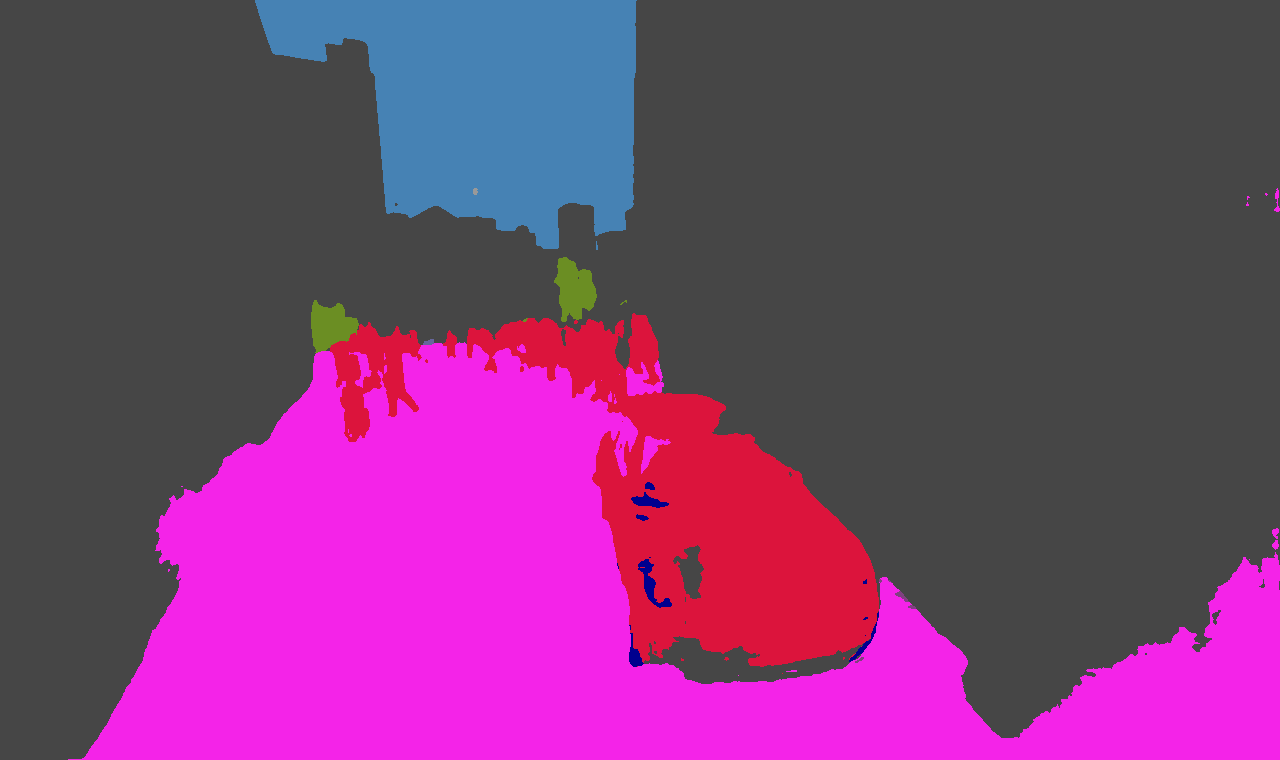}}
    \caption*{RobustNet}
    \label{fig:compare_r50_syn_isw}
  \end{subfigure}
  \hfill
  \begin{subfigure}{0.19\textwidth}
    \raisebox{-\height}{\includegraphics[width=\textwidth, height=0.55\textwidth]{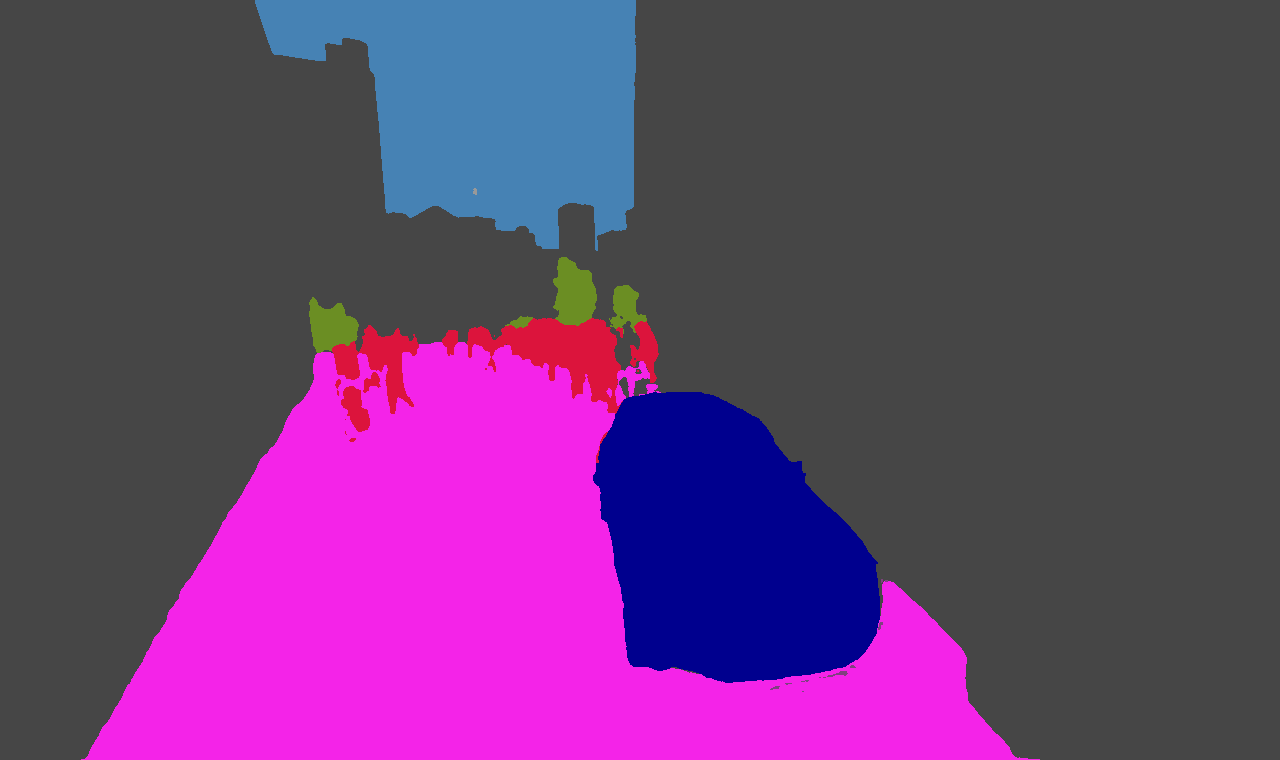}}
    \caption*{\textbf{Ours (WildNet)}}
    \label{fig:compare_r50_syn_ours}
  \end{subfigure}
  \hfill
  \begin{subfigure}{0.19\textwidth}
    \raisebox{-\height}{\includegraphics[width=\textwidth, height=0.55\textwidth]{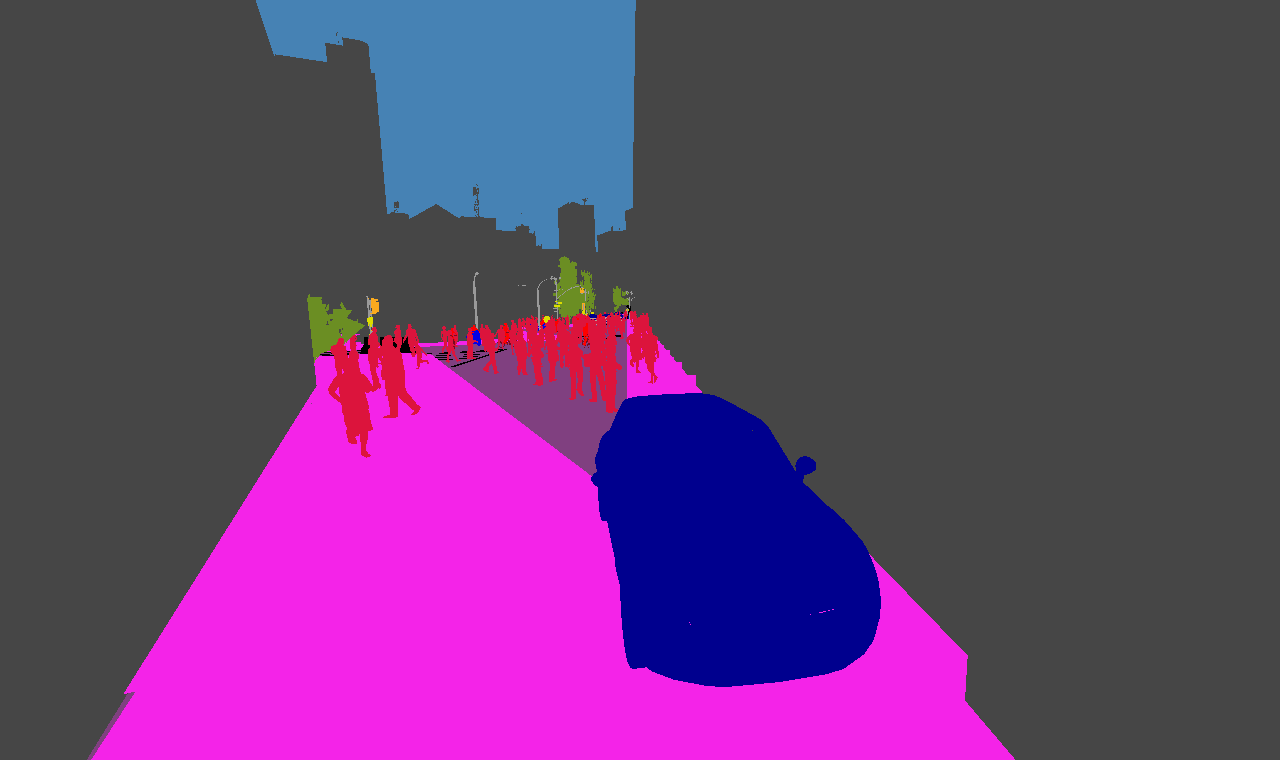}}
    \caption*{Ground truth}
    \label{fig:compare_r50_syn_gt}
  \end{subfigure}
  \caption{Semantic segmentation results on unseen domain images in SYNTHIA with the models trained on GTAV.
  }
  \label{fig:compare_r50_syn_img_base_isw_ours}
\end{figure*}

\begin{figure*}
  \centering
  \begin{subfigure}{0.19\textwidth}
    \raisebox{-\height}{\includegraphics[width=\textwidth, height=0.55\textwidth]{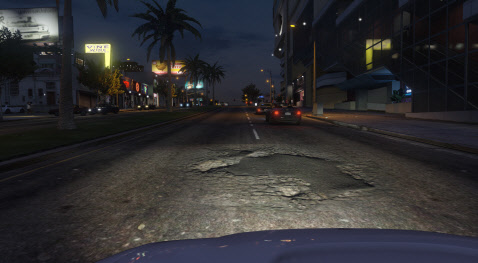}}
  \end{subfigure}
  \hfill
  \begin{subfigure}{0.19\textwidth}
    \raisebox{-\height}{\includegraphics[width=\textwidth, height=0.55\textwidth]{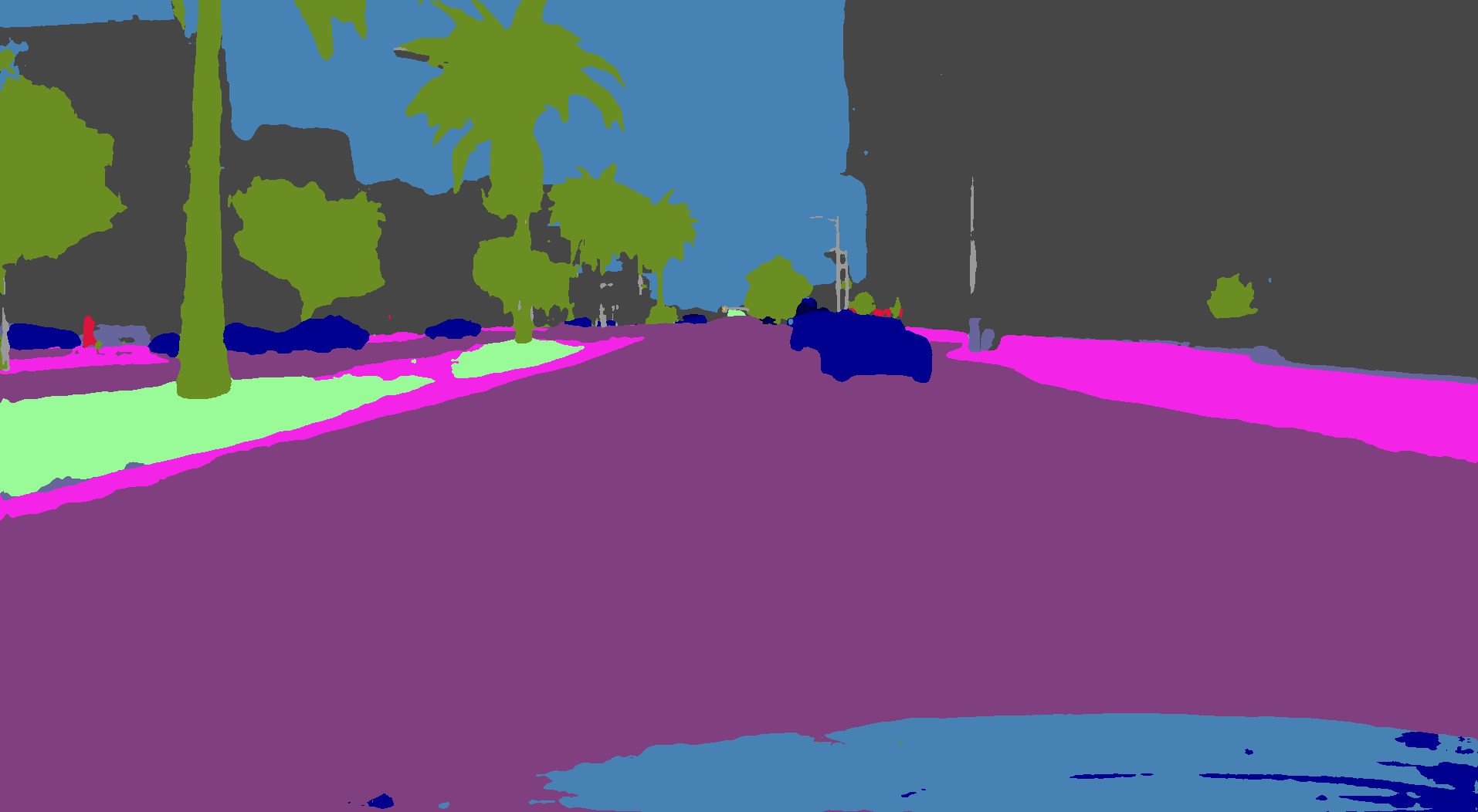}}
  \end{subfigure}
  \hfill
  \begin{subfigure}{0.19\textwidth}
    \raisebox{-\height}{\includegraphics[width=\textwidth, height=0.55\textwidth]{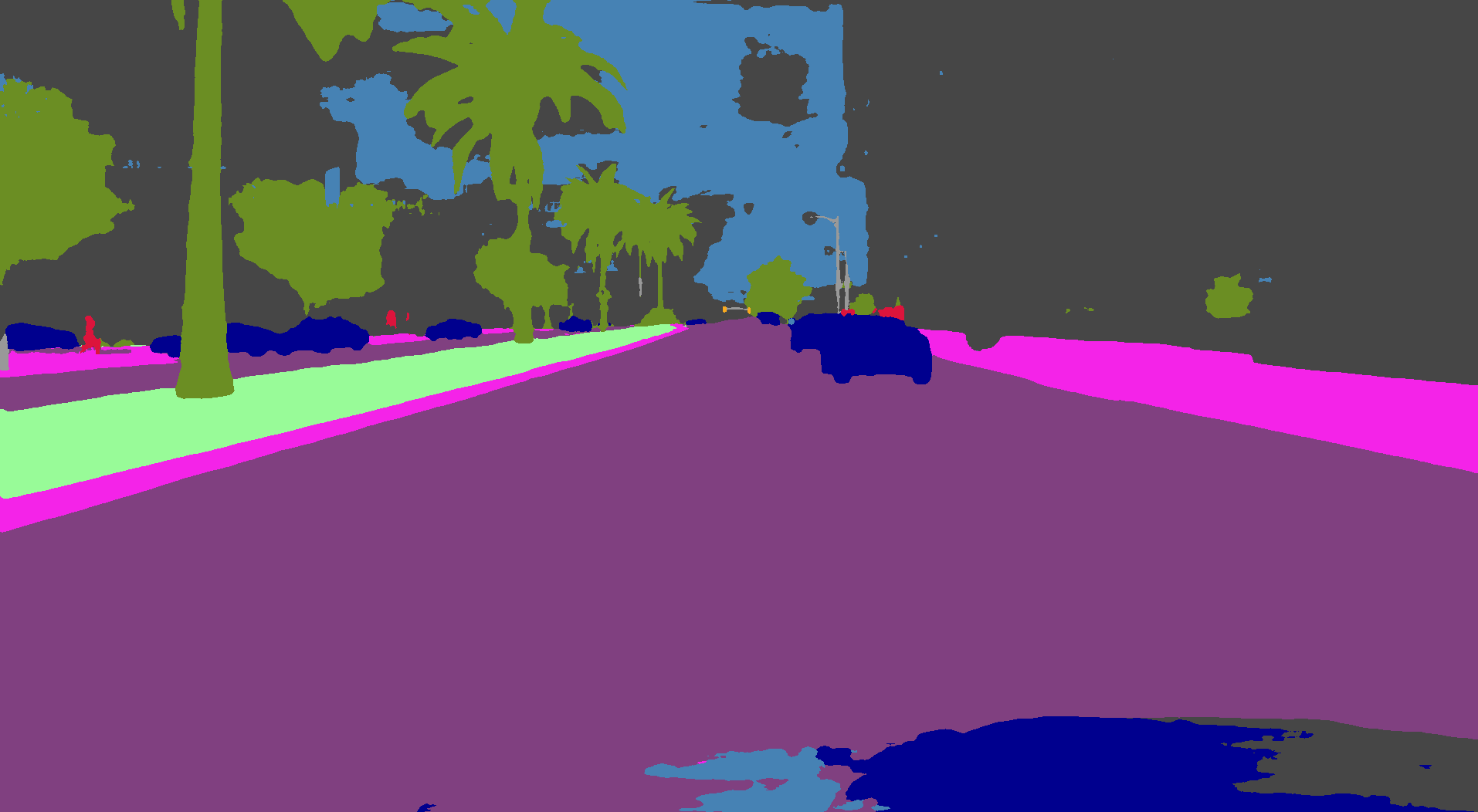}}
  \end{subfigure}
  \hfill
  \begin{subfigure}{0.19\textwidth}
    \raisebox{-\height}{\includegraphics[width=\textwidth, height=0.55\textwidth]{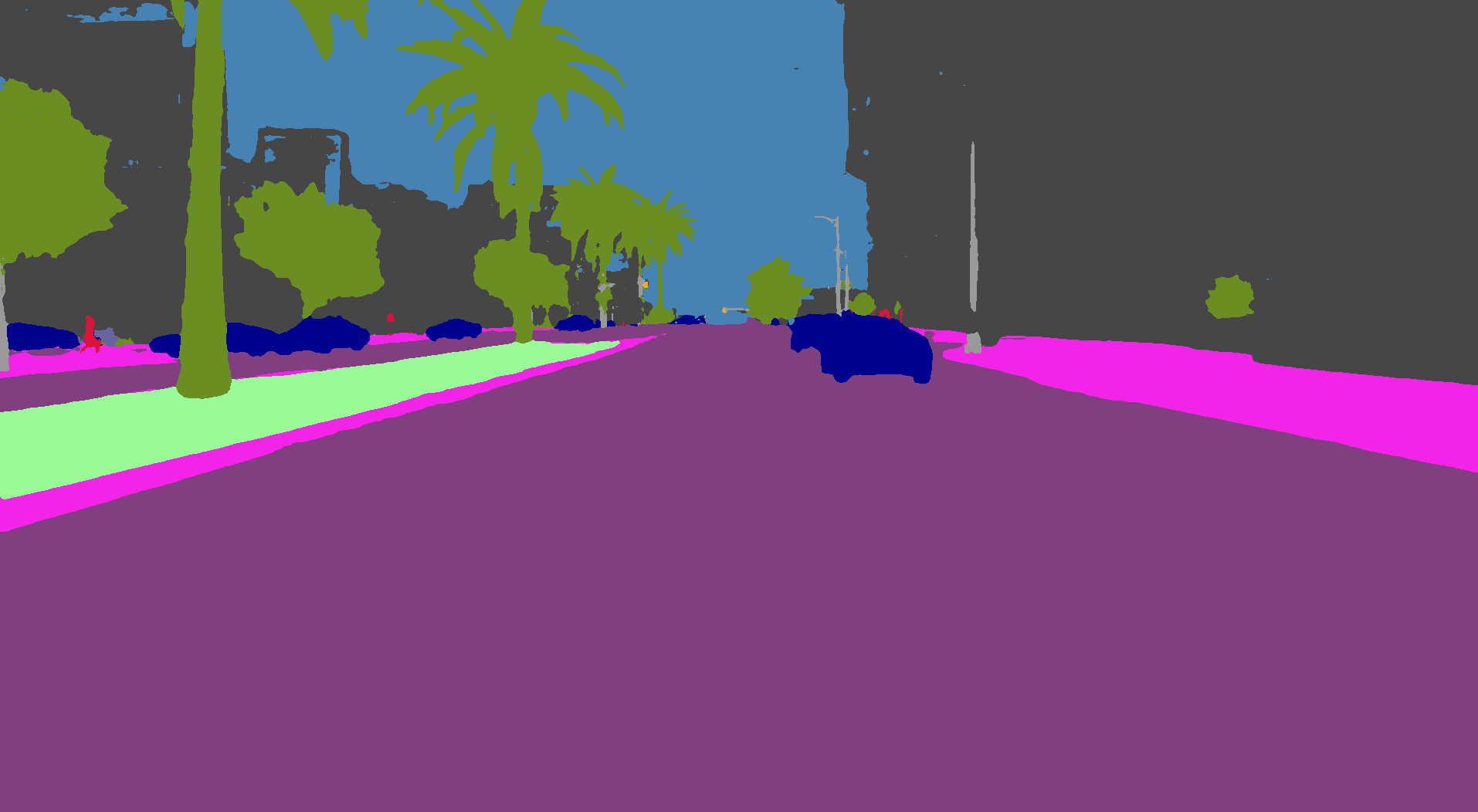}}
  \end{subfigure}
  \hfill
  \begin{subfigure}{0.19\textwidth}
    \raisebox{-\height}{\includegraphics[width=\textwidth, height=0.55\textwidth]{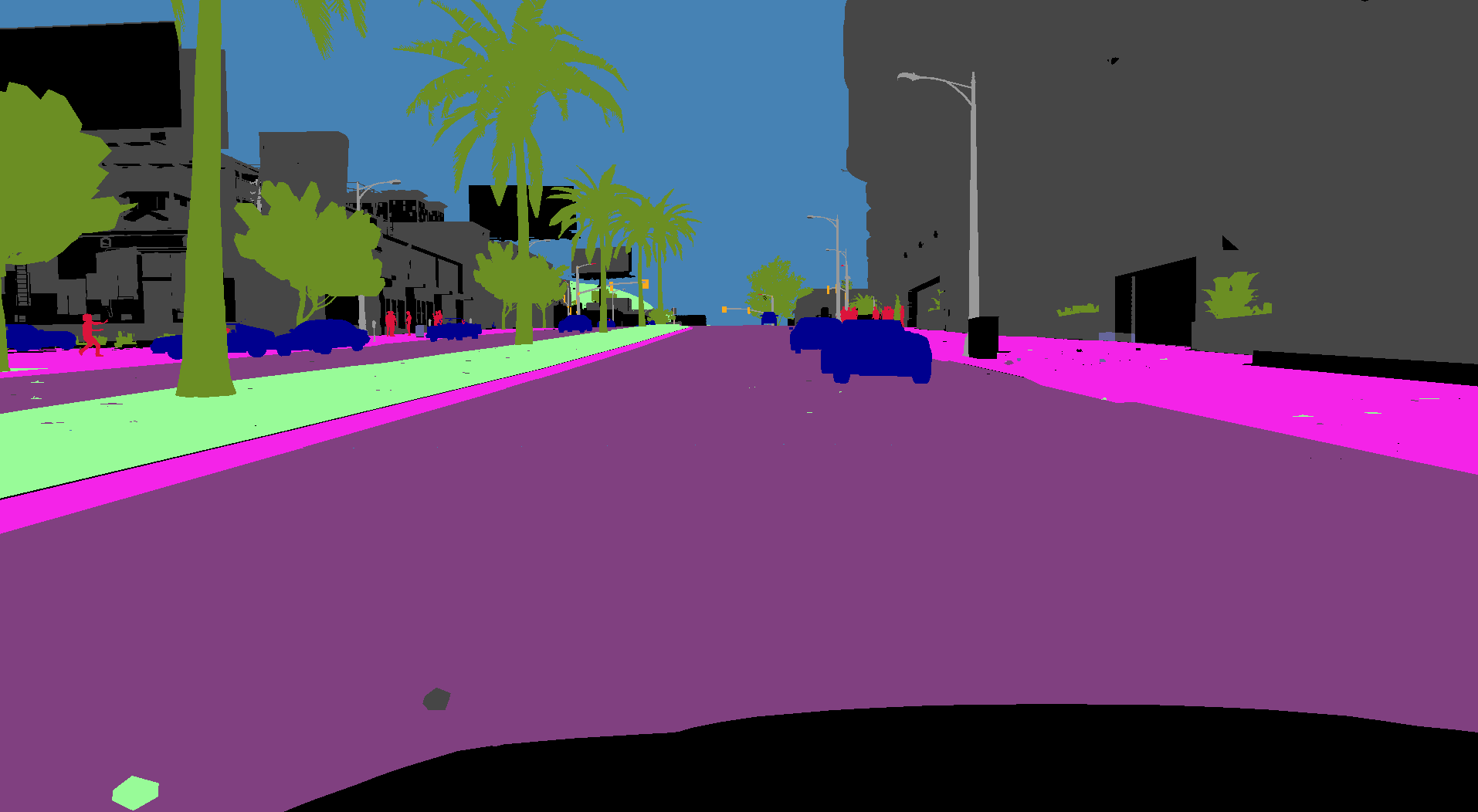}}
  \end{subfigure}
  \begin{subfigure}{0.19\textwidth}
    \raisebox{-\height}{\includegraphics[width=\textwidth, height=0.55\textwidth]{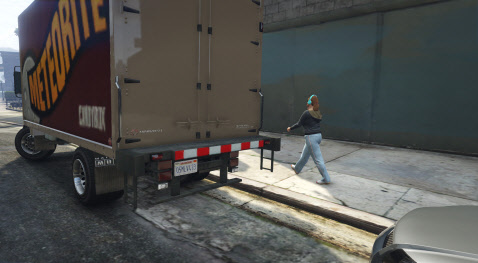}}
  \end{subfigure}
  \hfill
  \begin{subfigure}{0.19\textwidth}
    \raisebox{-\height}{\includegraphics[width=\textwidth, height=0.55\textwidth]{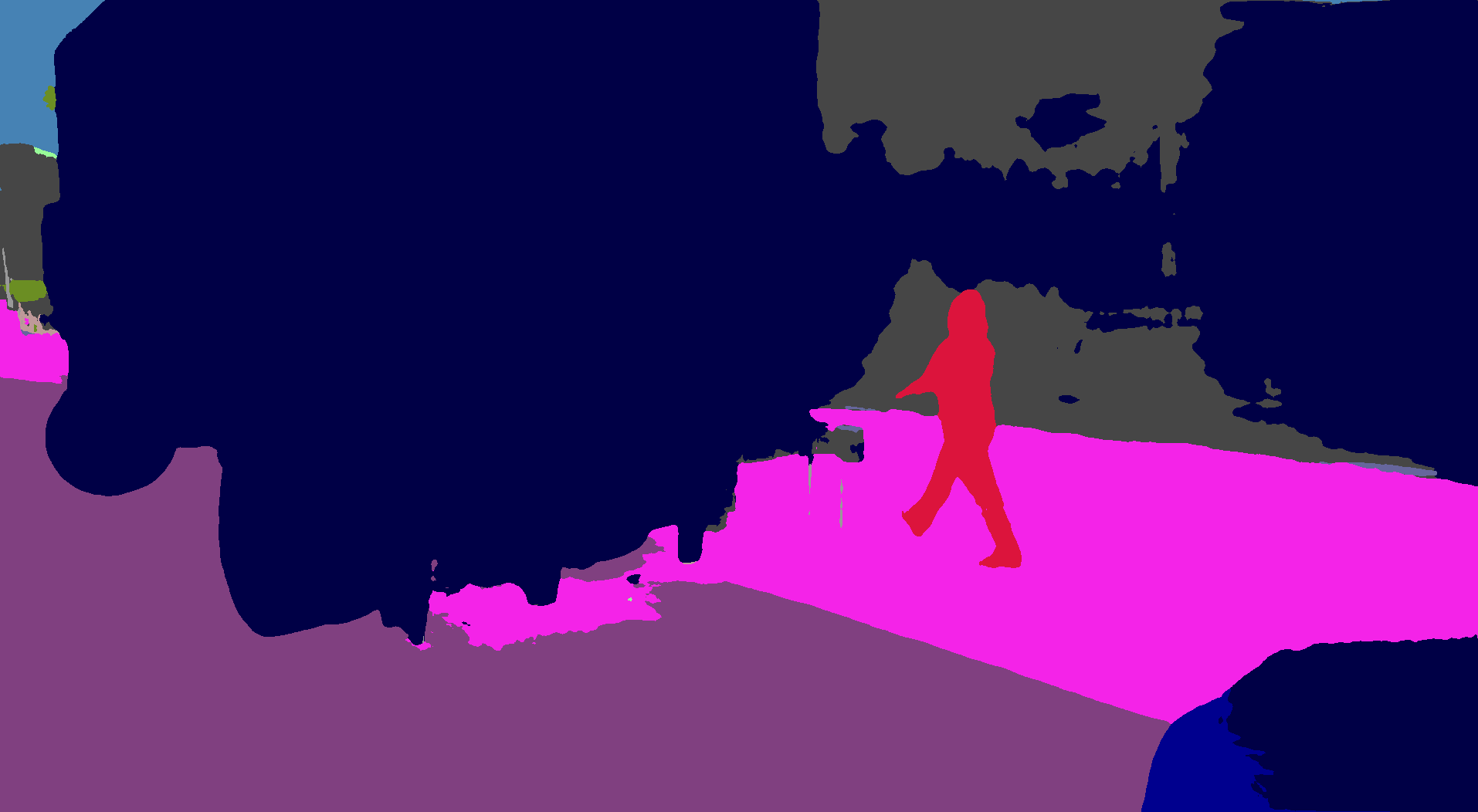}}
  \end{subfigure}
  \hfill
  \begin{subfigure}{0.19\textwidth}
    \raisebox{-\height}{\includegraphics[width=\textwidth, height=0.55\textwidth]{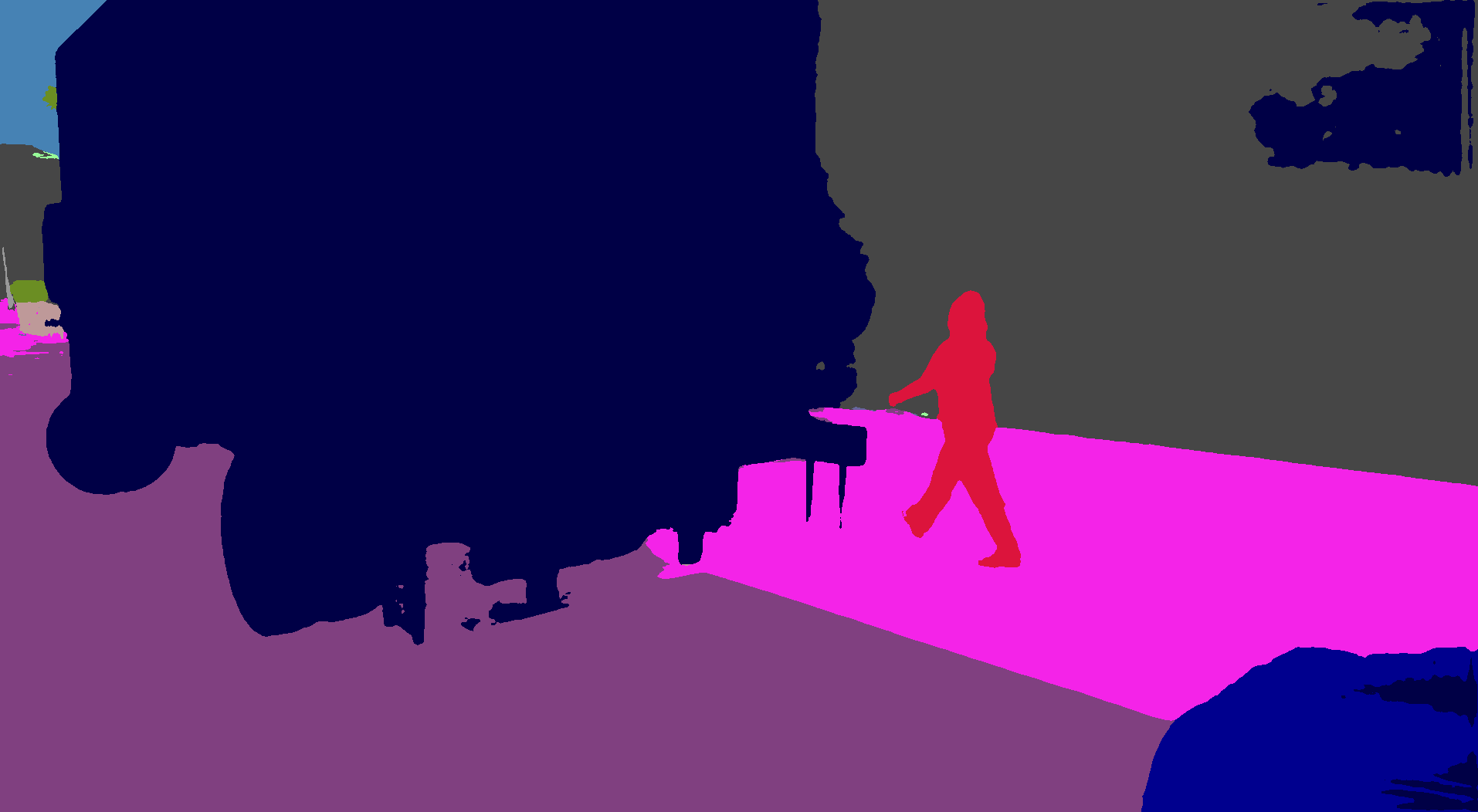}}
  \end{subfigure}
  \hfill
  \begin{subfigure}{0.19\textwidth}
    \raisebox{-\height}{\includegraphics[width=\textwidth, height=0.55\textwidth]{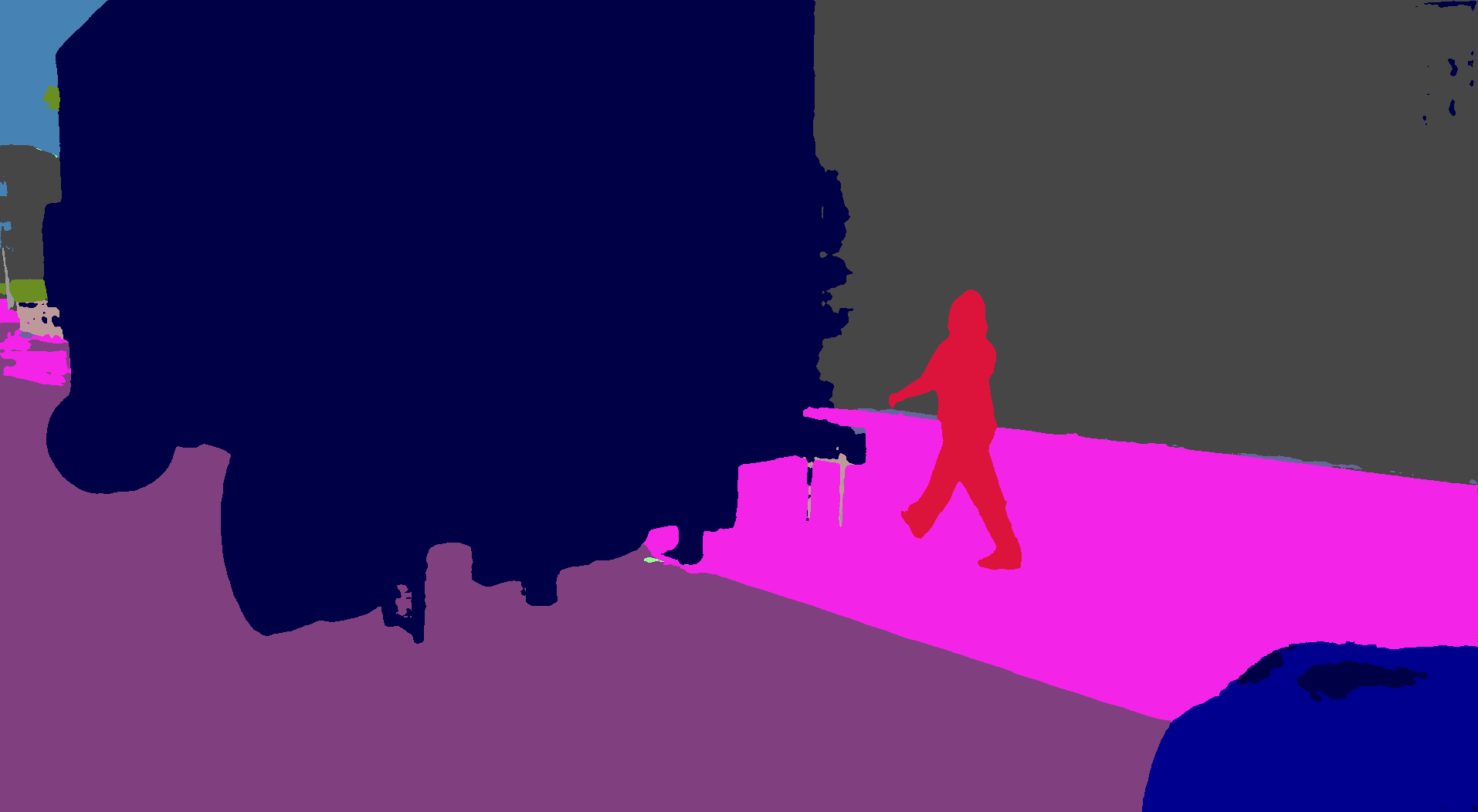}}
  \end{subfigure}
  \hfill
  \begin{subfigure}{0.19\textwidth}
    \raisebox{-\height}{\includegraphics[width=\textwidth, height=0.55\textwidth]{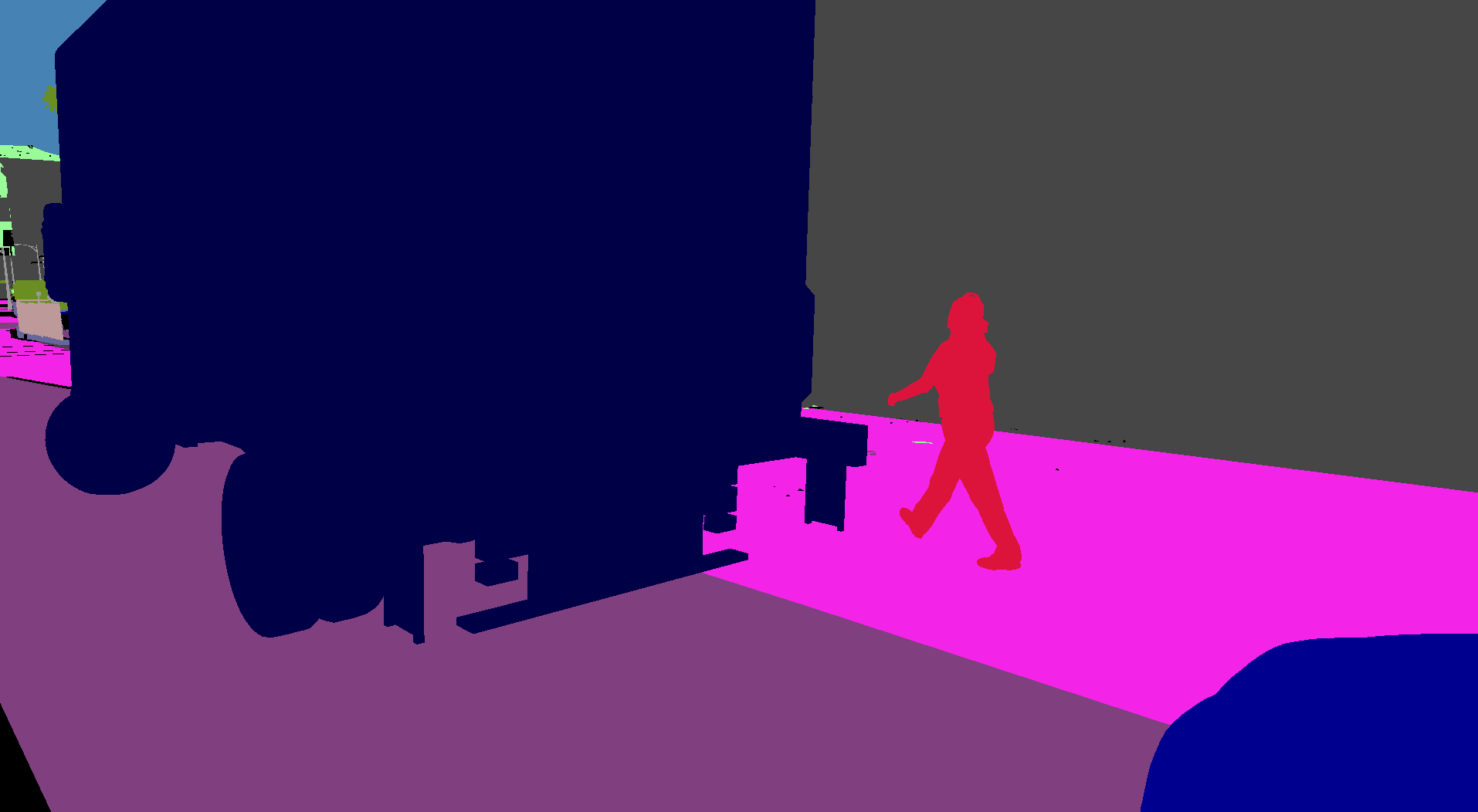}}
  \end{subfigure}
  \begin{subfigure}{0.19\textwidth}
    \raisebox{-\height}{\includegraphics[width=\textwidth, height=0.55\textwidth]{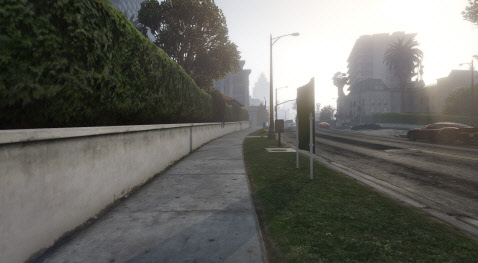}}
  \end{subfigure}
  \hfill
  \begin{subfigure}{0.19\textwidth}
    \raisebox{-\height}{\includegraphics[width=\textwidth, height=0.55\textwidth]{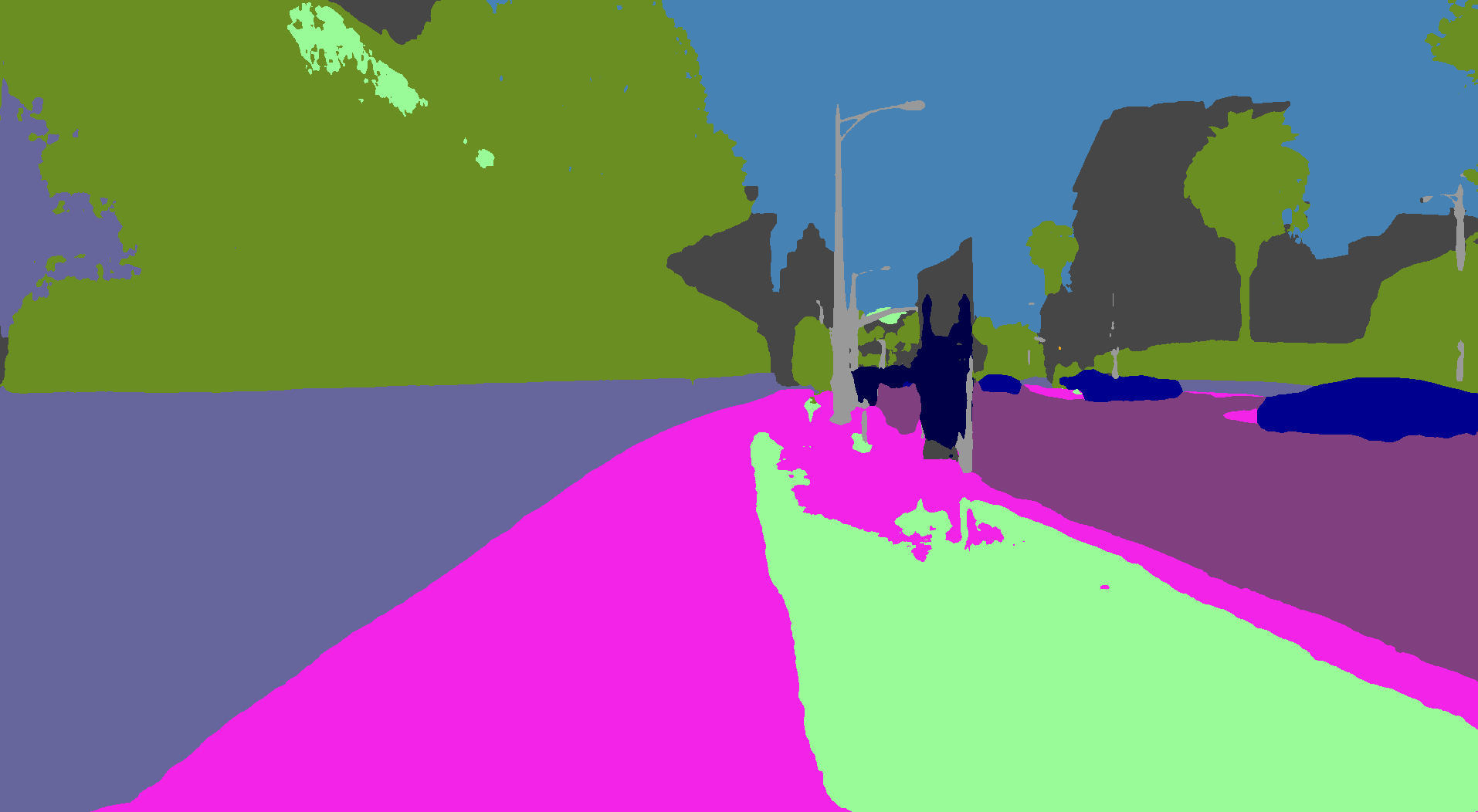}}
  \end{subfigure}
  \hfill
  \begin{subfigure}{0.19\textwidth}
    \raisebox{-\height}{\includegraphics[width=\textwidth, height=0.55\textwidth]{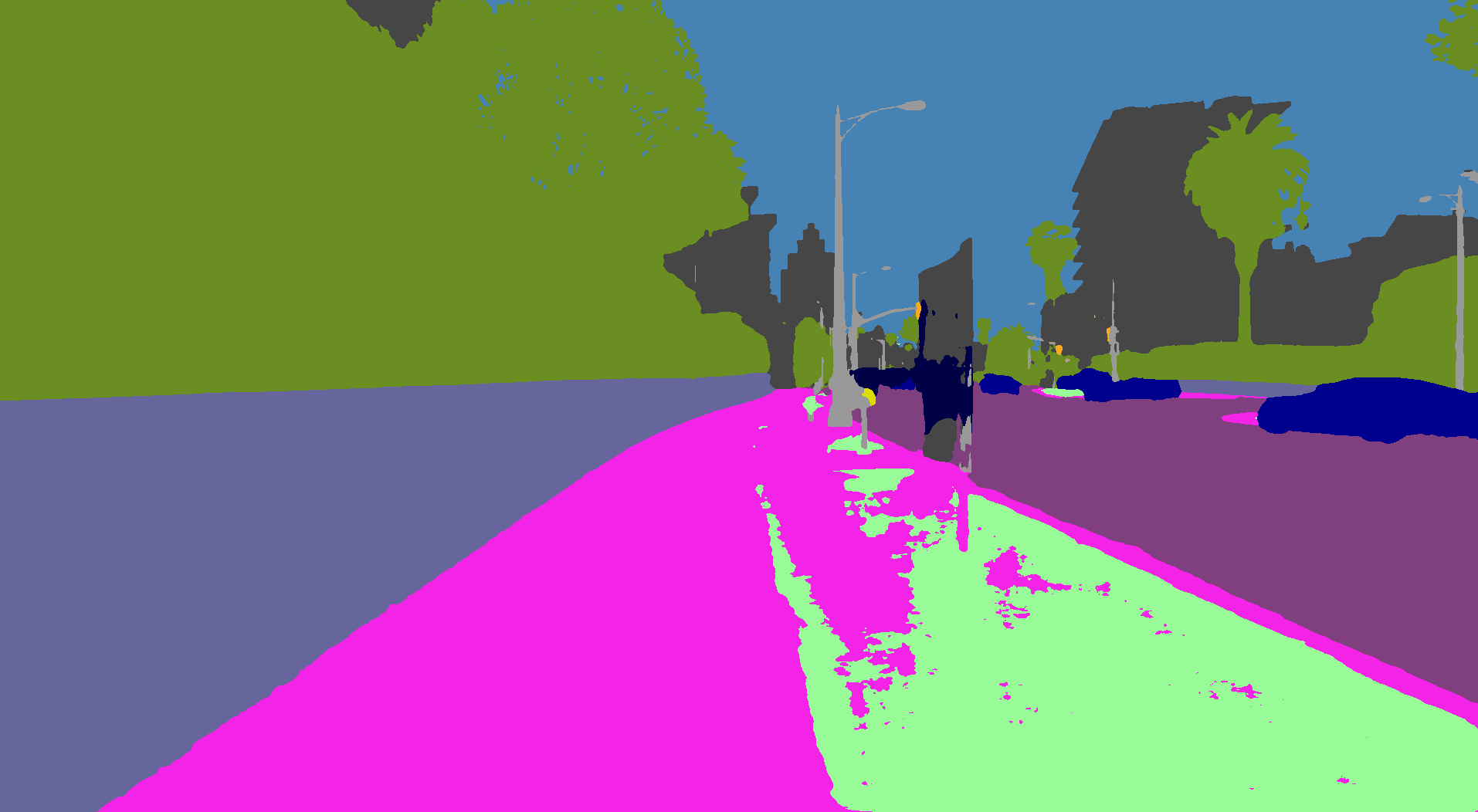}}
  \end{subfigure}
  \hfill
  \begin{subfigure}{0.19\textwidth}
    \raisebox{-\height}{\includegraphics[width=\textwidth, height=0.55\textwidth]{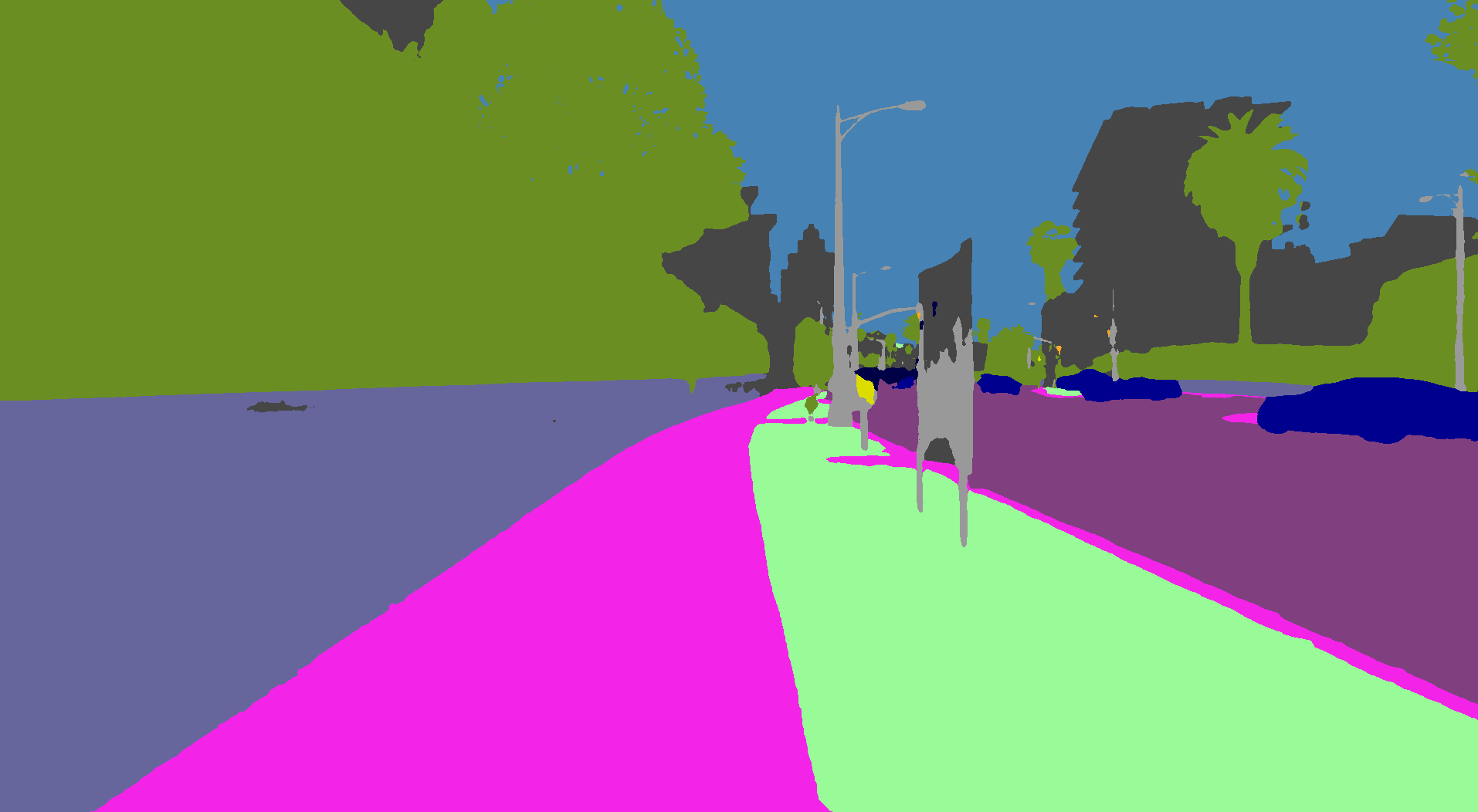}}
  \end{subfigure}
  \hfill
  \begin{subfigure}{0.19\textwidth}
    \raisebox{-\height}{\includegraphics[width=\textwidth, height=0.55\textwidth]{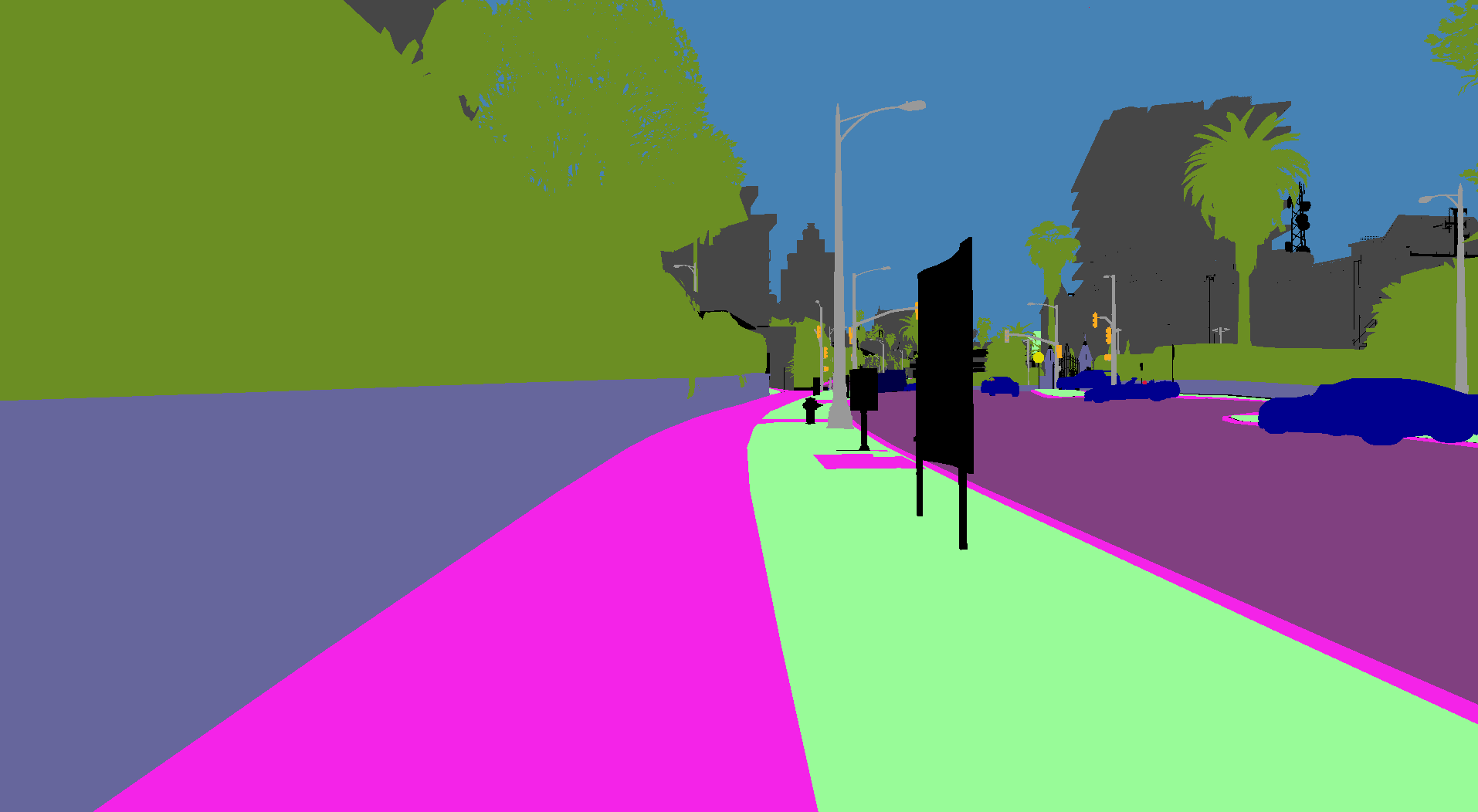}}
  \end{subfigure}
  \begin{subfigure}{0.19\textwidth}
    \raisebox{-\height}{\includegraphics[width=\textwidth, height=0.55\textwidth]{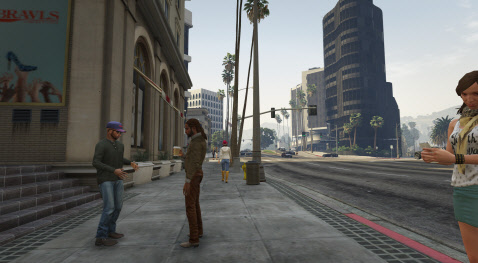}}
  \end{subfigure}
  \hfill
  \begin{subfigure}{0.19\textwidth}
    \raisebox{-\height}{\includegraphics[width=\textwidth, height=0.55\textwidth]{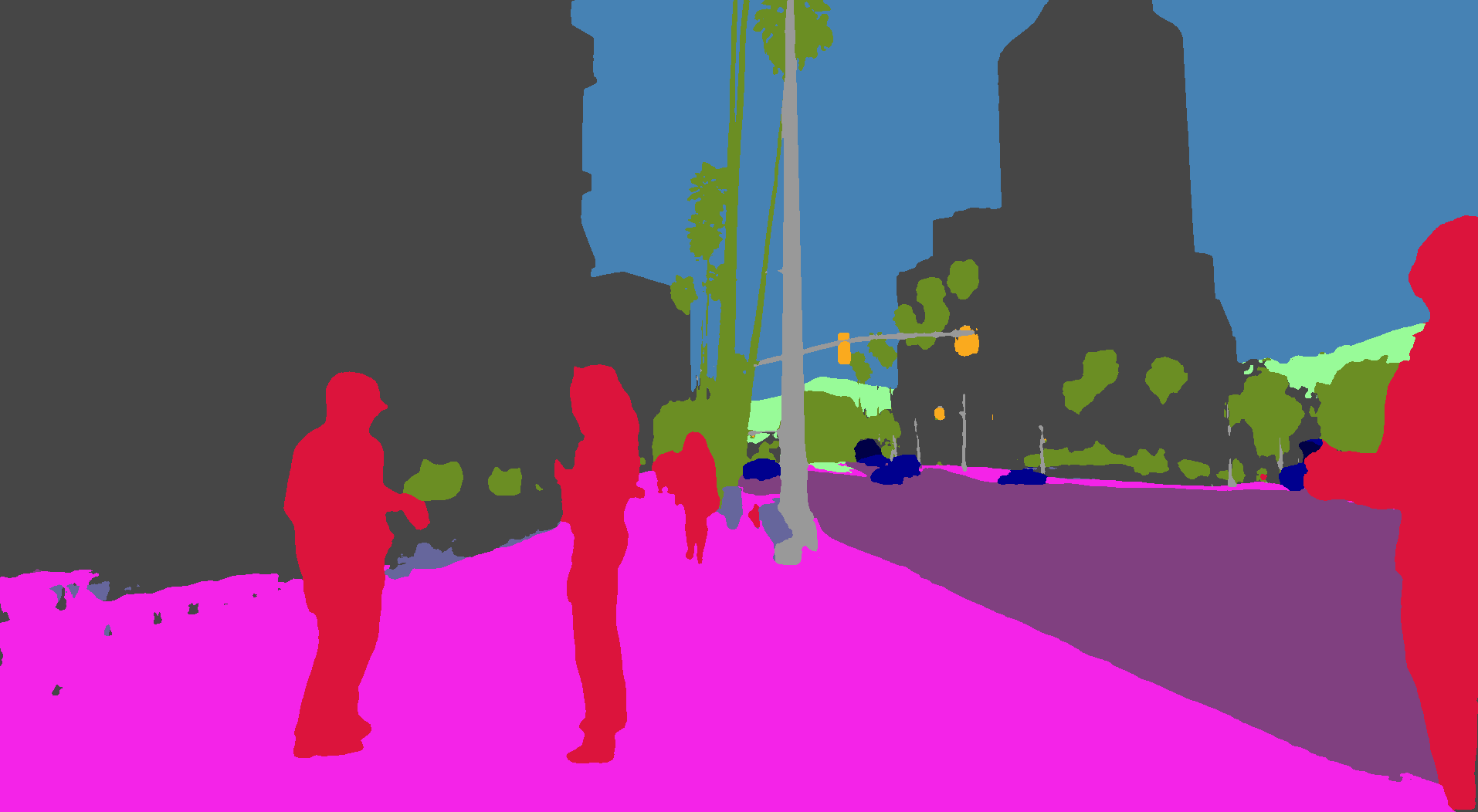}}
  \end{subfigure}
  \hfill
  \begin{subfigure}{0.19\textwidth}
    \raisebox{-\height}{\includegraphics[width=\textwidth, height=0.55\textwidth]{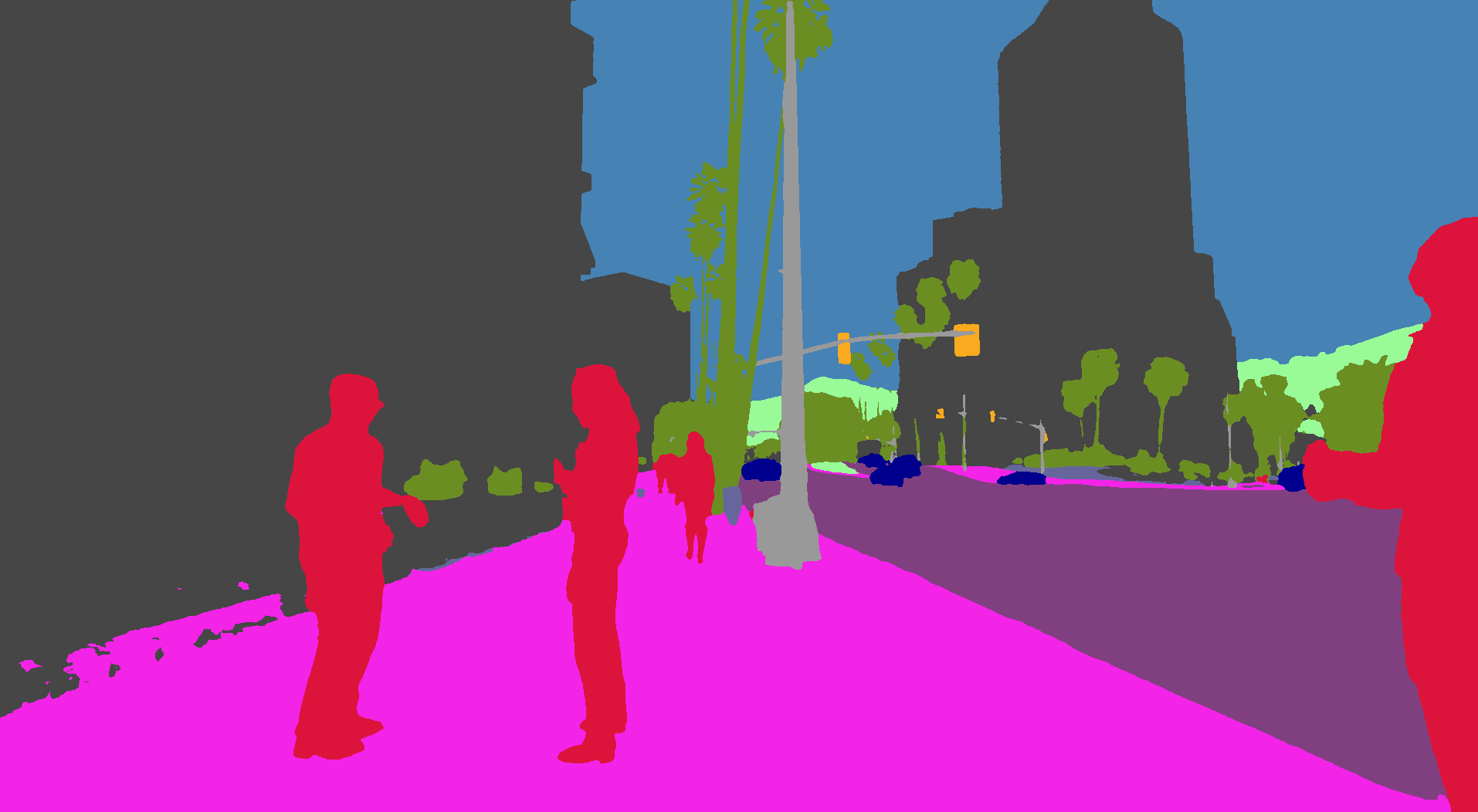}}
  \end{subfigure}
  \hfill
  \begin{subfigure}{0.19\textwidth}
    \raisebox{-\height}{\includegraphics[width=\textwidth, height=0.55\textwidth]{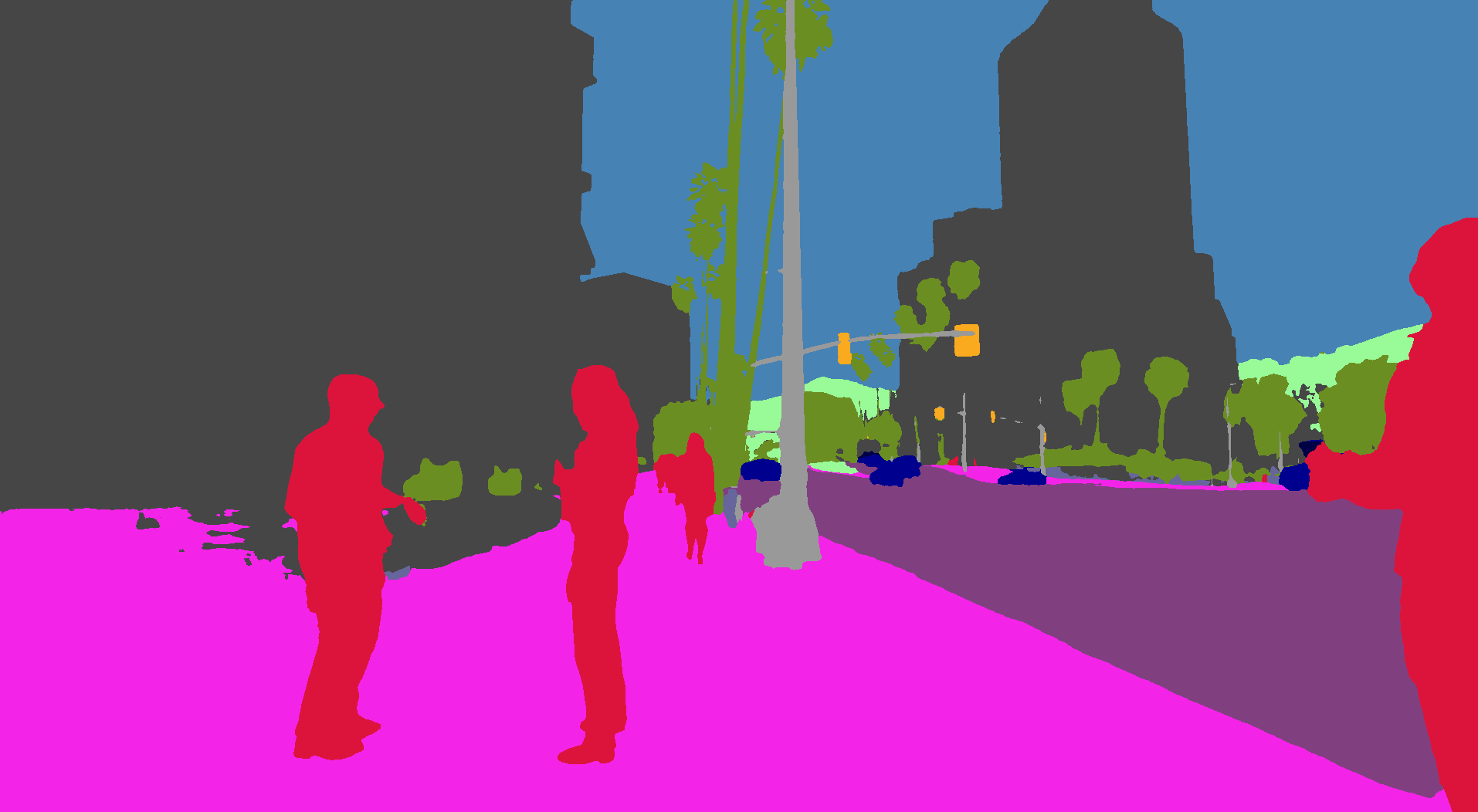}}
  \end{subfigure}
  \hfill
  \begin{subfigure}{0.19\textwidth}
    \raisebox{-\height}{\includegraphics[width=\textwidth, height=0.55\textwidth]{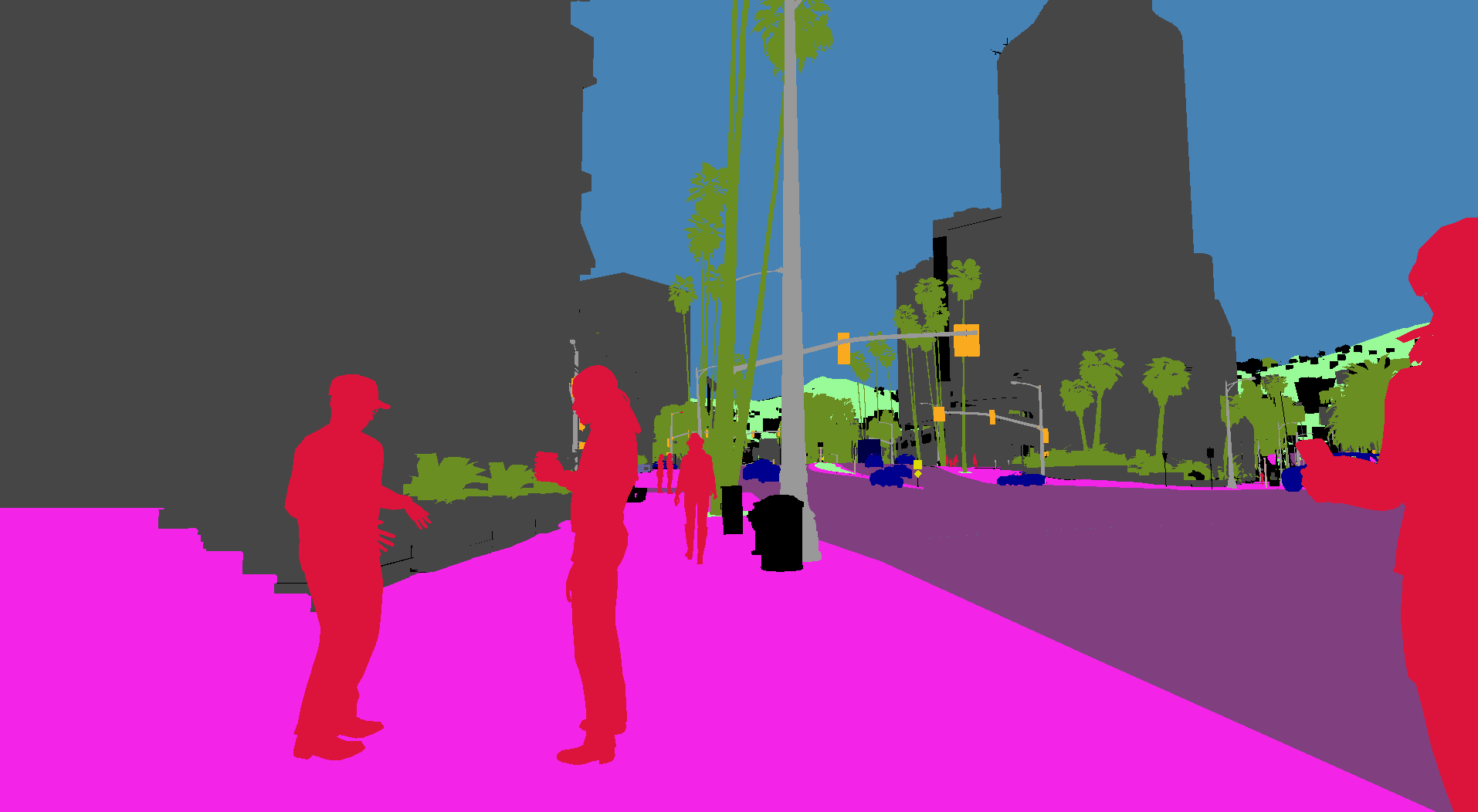}}
  \end{subfigure}
  \begin{subfigure}{0.19\textwidth}
    \raisebox{-\height}{\includegraphics[width=\textwidth, height=0.55\textwidth]{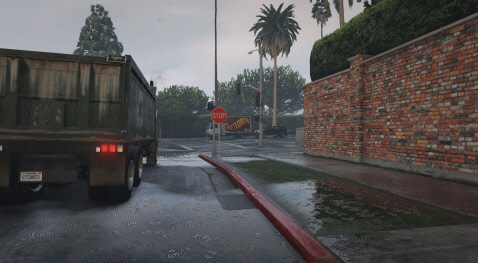}}
    \caption*{Seen domain image}
    \label{fig:compare_r50_gta_img}
  \end{subfigure}
  \hfill
  \begin{subfigure}{0.19\textwidth}
    \raisebox{-\height}{\includegraphics[width=\textwidth, height=0.55\textwidth]{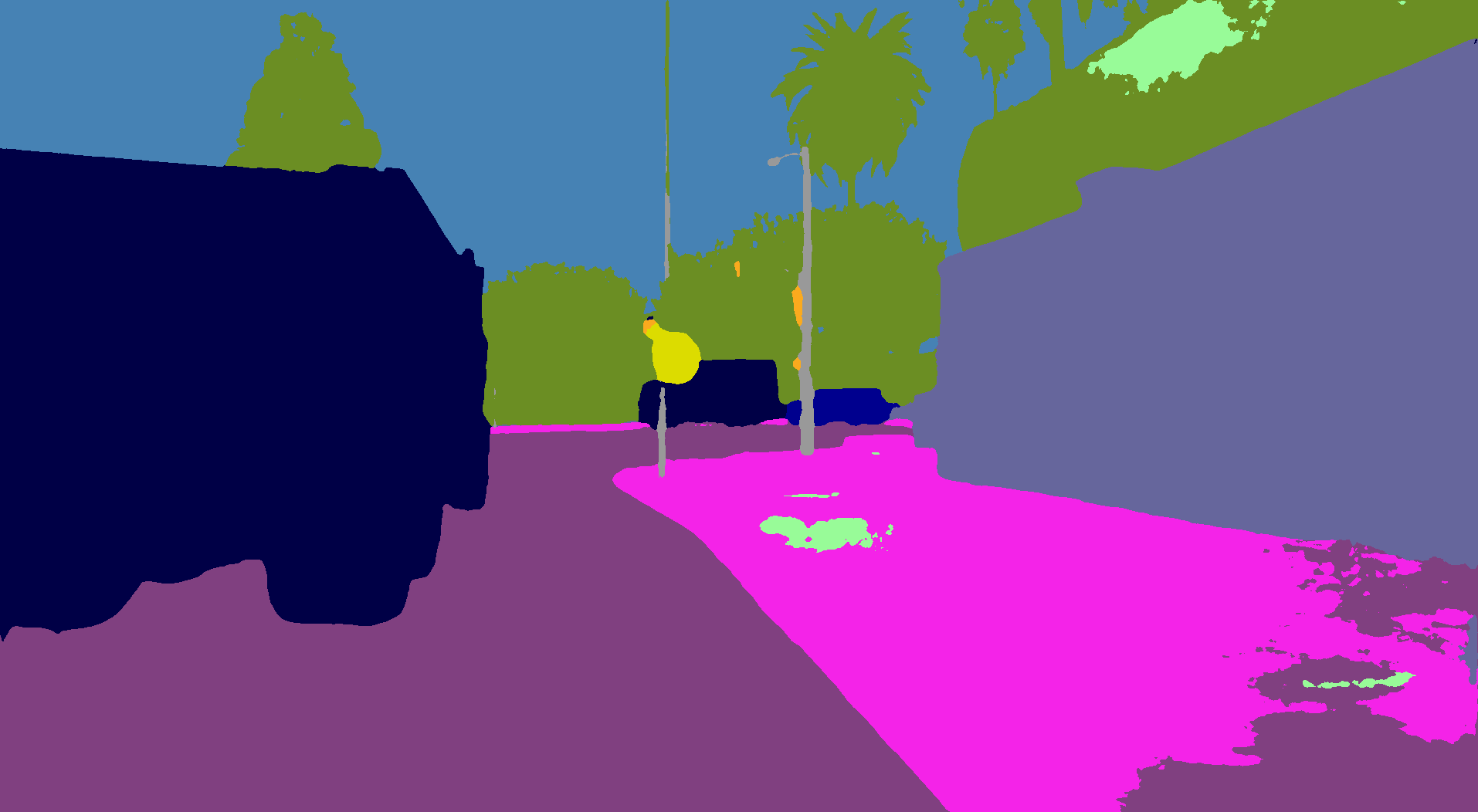}}
    \caption*{Baseline}
    \label{fig:compare_r50_gta_base}
  \end{subfigure}
  \hfill
  \begin{subfigure}{0.19\textwidth}
    \raisebox{-\height}{\includegraphics[width=\textwidth, height=0.55\textwidth]{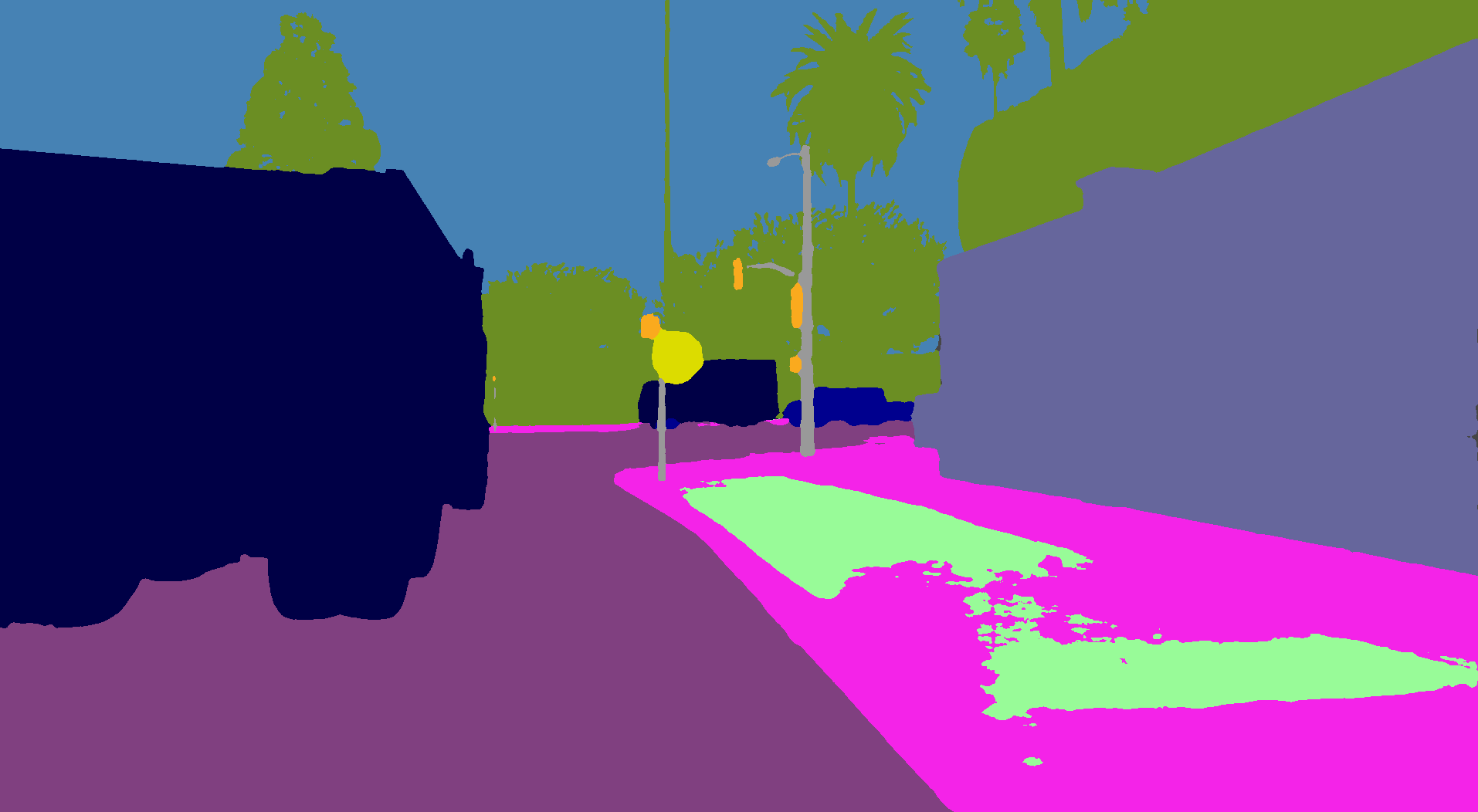}}
    \caption*{RobustNet}
    \label{fig:compare_r50_gta_isw}
  \end{subfigure}
  \hfill
  \begin{subfigure}{0.19\textwidth}
    \raisebox{-\height}{\includegraphics[width=\textwidth, height=0.55\textwidth]{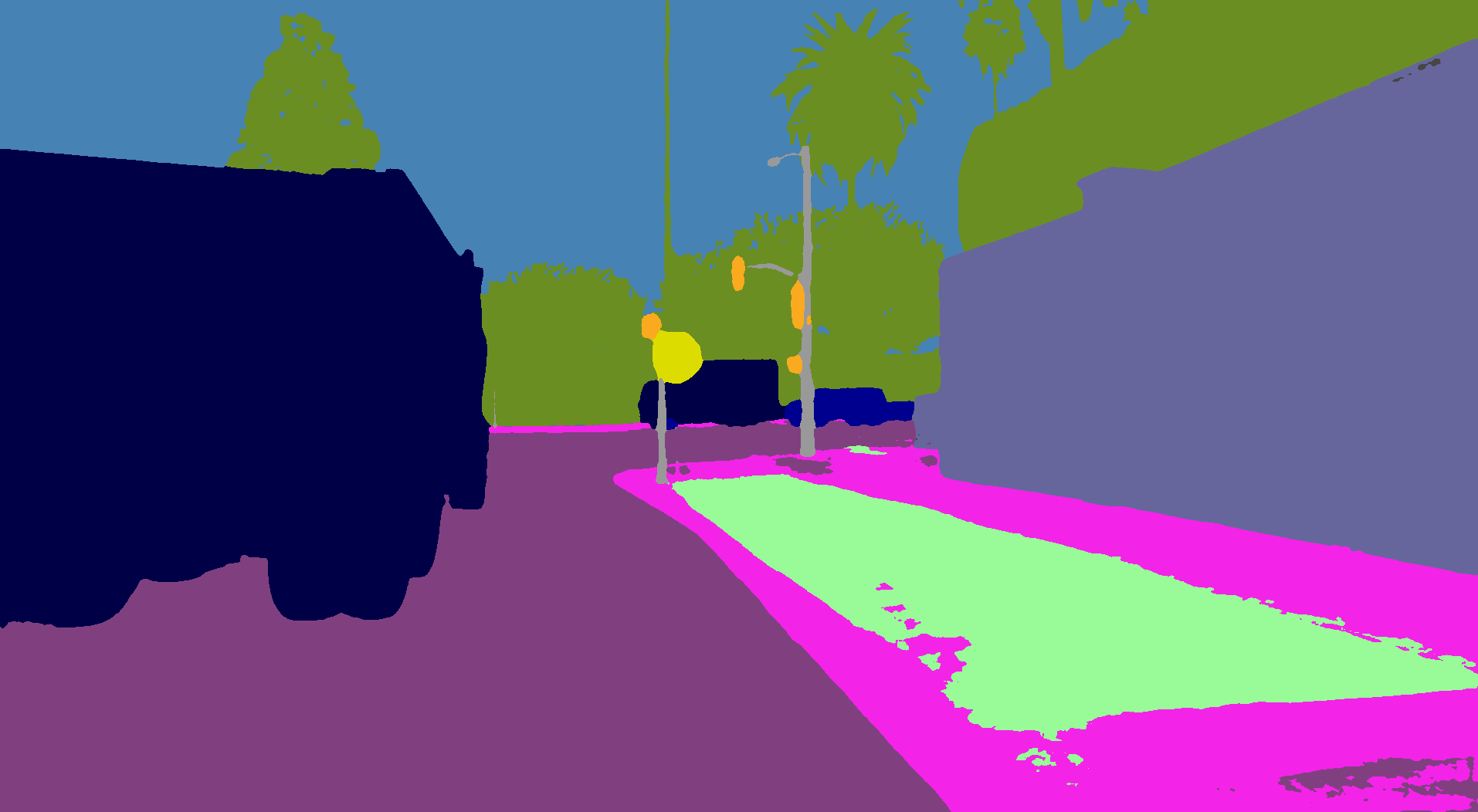}}
    \caption*{\textbf{Ours (WildNet)}}
    \label{fig:compare_r50_gta_ours}
  \end{subfigure}
  \hfill
  \begin{subfigure}{0.19\textwidth}
    \raisebox{-\height}{\includegraphics[width=\textwidth, height=0.55\textwidth]{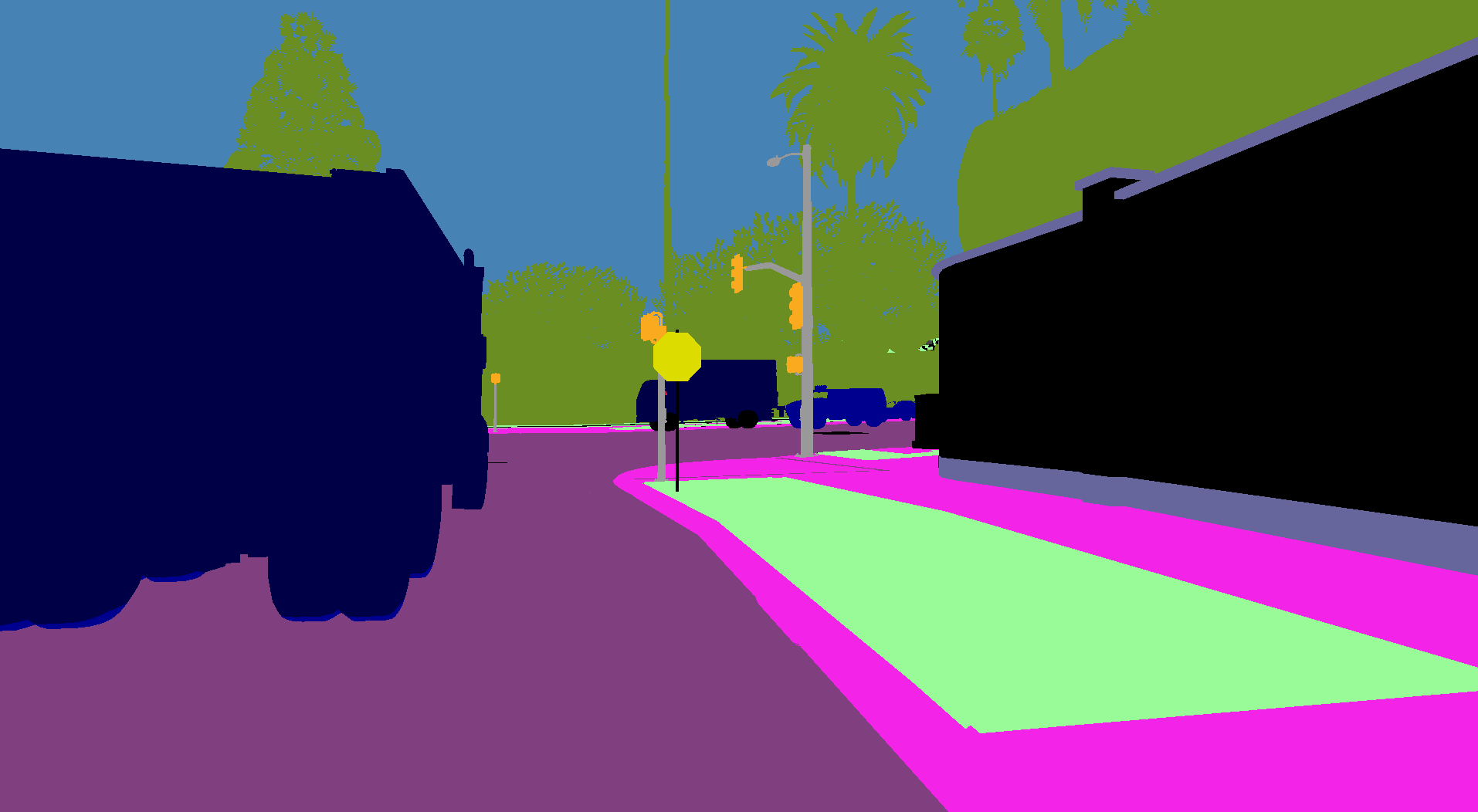}}
    \caption*{Ground truth}
    \label{fig:compare_r50_gta_gt}
  \end{subfigure}
  \caption{Semantic segmentation results on seen domain images in GTAV with the models trained on GTAV.
  }
  \label{fig:compare_r50_gta_img_base_isw_ours}
\end{figure*}

\vspace{-0.3em}
In~\cref{fig:compare_r50_bdd_img_base_isw_ours,fig:compare_r50_map_img_base_isw_ours,fig:compare_r50_cty_img_base_isw_ours,fig:compare_r50_syn_img_base_isw_ours,fig:compare_r50_gta_img_base_isw_ours}, we present semantic segmentation results on four unseen domain validation sets (\ie, Cityscapes~\cite{cordts2016cityscapes}, BDD100K~\cite{yu2020bdd100k},  Mapillary~\cite{neuhold2017mapillary}, and SYNTHIA~\cite{ros2016synthia}) and a seen domain validation set (\ie, GTAV~\cite{richter2016playing}).
We used ResNet-50 as the backbone network and trained on GTAV train set.
To show the efficacy of the proposed method, we additionally present the results of the baseline and RobustNet~\cite{choi2021robustnet}.
As shown in~\cref{fig:compare_r50_bdd_img_base_isw_ours,fig:compare_r50_map_img_base_isw_ours,fig:compare_r50_cty_img_base_isw_ours,fig:compare_r50_syn_img_base_isw_ours}, the baseline model works poorly on the unseen datasets, and RobustNet also often fails.
In contrast, WildNet can accurately segment the road and sidewalk (\eg, the top row in~\cref{fig:compare_r50_bdd_img_base_isw_ours}, the second row in~\cref{fig:compare_r50_map_img_base_isw_ours}, and the second row in~\cref{fig:compare_r50_syn_img_base_isw_ours}) and correctly classify instances (\eg, terrain in the fifth row of~\cref{fig:compare_r50_bdd_img_base_isw_ours}, riders and bicycles in the top row of~\cref{fig:compare_r50_cty_img_base_isw_ours}, and a car in the fifth row of~\cref{fig:compare_r50_syn_img_base_isw_ours}).
Furthermore, as shown in~\cref{fig:compare_r50_gta_img_base_isw_ours}, our WildNet performed well on the seen dataset even in some challenging cases, such as night-time (the top row in the figure), rainy (the fifth row), and backlight (the third row).

\end{document}